%% file: QuantumAgents.tex
\documentclass[12pt, a4paper]{article}
\usepackage[a4paper, top=2.0cm, bottom=1.0cm, left=1.8cm, right=1.5cm, footskip = 1.0cm, includehead, includefoot]{geometry}
\usepackage{helvet} 			

\usepackage{csquotes}
\usepackage[utf8]{inputenc}
\usepackage[final]{graphicx}
\usepackage[margin=20mm, font=footnotesize, labelfont=bf, justification=justified]{caption}
\usepackage{dirtytalk} 
\usepackage{pythonhighlight} 
\usepackage{csquotes}
\usepackage{times}
\usepackage[bottom]{footmisc}	
\usepackage{scrextend}
\deffootnote[1.8em]{1.8em}{1.8em}{\textsuperscript{\thefootnotemark}}
\usepackage{xcolor}			
\usepackage{newtxtext}
\usepackage{newtxmath}       	
\usepackage{braket}
\usepackage{float}			
\usepackage{mathtools}
\usepackage{amsmath}
\usepackage{gensymb}
\usepackage{subfigure}
\usepackage{tipa}
\usepackage{url}
\usepackage{authblk}
\usepackage{blindtext}
\usepackage{hyperref}
\usepackage{xcolor}
\usepackage{titlesec} 
\titleformat{\paragraph}
{\normalfont\normalsize\bfseries}{\theparagraph}{1em}{}
\usepackage{physics}
\usepackage{braket} 
\usepackage{musicography}  
\usepackage{tikz} 
\usetikzlibrary{quantikz}
\usetikzlibrary{matrix}
\usetikzlibrary{arrows.meta,positioning,shapes,calc}
\usepackage{MnSymbol} 
\usepackage{hyperref}
\usepackage{tikzfig}
\usepackage{relsize}
\usepackage[outline]{contour}
\contourlength{.12em}
\input{qcs.tikzstyles}

\input{dots.tikzstyles}
\input{colors.tikzstyles}
\makeatletter
\def\mathcolor#1#{\@mathcolor{#1}}
\def\@mathcolor#1#2#3{%
  \protect\leavevmode
  \begingroup
    \color#1{#2}#3%
  \endgroup
}
\makeatother
\usepackage{paralist}
\usepackage{hyperref}    
\usepackage{cleveref}   
\usepackage{booktabs}
\usepackage{enumitem}
\usepackage{algpseudocode}
\usepackage{algorithm}
\usepackage{physics}
\usepackage{listings}
\title{\huge Beyond Corpus-Driven Imitation: Evolving Musical Mindsets in the Agentic Orchestra via Quantum Reinforcement Learning\footnote{This is the authors' accepted, pre-publication version of a paper forthcoming in \textit{Computer Music Journal} (MIT Press). Beyond routine editorial and typesetting amendments, the content remains subject to further revision before final publication. Please consult the published version for definitive citation.}}
\author[]{Eduardo Reck Miranda\footnote{Corresponding author: eduardo.miranda@plymouth.ac.uk}}
\author[]{Moein Fahmideh Vatandoost}

\affil[]{Interdisciplinary Centre for Computer Music Research (ICCMR), University of Plymouth, Plymouth PL4 8AA, UK}

\begin{document}

\maketitle
\abstract{This paper presents a framework for computer-aided music composition and performance based on quantum reinforcement learning and multi-agent interaction. We introduce the \textit{agentic orchestra}, where musical agents evolve distinct mindsets through trial-and-error engagement with a musical environment, rather than through training on large datasets. The agents are implemented as parameterised quantum circuits. They evolve distinct compositional behaviours, referred to as \textit{mindsets}. This enables the exploration of musical decision spaces using quantum superposition and entanglement within a hybrid quantum-classical loop. We present two complementary agent architectures: atomistic agents, which process pitch and rhythm independently, and holistic agents, which handle pitch and duration as unified note events. Mindsets are shaped through musically grounded reward functions, which can be combined and weighted to develop different mindsets across agents. The resulting system positions quantum computing agents less as autonomous composers and more as adaptive musical partners, each developing its own musical personality through interaction rather than imitation. Demonstrations show how agents generalise their learned behaviour to previously unseen material, operating in interactive and offline batch modes using MIDI-encoded music. While quantum computers are still in development, experiments conducted on IBM quantum hardware confirmed that musically meaningful learning is achievable with existing quantum computers. For composers and performers, the agentic orchestra opens new possibilities for emergent, process-oriented musical creativity, offering a scalable foundation for quantum computer-aided music composition as quantum computing technology continues to advance.}

\section{Introduction}

Generative methods based on artificial intelligence (AI) have long been a core component of computer-aided music composition and performance systems, enabling the automated creation and transformation of musical material  \cite{Balaban_1992, Miranda2021, Miranda_2000, Nierhaus_2008}. The evolution of these methods has broadly followed the trajectory of AI research: from symbolic, rule-based systems \cite{Biles2007, Bouche_etal_2017, Cope1996, Hiller_Isaacson1959, Lerdahl_Jackendoff_1983, Lewis2000, Miranda1995, Rowe2004} through to connectionist architectures  \cite{Grittith_Todd1999, Todd1989}, and more recently to data-driven deep learning methods that currently dominate the field  \cite{Goodfellow_etal2016, Huang2017, Magenta,  Oord_etal2016}. Today, commercially available systems such as Suno and Udio  \cite{Vila2025} exemplify this trajectory, offering music generation through large-scale deep learning neural networks trained on vast corpora, or \enquote{big data}.

\medskip

Yet despite their often impressive outcomes, corpus-trained deep learning-based systems carry a fundamental limitation for composers and musicians seeking genuinely novel creative tools. These systems are engineered to optimise against patterns in training data. Results tend toward stylistic reproduction, recombination and pastiche. They often reflect the aesthetics of their training data rather than opening new artistic directions. From a compositional standpoint, this creates a feedback loop of stylistic stagnation that is difficult to circumvent within the data-driven paradigm. Such systems raise a fundamental question: what does it mean to build AI music systems that genuinely explore rather than merely reproduce or imitate?

\medskip

In this paper, we propose an alternative route. Rather than training a neural network on large corpora or encoding a fixed set of explicit musical rules, we present a quantum computing agentic system in which compositional behaviours emerge through interaction with a musical environment. The system learns from the consequences of its own musical actions, rather than through supervised imitation of existing music.

\medskip

We refer to our system as an \textit{agentic orchestra}. Each agent in the agentic orchestra is implemented as a parameterised quantum circuit (PQC) trained via quantum reinforcement learning (QRL). Instead of learning to reproduce patterns from existing music, agents evolve distinct musical \textit{mindsets}. Mindsets are learned dispositions toward particular compositional behaviours, such as rhythmic variety, melodic balance, or textural smoothness. Metaphorically, each agent develops its own musical skills and personality by engaging with musical material and learning from the consequences of its decisions.

\medskip

We are interested in systems that behave as instruments, environments, or collaborators rather than as replacement authors \cite{Miranda_2002, Miranda1995, Roads_2023}. Like the interactive improvisation systems and generative environments that have shaped the field of computer music (e.g., \cite{Assayag_Dubnov2004, Biles2007, Lewis2000, Rowe2004}), the agentic orchestra is conceived as an interactive partner. By exploiting quantum-mechanical phenomena such as superposition, entanglement, and interference, quantum computing enables agents to explore musical decision spaces in fundamentally different ways than traditional algorithms \cite{LaPierre2021, Miranda_2022, Weaver2022}. Moreover, they do so without the big data dependency that constrains contemporary deep learning approaches.

\medskip

The agentic orchestra does not require fault-tolerant quantum hardware. We demonstrate that musically meaningful agent behaviour can be achieved on current Noisy Intermediate-Scale Quantum (NISQ) devices \cite{Preskill2018} and we validate this with experiments on state-of-the-art IBM quantum hardware. The main contributions of this paper are as follows:

\begin{itemize}
    \item It introduces the concept of the \textit{agentic orchestra}, a framework whereby agents evolve reusable musical mindsets through QRL.
    \item It presents a PQC-based reinforcement learning architecture for music composition and performance.
    \item It defines atomistic and holistic agent designs for pitch, duration, and pitch-duration event (i.e., note) processing.
    \item It introduces musically interpretable action spaces and reward functions for shaping distinct compositional behaviours.
    \item It demonstrates how trained agents can transform MIDI-encoded musical material and act as adaptive musical partners within a creative workflow.
\end{itemize}

To the best of our knowledge, this is the first application of QRL to computer-aided music composition and performance. The remainder of the paper details the system architecture, the quantum learning process, and a series of musical demonstrations.

\section{System Overview: the Agentic Orchestra}

A group of agents comprises an \textit{agentic orchestra}. The notion of the agentic orchestra was conceived for interactive music with human performers. However, it can also be employed in batch processing mode to generate music from given musical recordings. The outputs from the orchestra can be played on the fly, saved as files, or both. 

\medskip

Members of the orchestra have different musical skills and preferences, referred to as \textit {mindsets}. Upon listening to an input, all agents may compose variations according to their learned mindsets. Then, the orchestra may elect one of them, sequence some of them, or sequence all of them for playback, respectively. The choices and sequencing can either be random, quantum-sampled, or guided by user-defined criteria. 

\medskip

In case there is more than one performer, the agents are subdivided and allocated accordingly. The current version of the system works with monophonic inputs and outputs. For convenience, for the demonstrations in Section \ref{sec:demos}, we used MIDI files{\footnote{MIDI (Musical Instrument Digital Interface) is a standard protocol that allows computers, electronic musical instruments, and software to communicate, record, and edit music.} for input and output in batch processing mode.

\begin{figure}[H]
\begin{center}\vspace{0.3cm}
\includegraphics[width=0.8\linewidth]{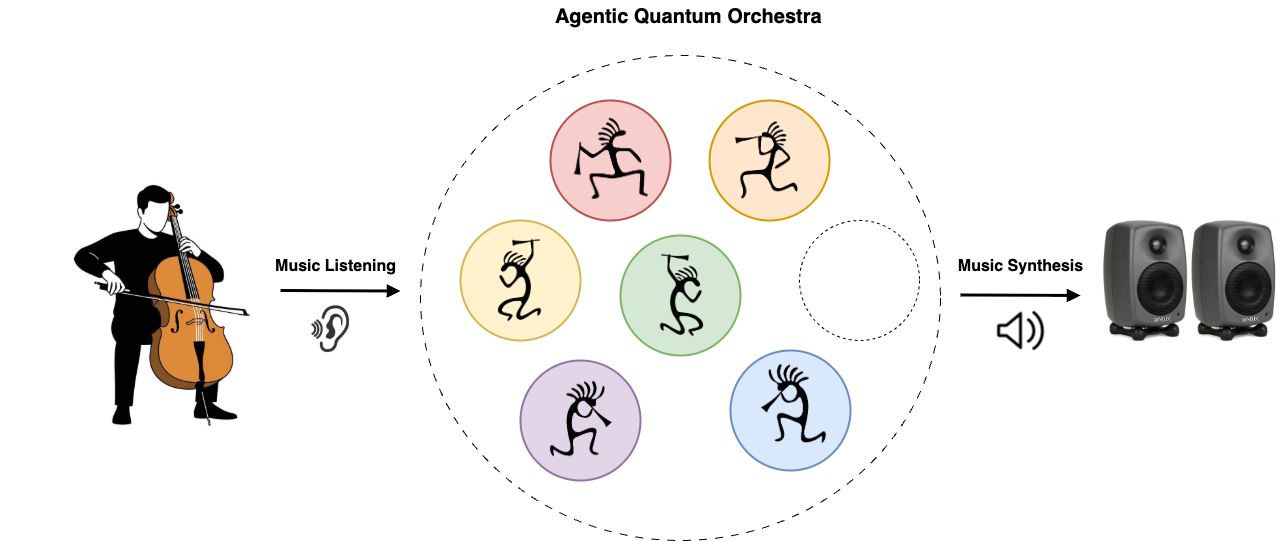}
\caption{The agentic orchestra supports interactive and batch-processing workflows.}
\label{fig:agentic-orchestra}
\end{center}
\end{figure}

Before joining an orchestra, agents first evolve their individual musical mindsets through QRL (Section \ref{sec:QRL}). During this training phase, an agent is given a snippet of music, a set of actions and one or more reward criteria. The actions are musical transformation operators that modify the snippet and can differ from agent to agent. Furthermore, the agents are set with criteria to evaluate the outcomes of the transformations. The criteria can also differ among the agents. The objective is to learn which actions need to be applied to a given input to maximise the reward associated with their mindset-defining criteria (Figure \ref{fig:QRL-agent}).

\begin{figure}[H]
\begin{center}\vspace{0.3cm}
\includegraphics[width=0.36\linewidth]{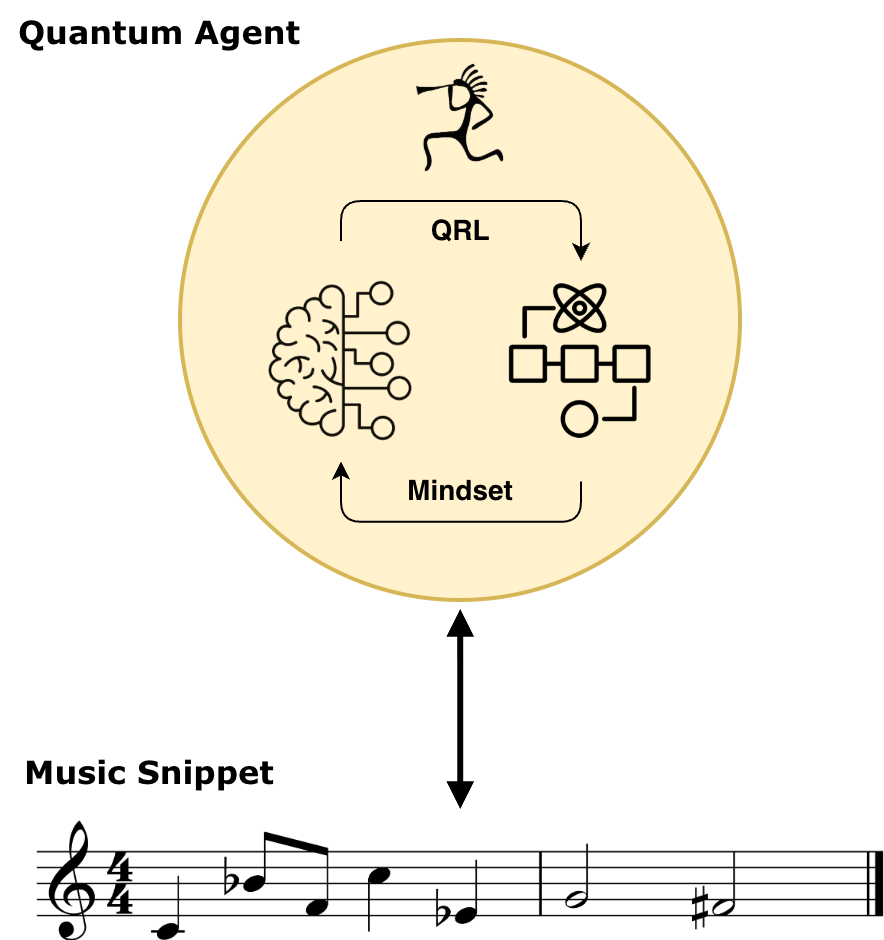}
\caption{Agents evolve mindsets by interacting with a snippet of music.}
\label{fig:QRL-agent}
\end{center}
\end{figure}

\medskip

The agents, therefore, evolve their mindset by trying out actions and combinations thereof on given snippets of music. After training, they will be able to apply their evolved knowledge to musical inputs other than the snippets used for training. In this sense, each trained agent functions as a reusable musical disposition: one may favour balanced rhythmic organisation, another may encourage melodic diversity, and another may produce smoother transitions. Metaphorically, each agent develops a musical personality through interaction with musical material. In the demonstrations reported in Section \ref{sec:demos}, each mindset is evolved from short snippets supplied as a MIDI file.

\section{Quantum Reinforcement Learning}
\label{sec:QRL}

Reinforcement learning (RL) is a well-studied learning paradigm in Cognitive Neuroscience \cite{Collins2024}. In computer modelling of RL, a software agent learns to act in an \textit{environment} by trial and error. The agent tries different actions and, based on whether those actions yield a reward (positive feedback) or a penalty (negative feedback), it iteratively refines its behaviour to accumulate the highest possible total reward over time.

\medskip

 RL is formally modelled as a Markov Decision Process (MDP). MDP is a mathematical framework used to model decision-making in situations where outcomes are partly random and partly controlled by an agent. It is characterized by the following tuple elements $\langle\mathcal{S},\mathcal {A}, \mathcal{T}, \mathcal{R},\gamma\rangle$:

\begin{itemize}
	\item{State space $\mathcal{S}$}: the complete set of possible states the environment can be in.
	\item{Action space $\mathcal{A}$}: the set of actions available to the agents.
	\item{Transition function $\mathcal{T}(s'|s, a)$}: the probability of transitioning to state $s'$ after applying action $a$ on state $s$.
	\item{Reward function $\mathcal{R}(s, a)$}: the scalar reward (or penalty) the agent receives upon transitioning from one state to another via action $a$.
	\item{Discount factor $\gamma \in \{0.0, \dots, 1.0\}$}:  weights the importance of future rewards relative to immediate ones.  A value close to 1.0 makes the agent \enquote{far-sighted}; a value close to 0.0 focuses it on immediate gains.  The default in our implementation is $\gamma = 0.99$.
\end{itemize}

\medskip

At each time-step $t$, the agent observes a state $s_t$, selects an action $a_t$, transitions to a new state $s_{t+1}$, and receives a scalar reward $r_t = \mathcal{R}(s_t,a_t)$. The goal is to learn a policy $\pi(a|s)$, that is, a mapping from states to actions, which maximises the expected sum of discounted future rewards.

\medskip
Early RL methods such as Q-learning \cite{watkins1992qlearning} and SARSA \cite{zhao_deep_sarsa} maintain a table of action-value estimates $Q(s,a)$. The optimal policy is then derived by selecting the action with the highest estimated value for a given state, as shown in Eq.~\ref{eq:action_values}. These methods iteratively update $Q(s, a)$ using observed transitions and their associated rewards.

\begin{equation}
 	\pi^*(s) = \arg {\max}_{a} \ Q(s,a)
	\label{eq:action_values}
\end{equation}

\medskip

While conceptually simple, classical Q-learning becomes impractical when the state or action spaces are large \cite{Sutton_Barto2018}. Deep Q-Networks (DQN) \cite{fan2020theoretical} overcome this by replacing the table with a neural network that \textit{approximates} $Q(s, a)$. This enables RL to scale to high-dimensional inputs such as images and raw sensor data.  Two key techniques known as \textit{experience replay} \cite{Lin1992, fan2020theoretical} and \textit{target networks} \cite{Mnih2015} have further stabilised training, making deep RL reliable in complex domains ranging from game-playing \cite{Silver2017} to robotics \cite{Wang2023}.

\medskip

Building on DQN, an emerging approach integrates quantum computing within RL. This approach, referred to as Quantum Reinforcement Learning (QRL), explores the unique properties of quantum mechanics, such as superposition, entanglement and interference, to process information \cite{LaPierre2021, nielsen_chuang}. Instead of a classical neural network, QRL uses a Parameterised Quantum Circuit (PQC), also referred to as a Variational Quantum Circuit (VQC), as the function approximator for $Q(s, a)$ or the policy $\pi(a | s)$ \cite{dong2008quantum, jerbi2019quantum, schuld2020circuit}.

\medskip

Quantum computing encodes information in qubits (i.e., quantum bits) instead of classical binary bits. A qubit can be represented as a superposition state $\ket{\psi} = \alpha\ket{0} + \beta\ket{1}$, where $\alpha,\beta \in \mathbb{C}$ and satisfy $|\alpha|^2 + |\beta|^2 = 1$. In simple terms, $\alpha$ and $\beta$ are probability amplitudes. Upon measurement in the computational basis, the qubit collapses to $\ket{0}$ with probability $|\alpha|^2$ and to $\ket{1}$ with probability $|\beta|^2$. Quantum computation proceeds by applying unitary operators $U$ to quantum states, such that $\ket{\psi'} = U\ket{\psi}$.

\medskip

When multiple qubits are combined, their joint state creates a $2^n$-dimensional Hilbert space, enabling exponentially large representations and correlations via entanglement such as, for example, the Bell state $\ket{\Phi^+} = \tfrac{1}{\sqrt{2}}(\ket{00} + \ket{11})$: if the measurement on one qubit yields $0$, the other is guaranteed to be $0$ too, and if it yields $1$, the other will be $1$ as well. Quantum algorithms manipulate these high-dimensional amplitudes through sequences of quantum gates so that measurement probabilities encode solutions to computational problems, given by the Born rule $P(i)=|\langle i|\psi\rangle|^2$. This mathematical structure is what makes quantum computing attractive for fields such as QRL, where quantum state spaces, amplitude encoding, and unitary operators may yield speedups in exploration, optimisation, and sampling. For a quantum computing primer, please refer to the first chapter of the book \textit{Quantum Computer Music} \cite{Miranda_2022}. The book \textit{Introduction to Quantum Computing} \cite{LaPierre2021} provides a comprehensive introduction to the field.

\medskip

When combined with standard RL frameworks such as Q-learning or policy gradients, PQCs serve as differentiable models that process quantum-encoded input states and output action-value estimates $Q(s, a)$ or policy distributions $\pi(a | s)$ \cite{dong2008quantum, jerbi2019quantum, schuld2020circuit}. PQCs introduce learnable parameters as rotation angles in quantum gates (i.e., unitary operators $U$,  $\ket{\psi'} = U \ket{\psi}$), which can be updated via gradient-based or gradient-free methods to minimise the loss function (Figure \ref{fig:qrl-pitch-circuit}). Early studies suggest these hybrid models may achieve superior performance to classical deep RL as quantum hardware scales up \cite{Kunczik2022, lockwood2020reinforcement}.
\medskip

Despite its relative nascency, QRL has been explored in grid-world tasks, robotic controls, and function approximation scenarios \cite{chen2020variational}. This paper reports what we believe to be the first application of QRL for music: each musical agent in our agentic orchestra is implemented as a PQC trained via QRL.

\section{The Agents}
\label{sec:agents}

Each agent in the agentic orchestra is implemented as a PQC whose trainable parameters $\theta$ determine its learned musical behaviour, or \textit{mindset}. The system supports two agent architectures: \textit{holistic} and \textit{atomistic} agents. Holistic agents process a musical note as a single entity, whereas atomistic agents handle the pitches and durations independently. Essentially, both architectures share the same underlying quantum circuit design. They differ in the type of musical information they encode and the action spaces available to them.

\medskip

In the descriptions below, we first define the number of qubits required to encode a single symbolic element: one pitch, one duration, or one pitch-duration event. During learning, however, the observable state is a sliding window of size $k$ (Figure \ref{fig:windows}). Therefore, if one symbolic element requires $m$ qubits, the full window state requires $n_q = k \times m$ qubits. For instance, if the default window size is $k=4$, then a four-element window is encoded rather than a single isolated symbol.

\medskip

\subsection{Holistic Agents}

The unique events alphabet is defined by $\mathcal E = \mathcal P \times \mathcal D$, giving $|\mathcal E| = |\mathcal P| \times |\mathcal D| = 64\times8 = 512$. The pitch alphabet is $\mathcal P=\{p_0,\dots,p_{63}\}$, with $|\mathcal P|=64$, where $p_0=\textit{rest}$ and $p_i\in\mathbb Z$ denotes a MIDI pitch number between 33 (A1) and 96 (C7); this range can be customized. In turn, the duration alphabet is defined by the ordered set $\mathcal D=\{d_0,\dots,d_7\}$, with $|\mathcal D|=8$, where $(d_0, \dots, d_7)=(\musSixteenth, \musEighth, \musEighthDotted, \musQuarter, \musQuarterDotted, \musHalf, \musHalfDotted, \musWhole)$.

\medskip

With a sliding window of size $k$, a holistic agent therefore operates on $n_q=9k$ qubits; for the demonstrations with $k=4$, this gives $n_q=36$.

\medskip

The action space for holistic agents comprises $|\mathcal A_\text{holistic}| = 57$ operators (see Appendix I).

\subsection{Atomistic Agents}

Atomistic agents also require up to nine qubits, but they work with two separate sub-registers:

\begin{itemize}
\item{\textbf{Pitch sub-register}: up to six qubits encode the 64-element pitch alphabet $|\mathcal P|$, with the same range as above.}
\item{\textbf{Duration sub-register}: up to three qubits to encode the 8-element duration alphabet $|\mathcal D|$.}
\end{itemize}

\medskip

With a sliding window of size $k$, the atomistic pitch register uses $n_q=6 \times k$ qubits, and the atomistic duration register uses $n_q=3 \times k$ qubits. Thus, for instance, if $k=4$, then pitch agents would use $24$ qubits and duration agents would use $12$ qubits.

\medskip

Because pitch and duration are encoded and acted upon independently, an atomistic agent can specialise entirely in one dimension. For instance, reshaping rhythms while leaving pitches unchanged, or vice versa.  The pitch action space has $|\mathcal A_\text{pitch}| = 50$ operators, and the duration action space has $|\mathcal A_\text{dur}| = 13$ operators (see the Appendix I).

\subsection{The Ansatz: Circuit Architecture}

Regardless of type, every agent is implemented as the same two-layer PQC ansatz $U_{\theta}$. The circuit operates on $n_q$ qubits and is defined as shown in Eq. \ref{eq:agent-ansatz}. 

\begin{equation}
U_{\theta}
=
\prod_{\ell=1}^{L}
\left[
\left(
\prod_{q=0}^{n_q-1}
R_Z(\theta_{\ell, q})
R_Y(\theta_{\ell, n_q+q})
\right)
\left(
\prod_{q=0}^{n_q-2}
CX_{q, q+1}
\right)
\right].
\label{eq:agent-ansatz}
\end{equation}

\medskip

The equation is applied for 2 layers ($L = 2, \ell = 1, 2$), as follows:

\begin{itemize}
\item{\textbf{Rotation layer:} for every qubit $q$ in layer $\ell$, two successive rotation gates are applied. First $R_Z(\theta_{\ell,q})$, which is a rotation about the Z-axis of the qubit by the trainable angle $\theta_{\ell,q}$, and then $R_Y(\theta{\ell, n_{q+q}})$, which is a rotation about the Y-axis by the trainable angle $\theta_{\ell, n_{q+q}}$.  The index $n_{q + q}$ distinguishes the Y-axis angles from the Z-axis angles within the same layer.  These parametrised rotations are the learnable \enquote{weights} of the quantum circuit.}
\item{\textbf{Entangling layer:} a chain of CNOT gates $CX_{q, q+1}$ is applied for $q = 0, 1, \ldots, n_{q − 2}$.  Each CNOT gate flips the target qubit ($q + 1$) conditionally on the control qubit ($q$).  This chain creates entanglement across all adjacent qubit pairs, allowing the circuit to learn correlations between different parts of the encoded musical context.}
\end{itemize}

\medskip

Because $L = 2$, the full circuit consists of two repetitions of [rotation layer $\rightarrow$ entangling layer], giving a total of $2Ln_q = 4n_q$ trainable angles $\theta \in \mathbb{R}^{2Ln_q}$.  All angles are initialised uniformly on $[0, 2\pi]$.

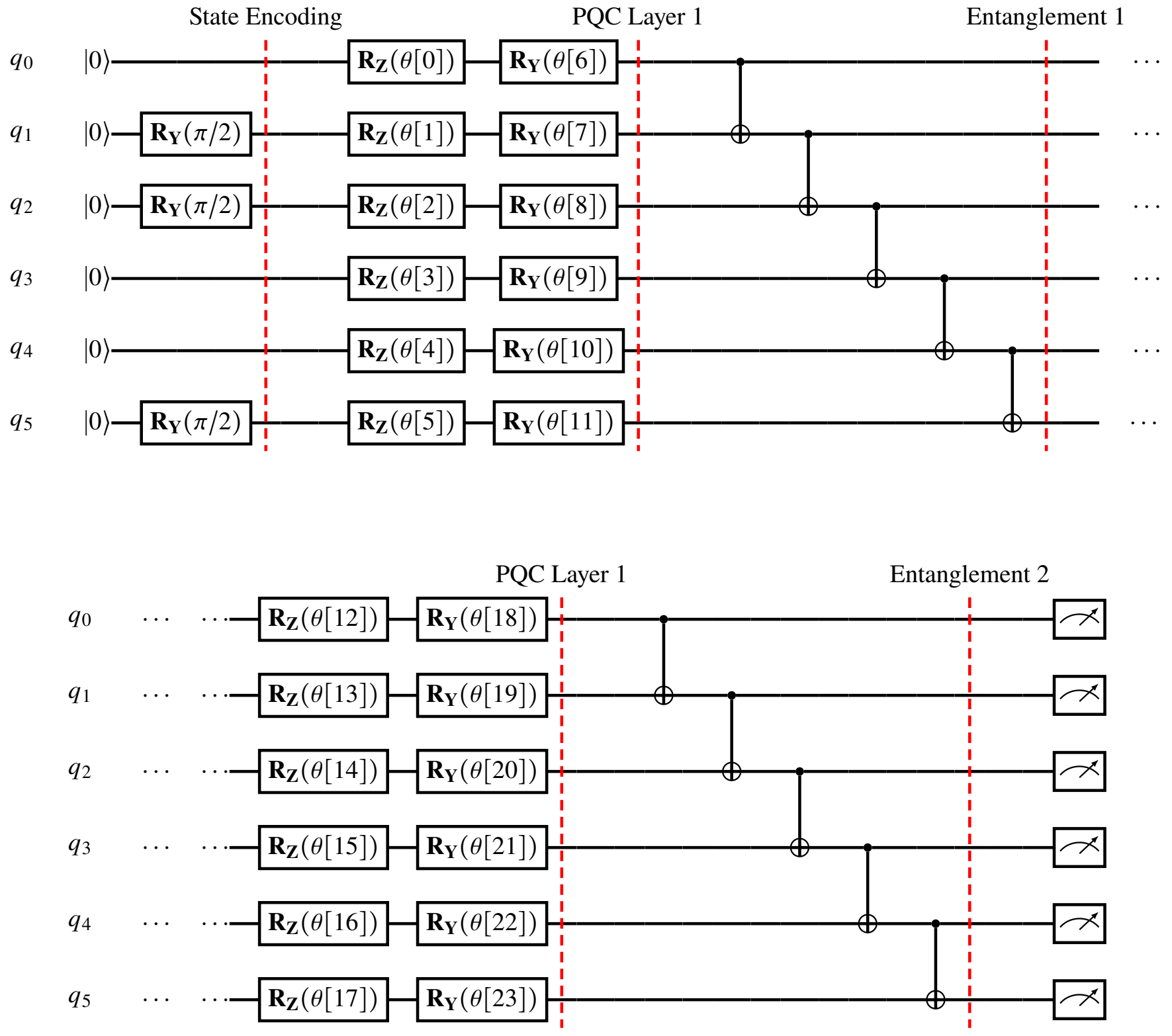
\begin{figure}[H]
\begin{center}\vspace{0.3cm}
    \begin{tikzpicture}
        \node[scale=0.6] 
        {
            \begin{quantikz}
                	\lstick{$q_0$} & \ket{0} &  \hphantomgate{}\slice{State Encoding}     & \hphantomgate{}  &\gate{\textbf{R}_\textbf{Z}(\theta[0])} &\gate{\textbf{R}_\textbf{Y}(\theta[6])}\slice{PQC Layer 1} & \hphantomgate{} & \ctrl{1} 		 & \hphantomgate{} & \hphantomgate{} & \hphantomgate{} & \hphantomgate{}\slice{Entanglement 1}  & \hphantomgate{} & \ldots \\
                	\lstick{$q_1$} & \ket{0} &  \gate{\textbf{R}_\textbf{Y}(\pi/2)} 		&  \hphantomgate{}  &\gate{\textbf{R}_\textbf{Z}(\theta[1])} &\gate{\textbf{R}_\textbf{Y}(\theta[7])}                               & \hphantomgate{} & \targ{} 		 & \ctrl{1}                 & \hphantomgate{} & \hphantomgate{} & \hphantomgate{} & \hphantomgate{} & \ldots  \\
                	\lstick{$q_2$} & \ket{0} &  \gate{\textbf{R}_\textbf{Y}(\pi/2)} 		&  \hphantomgate{}  &\gate{\textbf{R}_\textbf{Z}(\theta[2])} &\gate{\textbf{R}_\textbf{Y}(\theta[8])}                               & \hphantomgate{} & \hphantomgate{} & \targ{}                 & \ctrl{1}                 & \hphantomgate{} & \hphantomgate{} & \hphantomgate{} & \ldots \\
                	\lstick{$q_3$} & \ket{0} &  \hphantomgate{}                          		&  \hphantomgate{}  &\gate{\textbf{R}_\textbf{Z}(\theta[3])} &\gate{\textbf{R}_\textbf{Y}(\theta[9])}                               & \hphantomgate{} & \hphantomgate{} & \hphantomgate{} & \targ{}                 & \ctrl{1}                 & \hphantomgate{} & \hphantomgate{} & \ldots \\
	    	\lstick{$q_4$} & \ket{0} &  \hphantomgate{}                          		&  \hphantomgate{}  &\gate{\textbf{R}_\textbf{Z}(\theta[4])} &\gate{\textbf{R}_\textbf{Y}(\theta[10])}                             & \hphantomgate{} & \hphantomgate{} & \hphantomgate{} &\hphantomgate{} & \targ{}                  & \ctrl{1} & \hphantomgate{} & \ldots  \\		
		\lstick{$q_5$} & \ket{0} &  \gate{\textbf{R}_\textbf{Y}(\pi/2)} 		&  \hphantomgate{}  &\gate{\textbf{R}_\textbf{Z}(\theta[5])} &\gate{\textbf{R}_\textbf{Y}(\theta[11])}                             & \hphantomgate{} & \hphantomgate{} & \hphantomgate{} & \hphantomgate{} & \hphantomgate{} & \targ{} & \hphantomgate{} & \ldots	\
	     \end{quantikz}      
        };
    \end{tikzpicture} \\[1cm]
    \begin{tikzpicture}
        \node[scale=0.6] 
        {
            \begin{quantikz}
                	\lstick{$q_0$} & \ldots  & \ldots &\gate{\textbf{R}_\textbf{Z}(\theta[12])} &\gate{\textbf{R}_\textbf{Y}(\theta[18])}\slice{PQC Layer 1} & \hphantomgate{} & \ctrl{1} 		 & \hphantomgate{} & \hphantomgate{} & \hphantomgate{} & \hphantomgate{}\slice{Entanglement 2}  & \hphantomgate{} & \meter{} \\
                	\lstick{$q_1$} & \ldots & \ldots &\gate{\textbf{R}_\textbf{Z}(\theta[13])} &\gate{\textbf{R}_\textbf{Y}(\theta[19])}                               & \hphantomgate{} & \targ{} 		 & \ctrl{1}                 & \hphantomgate{} & \hphantomgate{} & \hphantomgate{} & \hphantomgate{} &\meter{} \\
                	\lstick{$q_2$} & \ldots & \ldots &\gate{\textbf{R}_\textbf{Z}(\theta[14])} &\gate{\textbf{R}_\textbf{Y}(\theta[20])}                               & \hphantomgate{} & \hphantomgate{} & \targ{}                 & \ctrl{1}                 & \hphantomgate{} & \hphantomgate{} & \hphantomgate{} & \meter{}\\
                	\lstick{$q_3$} & \ldots & \ldots &\gate{\textbf{R}_\textbf{Z}(\theta[15])} &\gate{\textbf{R}_\textbf{Y}(\theta[21])}                               & \hphantomgate{} & \hphantomgate{} & \hphantomgate{} & \targ{}                 & \ctrl{1}                 & \hphantomgate{} & \hphantomgate{} & \meter{} \\
	    	\lstick{$q_4$} & \ldots & \ldots &\gate{\textbf{R}_\textbf{Z}(\theta[16])} &\gate{\textbf{R}_\textbf{Y}(\theta[22])}                             & \hphantomgate{} & \hphantomgate{} & \hphantomgate{} &\hphantomgate{} & \targ{}                  & \ctrl{1} & \hphantomgate{} & \meter{}  \\		
		\lstick{$q_5$} & \ldots & \ldots &\gate{\textbf{R}_\textbf{Z}(\theta[17])} &\gate{\textbf{R}_\textbf{Y}(\theta[23])}                             & \hphantomgate{} & \hphantomgate{} & \hphantomgate{} & \hphantomgate{} & \hphantomgate{} & \targ{} & \hphantomgate{} & \meter{}	
            \end{quantikz}      
        };
        
    \end{tikzpicture}
\end{center}
\caption{Example six-qubit instance of the parameterised quantum circuit used by the atomistic pitch agent. The diagram is shown for a single encoded pitch-state index, corresponding to an illustrative sliding window size $k = 1$, in order to keep the circuit readable.}
\label{fig:qrl-pitch-circuit}
\end{figure}

\medskip

After the variational circuit is applied, all qubits are measured in the computational ($Z$) basis \cite{LaPierre2021}. The probability assigned to an action-associated measurement outcome is interpreted as a probability-derived action-value estimate:
$\hat{Q}_{\theta}(s,a)=P_{\theta}(a\mid s)$.

\medskip

Because the number of possible measurement outcomes is $2^{n_q}$ and $\lvert \mathcal{A} \rvert \le 2^{n_q}$  for every agent, the computational basis probabilities provide one-to-one action-associated outputs. A fixed rescaling converts these outputs to Q-value estimates as described in Section \ref{sec:measure-action}, without a trainable classical output layer.

\medskip

Figure~\ref{fig:qrl-pitch-circuit} shows a readable six-qubit instance of the ansatz for the atomistic pitch agent. The same circuit pattern is used for the duration and holistic agents, with the number of qubits adjusted according to the corresponding state encoding.  

\medskip

In the demonstrations in Section \ref{sec:demos}, the musical environment uses a sliding window of size $k=4$ for action application and reward evaluation. The circuit first encodes the discrete pitch state using fixed $R_Y(\pi/2)$ rotations, then applies two trainable PQC layers composed of $R_Z(\theta)$ and $R_Y(\theta)$ rotations. Each variational layer is followed by a nearest-neighbour CNOT entanglement chain. Finally, all qubits are measured, and the resulting measurement probabilities are interpreted as action-value estimates for selecting musical transformation actions.

\section{The Environment}
\label{sec:envs}

\medskip

The Environment defines what the agent acts on and how its actions are evaluated.  It is formalised as a Markov Decision Process (Section \ref{sec:QRL}), with three instantiations: one for the holistic agents and two for the atomistic agents (pitch and duration, respectively), as detailed in Section \ref{sec:agents}. Across all instantiations, the design follows the same principle: the agent processes a short music snippet through a sliding window  $W_n$ and receives a scalar reward that serves as a proxy for musical quality.

\medskip

For each episode, actions ($A$) and rewards ($\mathcal{R}(s, a)$) are computed for the sliding window $W_n$ before it slides one element in the snippet to compute $W_{n+1}$, and so on (Figure \ref{fig:windows}). The size of the window is set to $k=4$ by default. This is configurable, but larger $k$ values widen the musical context at the cost of a proportionally wider input state.  

\begin{figure}[H]
\begin{center}\vspace{0.3cm}
\includegraphics[width=0.36\linewidth]{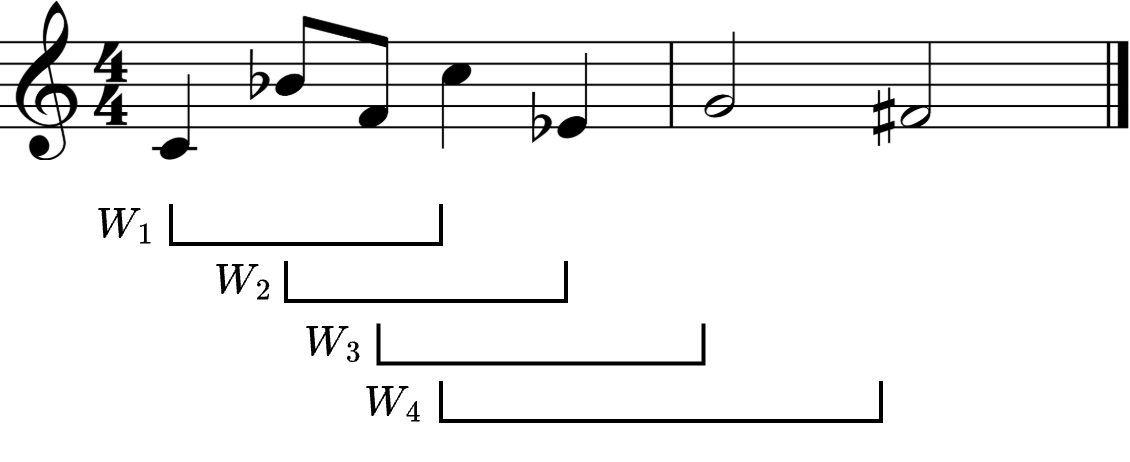}
\caption{A snippet of music is processed with a sliding window. In this example, the size of the window is $k=4$.}
\label{fig:windows}
\end{center}
\end{figure}

\subsection{State Space and Encoding}
\label{sec:state-and-encoding}

Each environment defines a discrete state space $\mathcal{S}$ whose alphabet matches the agent type:  $\mathcal{S}_{\text{pitch}} = \mathcal P$, with $|\mathcal P|=64$ and $\mathcal{S}_{\text{dur}} = \mathcal D$, with $|\mathcal D|=8$ for atomistic agents, and $\mathcal{S}_{\text{events}} = \mathcal E$, with $|\mathcal E|=512$ for holistic ones.

\medskip

A discrete state $s \in \mathbb{N}$ is represented as a fixed-width binary string of length $n_q$ bits.  This string is mapped onto the PQC input as follows: for every bit that is equal to 1, apply a single rotation gate $R_Y(\pi/2)$ to prepare the respective qubits as $\ket{+}$  \cite{LaPierre2021}. Bits that are equal to 0 prepare the respective qubits as $\ket{0}$.  The result is a product state that encodes the binary features in amplitude space with zero additional gate depth.  These encoding rotations act as a data-loading stage that precedes the ansatz $U_\theta$ (Eq. \ref{eq:agent-ansatz}). Once the input state is prepared, $U_\theta$ is applied on top to compute the Q-values.

\medskip

At time-step $t$, the agent does not observe a single state element but rather a sliding window $W$ of the last $k$ consecutive elements of the music snippet.  The state vector is the concatenated binary representation depicted in Eq. \ref{eq:bit-matrix}, where $x_i$ denotes a pitch $p_i$, a duration $d_i$, or a unique event $(p_i, d_i)$, depending on the instantiation.  The agent applies one local transformation (an action $a \in A$) to the current window, the environment updates the global music sequence accordingly, and the reward is recomputed on the whole updated window.

\begin{equation}
s_t=\bigl[\operatorname{bin}(x_{t-k+1}),\dots, \operatorname{bin}(x_t)\bigr]
\label{eq:bit-matrix}
\end{equation}

\subsection{Action Space}

The action space $\mathcal{A}$ defines all admissible musical transformations for pitches, durations and notes (i.e., pitch-duration compounds), respectively. Each action is an operator that modifies the notes in the current window $W$, which are detailed in the Appendix: 

\begin{itemize}
\item{50 operators for pitches: $\mathcal A_\text{pitch}=\{{a^p}_{0}, \dots, {a^p}_{49}\}$}
\item{13 operators for durations: $\mathcal A_\text{dur}=\{{a^d}_{0}, \dots, {a^d}_{12}\}$} 
\item{57 operators for notes: $\mathcal A_\text{events}=\{{a^e}_{0}, \dots, {A^e}_{56}\}$}
\end{itemize}

\medskip

\subsection{Reward Functions}
\label{sec:rewards}

Together with the action operators and the system parameters described in Section \ref{sec:trainloop}, the reward functions are the primary driver of mindset differentiation.  Agents may be trained with any single reward or any weighted combination  $\mathcal R_{\Sigma}$ (Eq.   \ref{eq:total_reward}).

\medskip

Currently, the system supports three reward functions: Diversity ($ \mathcal R_{\text{div}}$), Perfect Balance ($ \mathcal R_{\text{pb}}$) and Smoothness ($ \mathcal R_{\text{sm}}$). These rewards can be applied to pitch sequences, duration sequences, or note sequences. Agents may be trained with any combination of the three reward functions and assign custom weights to compute a total reward $\mathcal R_{\Sigma}$ per timestep, allowing distinct musical mindsets to emerge. (Eq. \ref{eq:total_reward}).

\medskip

Diversity (Eq. \ref{eq:wf}) rewards variety in a symbolic sequence.  It is the ratio of unique elements to total elements in a sequence. For instance, if we have eight notes and six of them are distinct, we score 6/8 = 0.75. If all eight are the same note, we score 1/8 = 0.125. The agent is incentivised to avoid repetition: it gets a higher score the more it spreads its material across different elements rather than repeating the same element.

  \begin{equation}
    \mathcal R_{\text{div}} = \frac{|\text{unique}(\texttt{values})|}{|\texttt{values}|}
    \label{eq:wf}
  \end{equation}

\medskip

Perfect Balance, or \enquote{evenness reward}, encourages even temporal spacing of elements. Elements should be distributed proportionally across the length of the sequence rather than bunched together or unevenly spread. 

\medskip

Equation \ref{eq:pb} maps each element's position in the sequence onto a point on a circle, like plotting hours on a clock face. Then it measures how evenly spread those points are around the full $360\degree$. The $x$ and $y$ components are the sum of the cosines and sines of all those angular positions. If all elements were perfectly evenly spaced around the circle, those sums would cancel out to zero, giving a score of 1 (maximum). If all elements cluster at the same point, the vector is long, and the score drops toward 0. The variable $\theta_i$ denotes the angular position of each element relative to the sequence length.
  
  \begin{equation}
    \mathcal R_{\text{pb}} = 1 - \frac{\sqrt{x^2 + y^2}}{N} , \quad \text{where } x = \sum_i \cos(\theta_i),\ y = \sum_i \sin(\theta_i)
    \label{eq:pb}
  \end{equation}
  
 \medskip
 
The Smoothness reward measures a decrease in distributional change between successive evaluated windows. Let $p_{t}(i)$
denote the normalised frequency of symbol $𝑖$ in the active window after the action at timestep $𝑡$. The implementation compares this histogram with the histogram stored at the preceding evaluated window. It computes their total variation distance and rewards a decrease relative to the previously computed distance, as shown in Eq. \ref{eq:td}.

 \begin{equation}
    \mathcal D_{t} = \frac{1}{2} \sum_i | p_{t}(i) - p_{t-1}(i) | ,  \quad R_{\text{sm}, t} = D_{t - 1} - D_t
    \label{eq:td}
 \end{equation}
 
  \medskip
  
 A positive reward indicates that the distance between successive window distributions has decreased, whereas a negative reward signifies an increase. An unchanged distance produces zero reward, even when the distance itself is large. Consequently, this reward measures a change in local distributional variation rather than an absolute smoothness score for the complete musical sequence.
 
  \medskip

Since the histograms are normalised, $D_t \in [0,1]$ and $R_{\text{sm}, t} \in [-1,1]$, with no additional rescaling or offset applied. Given that both Diversity and Perfect Balance lie in $[0,1]$, an equally weighted three-component reward combining these terms falls within $[- \frac{1}{3}, 1]$.
 
 \medskip
   
If more than one reward is employed, the system computes the total reward as a weighted sum of the selected components (Eq. \ref{eq:total_reward}). By default, the selected reward components are equally weighted, although the weights can be customised. This allows the system to adapt across different agents and enables multi-objective optimisation for evolving distinct mindsets.

 \begin{equation}
  \mathcal R_{\Sigma} = \sum_{x} \lambda_x \mathcal R_x, \quad \text{where  } \lambda_x \text{ are user-defined weights for different reward measures}. 
  \label{eq:total_reward}
 \end{equation}
 
\subsection{Learning Update}
 
 The agent learns with off-policy Q-learning and a uniform replay buffer \cite{fan2020theoretical, kaelbling1996reinforcement, Lin1992, watkins1992qlearning}.  At each time step $t$, the active agent observes a state $s_t$, selects an action $a_t$ via an $\epsilon$-greedy policy, exploiting the highest Q-value with probability $1-\epsilon$, and exploring randomly with probability $\epsilon$. It stores the transition tuple $\langle s_t, a_t, r_t, s_{t+1}, d_t \rangle$ in an agent-specific replay buffer.  A slowly updated target network with frozen parameters $\bar{\theta}$ is used to generate stable Q-value targets.

\medskip
Training proceeds by periodically sampling a mini-batch $B$ of stored transitions and minimising the target-network temporal-difference (TD) loss (Eq.~\ref{eq:double-dqn-loss}).

\medskip

\begin{equation}
\mathcal{L} = \Bigl(r + \gamma \max_{a'} Q_{\bar{\theta}}(s', a') - Q_{\theta}(s, a) \Bigr)^2
\label{eq:double-dqn-loss}
\end{equation}

\medskip

Equation \ref{eq:double-dqn-loss} has the following components:

\begin{itemize}
\item{$r$: the immediate reward received after taking action $a$ in state $s$.}
\item{$\gamma \max_{a'} Q_{\bar{\theta}}(s', a')$: the discounted estimate of the best future reward achievable from next state $s'$, computed using the frozen target parameters $\bar{\theta}$. Using $\bar{\theta}$ instead of the current parameters $\theta$ prevents instability caused by the target \enquote{chasing itself}.}
\item{$Q_{\theta}(s, a)$: the current Q-value predicted by the circuit for the taken action a in state $s$.}
\item{The squared difference, the TD error, is the loss to be minimised: it measures how far the agent's current prediction deviates from the bootstrapped target.}
\end{itemize}

\medskip

The circuit parameters $\theta$ are updated using the parameter-shift rule, which provides an analytic gradient of $\hat{Q}_\theta$ with respect to any single rotation angle $\theta_k$ (Eq. \ref{shift-gradients})
\begin{equation}
\frac{\partial Q_{\theta}(s,a)}{\partial \theta_k}
=
\frac{1}{2}
\left[
Q_{\theta + \frac{\pi}{2}e_k}(s,a)
-
Q_{\theta - \frac{\pi}{2}e_k}(s,a)
\right].
\label{shift-gradients}
\end{equation}

\medskip

Equation \ref{shift-gradients} involves the following quantities:

\begin{itemize}
\item{$\theta_k$: the $k$-th trainable rotation angle in the PQC, that is, one of the $2Ln_q$ angles defined in the ansatz Eq. \ref{eq:agent-ansatz}).}
\item{$e_k$: the unit vector in the direction of $\theta_k$, so that $\theta \pm \frac{\pi}{2}e_k$ means \enquote{shift only angle $\theta_k$ by $\pm \frac{\pi}{2}$ while keeping all other angles fixed}.}
\item{$\hat{Q}_{\theta \pm \frac{\pi}{2}e_k}(s, a)$: the Q-value of action $a$ in state $s$ when the circuit is evaluated with the shifted angle.  Each gradient evaluation, therefore, requires three circuit executions: the current circuit and the two shifted variants.}
\item{The factor ½ is a constant arising from the specific form of quantum rotation gates; it is the quantum analogue of a finite-difference coefficient for sinusoidal functions.}
\end{itemize}

\medskip

Combining Eqs. \ref{eq:double-dqn-loss} and \ref{shift-gradients}, the angle update at each training step follows stochastic gradient descent with learning rate $\eta$ (Eq. \ref{eq:combined}), where $\delta = y − Q_\theta(s, a)$ is the TD error and $g_k = ½(Q^+_{a} − Q^−_{a})$ is the parameter-shift gradient estimate.  

\begin{equation}
\theta_k  \leftarrow  \theta_k  +  2\eta \times \delta \times g_k  
\label{eq:combined}
\end{equation}

\medskip

The exploration coefficient $\epsilon$ decays exponentially as $\epsilon_t = max(0.01, 0.99^t)$, transitioning the agent from pure exploration to increasingly greedy exploitation as training progresses.  The target parameters $\bar{\theta}$ are synchronised with $\theta$ every $\tau = 500$ environment steps.

\section{Unpacking the Learning Process}
\label{sec:trainloop}

This section details how learning unfolds in practice, following the trajectory of a concrete run and exposing the full chain from musical input to quantum measurement and parameter update.

\medskip

Conceptually, the learning process implements the Markov Decision Process defined in Section \ref{sec:QRL} using a hybrid quantum--classical loop. The PQC plays the role of a function approximator for the action–value estimate $\hat{Q}_\theta(s, a)$, while the musical environment provides state transitions and rewards as defined in Section \ref{sec:envs}.

\medskip

The circuit is laid out in four phases: state encoding, variational rotations, entanglement, and measurement (e.g., Figure \ref{fig:qrl-pitch-circuit}).

\subsection{Overview of the Learning Loop}

Each learning episode proceeds as a sequence of timesteps $t=1,\dots,T$, where at every timestep the agent:

\begin{itemize}
\item{Observes a state $s_t$ through a sliding window $W_t$.}
\item{Encodes the state $s_t$ into a quantum register.}
\item{Evaluates action-value scores using its PQC.}
\item{Selects and applies an action $a_t$.}
\item{Receives a scalar reward $r_t$.}
\item{Stores the resulting experience for later learning.}
\end{itemize}

\medskip

Interaction with the environment and parameter optimisation are decoupled. The agent acts sequentially on the musical material, while parameter updates are performed statistically from replayed experience.
 
 \subsection{Learning Timestep}
 
Let us examine a single timestep $t$ in detail. First, consider an atomistic pitch agent with $W$ of size $k = 4$.  At time step $t$, the active window is (Eq. \ref{eq:active-window}):

\begin{equation}
W_t = [p_{t-3}, p_{t-2}, p_{t-1}, p_t] = [C_4, D_4, E_4, G_4]
\label{eq:active-window}
\end{equation}

\medskip

Each pitch $p_i \in S_{\text{pitch}}$ is mapped to its integer index within the 64-element pitch alphabet and encoded as a 6-bit binary string (since $2^6 = 64$). The observable state (Eq. \ref{eq:bit-matrix}) is the concatenation of these encodings, as shown in Eq. \ref{eq:binary-representation}. This binary representation captures the context on which the agent will act.

\begin{equation}
s_t = [\text{bin}(p_{t-3}), \text{bin}(p_{t-2}), \text{bin}(p_{t-1}), \text{bin}(p_t)]
\label{eq:binary-representation}
\end{equation}

\medskip

Next, the binary state $s_t$ is loaded directly into the quantum register. For each qubit in state $\ket{1}$, a gate $R_Y(\pi/2)$ is applied, preparing the corresponding qubit in the superposition state $\ket{+}$; qubits in state $\ket{0}$ are not changed. Thus, the resulting initial quantum state is as shown in Eq. \ref{eq:initial-quantum-state}. At this stage, no trainable parameters are involved: the state purely represents the observed window.

\medskip

Now comes the PQC evaluation. The prepared input state is processed by the agent's ansatz $U_\theta$ (Eq.~\ref{eq:agent-ansatz}), which consists of two repetitions ($L=2$) of parameterised single-qubit rotations $R_Z(\theta)$ and $R_Y(\theta)$ followed by a nearest-neighbour chain of entangling CX gates. The output state is $\ket{\psi_{\text{out}}} = U_\theta \ket{\psi_{\text{in}}(s_t)}$. The measurement distribution of this output state provides the agent's current action-value scores for the available musical transformations.

\medskip

To clarify the mechanism above, Appendix II details a specific parameterised quantum circuit executed by an atomistic pitch agent at a single time step.

\medskip

Considering the atomistic pitch agent above (Eq. \ref{eq:active-window}), each pitch is mapped to its index in the 64-element pitch alphabet and encoded as a 6-bit binary number. The four binary strings are concatenated into a 24-bit state vector. Thus, $n_q = 6 \times 4 = 24$. According to the encoding rule in Section \ref{sec:state-and-encoding}, this process produces the input state as shown in Eq. \ref{eq:initial-quantum-state}.

\begin{equation}
\ket{\psi_{\text{in}}(s_t)} = \bigotimes_{q=0}^{23} \ket{\phi_q}, \quad \ket{\phi_q} \in \{\ket{0}, \ket{+}\}
\label{eq:initial-quantum-state}
\end{equation} 

\medskip

The circuit then applies two identical layers, each consisting of a rotation layer, where for every qubit $q$, $R_Z(\theta_{\ell,q})$ and $R_Y(\theta_{{\ell, n_q}+q})$, and an entangling layer, with the nearest-neighbour chain in Eq.\ref{eq:neighbor-chain}. The total number of trainable parameters is $2Ln_q = 4 \times 24 = 96$.

\begin{equation}
 CX_{0, 1}, CX_{1, 2}, \dots, CX_{22, 23}
\label{eq:neighbor-chain}
\end{equation}

\medskip

At each learning timestep, the agent's current musical context, defined by a sliding window of events, is embedded into the quantum register as a structured quantum state. Binary features associated with the window are mapped onto the computational basis, with active features promoted to superposition via fixed $R_Y(\pi/2)$ rotations. As a result, the observed musical context is represented not as a single configuration but as a quantum state whose amplitudes simultaneously encode multiple potential interpretations. This superpositional representation enables the circuit to evaluate many candidate action tendencies in parallel, exploiting quantum interference as part of the decision‑making process. The subsequent trainable rotations $R_Z(\theta)$ and $R_Y(\theta)$ constitute the agent's learned musical mindset: properties such as smoothness, diversity, or balance are not encoded explicitly as rules, but implicitly, through how amplitudes interfere and redistribute across the state space as learning progresses.

\medskip

Contextual musical relationships are modelled through entanglement. The linear chain of CNOT gates couples adjacent qubits, thereby entangling information associated with different notes within the window. This allows the agent to capture dependencies across the musical context, such that the desirability of an action can depend on patterns spanning multiple positions; for example, favouring a transposition only when earlier pitches form a leap, or avoiding repetition when regularity is already present. 

\medskip

Repeated circuit executions yield a probability distribution over bitstrings, each of which is mapped, via a fixed measurement-to-action mapping, to a discrete musical action. These measurement probabilities define the unscaled probability-derived outputs $\left( \hat{Q}_\theta(s_t, a)=P_\theta(j=M(a)\mid s_t) \right)$, with no intermediate neural network, softmax, or trainable classical output layer; a fixed rescaling is applied before temporal-difference learning, as described in Section \ref{sec:measure-action}. In this sense, action valuation is performed intrinsically by the quantum circuit itself: quantum probabilities are the agent's estimates of action utility.

 \subsection{From Quantum Measurement to Action Selection}
 \label{sec:measure-action}
 
All qubits are measured in the computational basis (Z), yielding bitstrings such as (Eq. \ref{yielded-bitstrings}): 

\begin{equation}
b = b_{n_q-1}\dots b_1 b_0 \in \{0, 1\}^{n_q},
\quad
j = \operatorname{int}(b),
\label{yielded-bitstrings}
\end{equation}
\noindent

where $j$ is the integer index associated with the measured bitstring.

\medskip

Repeating the circuit execution yields probabilities $P(j \mid s_t,\theta)$. Then, for action mapping, the agent defines a fixed measurement-to-action mapping $\mathcal{M}: \mathcal{J}_{\mathcal A} \rightarrow \mathcal A, \quad \mathcal{J}_{\mathcal A} \subseteq \{0,\dots,2^{n_q}-1\}$. This maps measurement outcomes to actions, as shown in the example in Table \ref{tab:fixed-bijection}. 

\begin{table}[h]
\centering
\small
\begin{tabular}{lccc}
\toprule
Bitstring & Index $j$ & Action \\ \midrule
000010  &  2  & \texttt{Shuffle}  \\
000111  &  7 &  \texttt{Singe-transpose +3} \\
001101  & 13  & \texttt{Invert} \\
\bottomrule
\end{tabular}
\caption{Mapping measurement outcomes to actions.}
\label{tab:fixed-bijection}
\end{table}

\medskip

Equation \ref{eq:q-value-action} defines the unscaled probability-derived output associated with each action. No trainable classical
output layer is used.

\begin{equation}
\hat{Q}_\theta(s_t,a) = P(j = \mathcal{M}^{-1}(a) \mid s_t,\theta)
\label{eq:q-value-action}
\end{equation}

\medskip

Before temporal-difference learning, the unscaled output is mapped to the return scale through a fixed affine transformation, as shown in Eq. \ref{eq:affine-transform}.

\begin{equation}
\hat{Q}_{\theta}^\texttt{scaled}(s ,a) = L + (U - L)\hat{Q}_\theta(s, a), \quad L = \frac{\texttt{min}(0, r_{\texttt{min}})}{1 - \gamma}, \quad U = \frac{\texttt{max}(0, r_{\texttt{max}})}{1 - \gamma}
\label{eq:affine-transform}
\end{equation}

\medskip

In this case, $[r_{\texttt{min}}, r_{\texttt{max}}]$ is the theoretical range of the selected reward and $0 \leq \gamma < 0$. These conservative bounds follow from the discounted reward sum and include zero return at termination. They are fixed for each configuration and applied to both main and target outputs, with no additional trainable parameters. For the equally weighted three-component reward, $r \in [- \frac{1}{3}, 1]$ and $\gamma = 0.99$ give $L = - \frac{100}{3}$ and $U = 100$, allowing negative Q-value estimates and values greater than one. Since $U > L$, the transformation preserves the action ranking used by the $\epsilon$-greedy policy.

\medskip

Using an $\epsilon$-greedy policy, the agent \textit{explores} with probability $\epsilon$ or \textit{exploits} with probability $1-\epsilon$. Let us suppose the quantum evaluation yields  $\hat{Q}_\theta = 0.18$ and the action \texttt{Shuffle} is selected (Eq. \ref{eq:yielded-evaluation}): 
 
 \begin{equation}
 \arg\max_a \hat{Q}_\theta(s_t,a) = \text{\texttt{Shuffle}}, \quad \hat{Q}_\theta = 0.18
 \label{eq:yielded-evaluation}
 \end{equation}
 
Thus, the action \texttt{Shuffle} is applied to the current window, resulting in an updated window $W_t'$. The environment evaluates the result with the respective reward function(s)
(Section \ref{sec:rewards}), returns a scalar reward $r_t$, and advances the sliding window to form $s_{t+1}$.

 \subsection{Transition, Replay Buffer and Parameter Updates}
 
Next the transition $\langle s_t, a_t, r_t, s_{t+1}, d_t  \rangle$ is appended to the replay buffer. The replay buffer promotes efficient use of quantum resources. Each parameter update requires multiple circuit executions (parameter-shift rule). Reusing stored experience increases learning yield per quantum run. And it promotes the stabilisation of learning targets. Mixing past and recent experience smooths TD targets and prevents repetitive behaviour. Learning is statistical and asynchronous, rather than immediate.

 \medskip
 
When a batch $B$ is sampled, parameter updates occur. For each sampled transition, a TD target is computed using the frozen target parameters $\bar{\theta}$ and the TD error $\delta$ is evaluated. Then, a single circuit parameter $\theta_k$ is selected, gradients are estimated via the parameter‑shift rule (Eq. \ref{shift-gradients}), and $\theta_k$ is updated using stochastic gradient descent (Eq. \ref{eq:combined}).

 \subsection{System Parameters}

The system parameters are as follows:

\begin{itemize}
	\item{\textbf{Batch size} ($B$): The number of past experiences sampled from memory during each training step (Eqs. \ref{eq:double-dqn-loss} and \ref{shift-gradients}). In the demonstrations below, the system picks four previous experiences to compute the gradient: $B = 4$.}
	 \item{\textbf{Target update} ($\tau$): How often, in terms of episodes, the agent copies its current parameters into the target network. The target network should be updated less frequently than the main network to yield more stable Q-value targets during learning. In the examples below, $\tau = 2$. That is, the target is updated every two episodes. }	 
	\item{\textbf{Exploration coefficient (initial value) $\epsilon$} ($\epsilon_{\text{start}}$): How often the agent explores new actions versus sticking to the ones that it has already tried before. In the demonstrations below, $\epsilon_{\text{start}} = 1.0$ by default. That is, the agent initially takes completely random actions, exploring a wide range of possibilities.}	
	\item{\textbf{Exploration decay $\epsilon$} ($\epsilon_{\text{decay}}$): As the training develops, $\epsilon$ is gradually reduced by multiplying $\epsilon_{\text{start}}  \times \epsilon_{\text{decay}}$. For the demonstrations below, $\epsilon_{\text{decay}} = 0.955$.}
 	\item{\textbf{Number of episodes} ($E$): The number of interactions between the agent and the environment. Unless stated otherwise, $E = 50$ in the examples below.}
	\item{\textbf{Sliding window $W$ size} ($k$): The size of the sliding window in terms of the number of notes. For the demonstrations below, $k = 4$.}
    	\item{\textbf{Steps factor} ($\xi$): The number of passes the agent makes over the snippet within one episode. If ξ = 1, the agent acts once per window position. In the demonstrations below, $\xi = 2$, which allows the agent to act twice before proceeding.}
	\item{\textbf{Discount factor:} ($\gamma$): Weighting of future rewards. Default $\gamma = 0.99$}	
	\item{\textbf{Learning rate:} ($\eta$). Step size for updating circuit parameters. Default: implementation-dependent.}
	\item{\textbf{Rewards}: The set of reward functions to train the agent; $\mathcal R_{\text{div}}$, $\mathcal R_{\text{pb}}$, $\mathcal R_{\text{sm}}$, or a combination thereof. In the demonstrations below, combinations $\mathcal R_{\Sigma}$ are equally weighted.}
	
	\item{\textbf{Actions}: The action operators (i.e., action space $\mathcal A$) for the agent; see Appendix I.}
\end{itemize}

\section{Demonstrations}
\label{sec:demos}

\subsection{Evolving Rhythmic Mindsets Atomistically}
\label{subsec:rhythmic-mindsets}

To evaluate how atomistic duration agents develop rhythmic mindsets, we conduct three experiments. First, we investigate how an agent discovers balanced variants of a given rhythmic snippet. Next, we examine how multi-objective reward functions perform when the available action space is constrained. Finally, we analyse how expanding this rhythmic action space affects the agent's capacity for exploration.

\medskip

All three experiments use the same initial snippet, shown in Figure~\ref{fig:random-rhythmic-snippet}. The agents are trained for 50 episodes across 30 independent random-number-generated seeds for the QRL algorithm. The input and settings remained the same, but the stochastic parts of the algorithm, including parameter initialisation, exploration and action selection, varied between runs. Repeating the experiment with different seeds produces a more reliable footprint of each agent's behaviour than a single run would. The reported curves show the mean across runs, with the shaded region indicating one standard deviation.

\begin{figure}[H]
\begin{center}\vspace{0.3cm}
\includegraphics[width=0.7\linewidth]{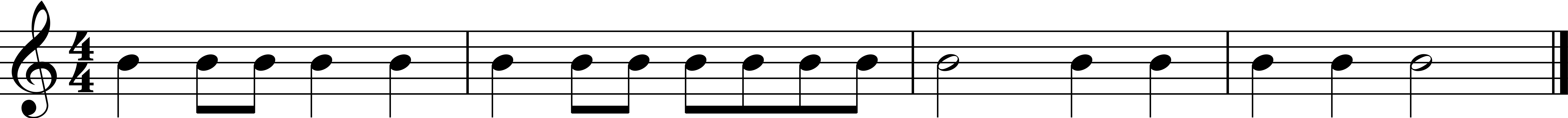}
\caption{Rhythmic snippet used for evolving atomistic rhythmic mindsets.}
\label{fig:random-rhythmic-snippet}
\end{center}
\end{figure}

The main evaluation focuses on the best-so-far score, which records the highest-scoring output discovered at any point during training. This is particularly important in a compositional setting, because the agentic orchestra can preserve, select, or repurpose outputs generated during training rather than using only the final episode. This is consistent with the broader aim of our research: to develop agents as sources of musical behaviours and materials, rather than as optimisers that must always converge monotonically to a single final solution. For completeness, we also report the episode-50 score as a fixed-budget reference, indicating the quality of the output obtained at the end of the training run. Because best-so-far may rise even under random exploration, an ascending curve alone therefore does not establish that a policy has learned. To assess the contribution of training without adding comparisons for every agent, we include a representative numerical comparison for the agent Shin against two non-learning baselines in Section \ref{subsubsec:shin-nonlearning-baselines}
.

\medskip

The three agents are summarised as follows (Table \ref{tab:rhythmic-agent-configs}): Alif is a single-objective agent trained only with the Perfect Balance reward, using a restricted action set. Bob uses the same restricted action set, but is trained with a composite reward combining Diversity, Perfect Balance and Smoothness. Shin is also trained with the composite reward, but is given a more expressive action set that can alter duration values rather than merely reorder them.

\begin{table}[H]
\centering
\caption{Summary of the atomistic rhythmic agents.}
\label{tab:rhythmic-agent-configs}
\begin{tabular}{llll}
\hline
Agent & Reward & Actions & Mindset \\
\hline
Alif & $\mathcal R_{\mathrm{pb}}$ &
$\{\texttt{\footnotesize{Rotate}}, \texttt{\footnotesize{Shuffle}}, \texttt{\footnotesize{Retrograde}}\}$ &
Balanced \\

Bob & $\mathcal R_{\Sigma} =
(\mathcal R_{\mathrm{div}}+\mathcal R_{\mathrm{pb}}+\mathcal R_{\mathrm{sm}}) \div 3$ &
$\{\texttt{\footnotesize{Rotate}}, \texttt{\footnotesize{Shuffle}}, \texttt{\footnotesize{Retrograde}}\}$  &
 Constrained \\

Shin & $\mathcal R_{\Sigma} =
(\mathcal R_{\mathrm{div}}+\mathcal R_{\mathrm{pb}}+\mathcal R_{\mathrm{sm}}) \div 3$ &
$\{\texttt{\footnotesize{Shuffle}}, \texttt{\footnotesize{Elongate}}, \texttt{\footnotesize{Shorten}}\}$  &
Expressive  \\
\hline
\end{tabular}
\end{table}

\subsubsection{Alif: Single-Objective Rhythmic Balance}

The first agent, Alif, was trained to develop a rhythmic mindset based solely on Perfect Balance. Alif's reward function was therefore $\mathcal R = \mathcal R_{\text{pb}}$. Its action set was restricted to simple reordering operations (Eq. \ref{eq:alif-operations}):

\begin{equation}
\mathcal A_{\text{Alif}} = \{\texttt{Rotate}, \texttt{Shuffle}, \texttt{Retrograde}\}.
\label{eq:alif-operations}
\end{equation}

\medskip

These actions do not introduce new duration values. They only reorganise the temporal order of the existing rhythmic material within the sliding window. Alif therefore tests whether it can discover more temporally balanced rhythmic arrangements when the rhythmic vocabulary itself remains fixed.

\medskip

Across 30 seeds, Alif reached a mean best-so-far Perfect Balance score of $\mu_{\text{best}} = 0.9614$, with standard deviation $\sigma_{\text{best}} = 0.0226$.

\medskip

The learning curve is provided in Appendix III, Figure \ref{fig:app-alif-best-so-far}. The curve rises quickly and stabilises near the upper end of the Perfect Balance scale, indicating that Alif can reliably discover rhythmically balanced configurations using only reordering operations. The best-discovered rhythmic output is shown in Figure~\ref{fig:alif-best-output}.

\begin{figure}[H]
\begin{center}\vspace{0.3cm}
\includegraphics[width=0.7\linewidth]{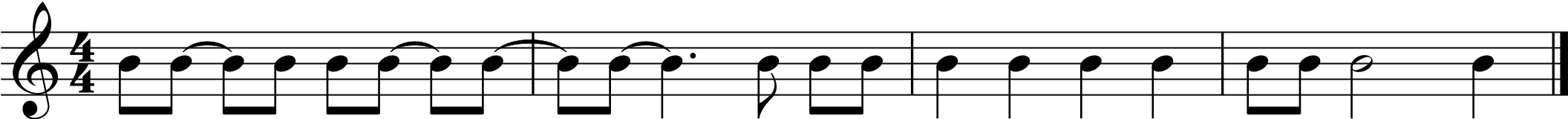}
\caption{Best-discovered output from Alif, with Perfect Balance $\mathcal R_{\mathrm{pb}} = 0.9893$.}
\label{fig:alif-best-output}
\end{center}
\end{figure}

\medskip

For reference, after the fixed training budget of 50 episodes, Alif achieved a mean episode-50 Perfect Balance score of $\mu_{\text{ep50}} = 0.8453$, with standard deviation $\sigma_{\text{ep50}} = 0.0482$. The episode-50 score remained high, but lower than the best-so-far score. This is expected because continued exploration and parameter updates can move the agent away from an earlier high-scoring configuration.

\medskip

Alif therefore demonstrates a focused rhythmic mindset. It is not an expressive duration transformer, because it preserves the original duration vocabulary, but a stable balance-seeking agent. This makes Alif useful in compositional contexts where the rhythmic material should remain recognisable while its temporal organisation is refined.

\subsubsection{Bob: Multi-Objective Learning with a Restricted Action Space}

Bob was trained with the same reordering action set as Alif (Eq. \ref{eq:alif-operations}). However, Bob dealt with a composite reward combining Diversity, Perfect Balance, and Smoothness: $\mathcal R_{\Sigma} = (\mathcal R_{\text{div}} + \mathcal R_{\text{pb}} + \mathcal R_{\text{sm}}) \div 3$.

\medskip

Bob therefore tests whether the same restricted reordering actions can support a broader rhythmic mindset. Unlike Alif, Bob is not rewarded only for temporal balance; it is also rewarded for variety and smoothness. However, because Bob's action set cannot introduce new duration values, its ability to improve Diversity is structurally limited.

\medskip

Across 30 seeds, Bob reached a mean best-so-far composite score of $\mu_{\text{best}} = 0.4504$, with standard deviation $\sigma_{\text{best}} = 0.0130$.

\medskip

The learning curve is provided in Appendix III, Figure \ref{fig:app-bob-best-so-far}. Bob consistently discovered improved rhythmic configurations during training. The relatively small standard deviation also indicates that Bob's best discoveries were stable across seeds, but remained constrained by the fact that its actions can only reorder existing durations. The best-discovered rhythmic output is shown in Figure~\ref{fig:bob-best-output}.

\begin{figure}[H]
\begin{center}\vspace{0.3cm}
\includegraphics[width=0.7\linewidth]{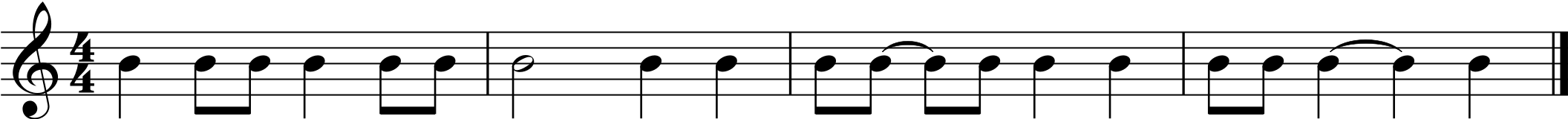}
\caption{Best-discovered output from Bob, with composite reward $\mathcal R_{\Sigma} = 0.4717$.}
\label{fig:bob-best-output}
\end{center}
\end{figure}

\medskip

Inspection of the reward components clarifies the nature of this improvement. Because Bob only reorders existing durations, the Diversity component remains effectively fixed. The first-episode Diversity score is equal to $\mathcal R_{\text{div}} = 0.1667$ and remains so at episode 50. Bob's improvement therefore comes mainly from Perfect Balance and Smoothness rather than from increased rhythmic variety. In the best-so-far outputs, the mean component scores were approximately $\mathcal R_{\text{div}} = 0.1667$, $\mathcal R_{\text{pb}} = 0.9346$, and $\mathcal R_{\text{sm}} = 0.2500$.

\medskip

For reference, after the fixed training budget of 50 episodes, Bob achieved a mean episode-50 composite score of $\mu_{\text{ep50}} = 0.3452$, with standard deviation $\sigma_{\text{ep50}} = 0.0490$.

\medskip

Bob is therefore best understood as a stable but constrained multi-objective rhythmic agent. It can improve the temporal organisation and smoothness of the rhythm, but it cannot substantially increase the diversity of duration values because its action set does not permit such transformations. This illustrates the importance of matching reward design with action-space expressivity.

\subsubsection{Shin: Multi-Objective Learning with Expressive Duration Actions}

The third duration agent, Shin, was trained with the same composite reward as Bob: $\mathcal R_{\Sigma} = \frac{1}{3}(\mathcal R_{\text{div}} + \mathcal R_{\text{pb}} + \mathcal R_{\text{sm}})$. However, Shin's action set was more expressive (Eq. \ref{eq:shin-operations}):

\begin{equation}
\mathcal A_{\text{Shin}} = \{\texttt{Shuffle}, \texttt{Elongate}[\musSixteenth,\musEighth,\musQuarter,\musHalf], \texttt{Shorten}[\musSixteenth,\musEighth,\musQuarter,\musHalf]\}.
\label{eq:shin-operations}
\end{equation}

\medskip

Unlike Alif and Bob, Shin can change the duration values themselves. This allows the agent not only to reorganise the rhythmic material, but also to transform the rhythmic vocabulary. Shin therefore tests whether a richer duration action space enables stronger multi-objective rhythmic discovery.

\medskip

Across 30 seeds, Shin reached a mean best-so-far composite score of $\mu_{\text{best}} = 0.6276$, with standard deviation $\sigma_{\text{best}} = 0.0714$.

\medskip

The learning curve is provided in Appendix III, Figure \ref{fig:app-shin-best-so-far}. Shin consistently discovered improved rhythmic configurations during training. The average improvement from the first episode to the best discovered output was approximately 0.3205. This improvement is substantially larger than Bob's, indicating that the richer action space allows Shin to discover more transformative rhythmic variants. The best-discovered rhythmic output is shown in Figure \ref{fig:shin-best-output}.

\begin{figure}[H]
\begin{center}\vspace{0.3cm}
\includegraphics[width=0.7\linewidth]{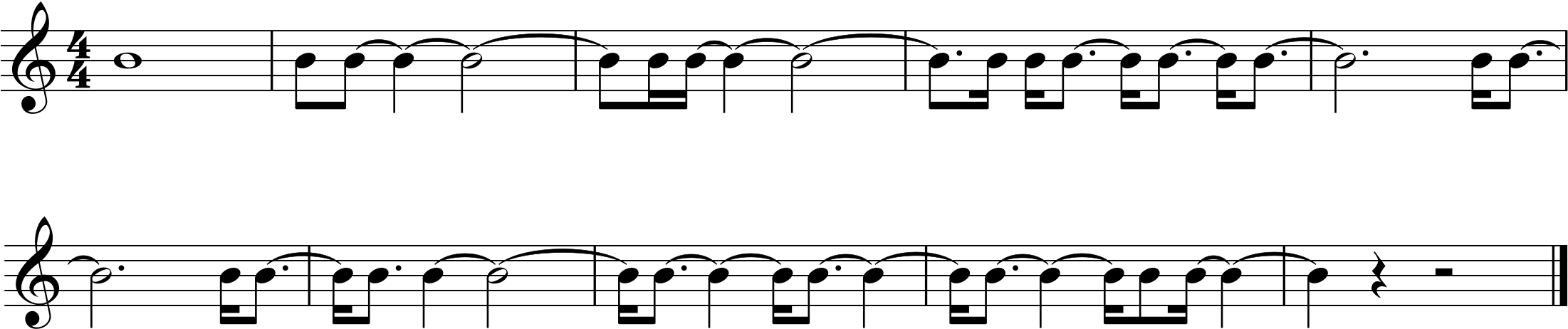}
\caption{Best-discovered output from Shin, with composite reward $\mathcal R_{\Sigma} = 0.7259$.}
\label{fig:shin-best-output}
\end{center}
\end{figure}

Component-level analysis shows that Shin's best outputs were driven mainly by improvements in Perfect Balance and Smoothness. In the best-so-far outputs, the mean component scores were approximately $\mathcal R_{\text{div}} = 0.2222$, $\mathcal R_{\text{pb}} = 0.8440$, and $\mathcal R_{\text{sm}} = 0.8167$.

\medskip

Compared with Bob, Shin achieves a much higher Smoothness score and a stronger overall best-so-far composite score. This shows that the expressive duration-changing actions allow the agent to reshape the rhythmic material in ways that are not available to a reordering-only agent. Shin also improves Diversity slightly, because its action set can alter duration values rather than only rearranging the existing vocabulary.

\medskip

For reference, after the fixed training budget of 50 episodes, Shin achieved a mean episode-50 composite score of $\mu_{\text{ep50}} = 0.3218$, with standard deviation $\sigma_{\text{ep50}} = 0.1519$.

\medskip

The episode-50 scores ranged from $-0.1380$ to $0.6041$, showing considerably higher variability than Bob. In Shin's case, this variability is also related to the expressivity of the action space: the same duration-changing operations that enable stronger discoveries can also move the rhythm into lower-scoring regions.

\medskip

Shin is therefore best understood as an exploratory multi-objective rhythmic agent. Unlike Alif and Bob, it can transform the duration vocabulary itself, which gives it greater potential for rhythmic discovery. Musically, Shin behaves less like a conservative optimiser and more like a transformative rhythmic agent.

\subsubsection{Comparing the Rhythmic Agents}

Figure \ref{fig:rhythm-best-so-far-comparison} compares the best-so-far learning curves for all three rhythmic agents. The figure shows three distinct profiles. Alif rises quickly towards a very high Perfect Balance score, reflecting the strong alignment between its single reward and its restricted reordering actions. Bob improves more modestly because it optimises a composite reward but cannot affect all reward components. Shin improves more substantially than Bob because its duration-changing actions allow it to reshape the rhythmic vocabulary itself.


\begin{figure}[H]
\begin{center}\vspace{0.3cm}
\includegraphics[width=0.7\linewidth]{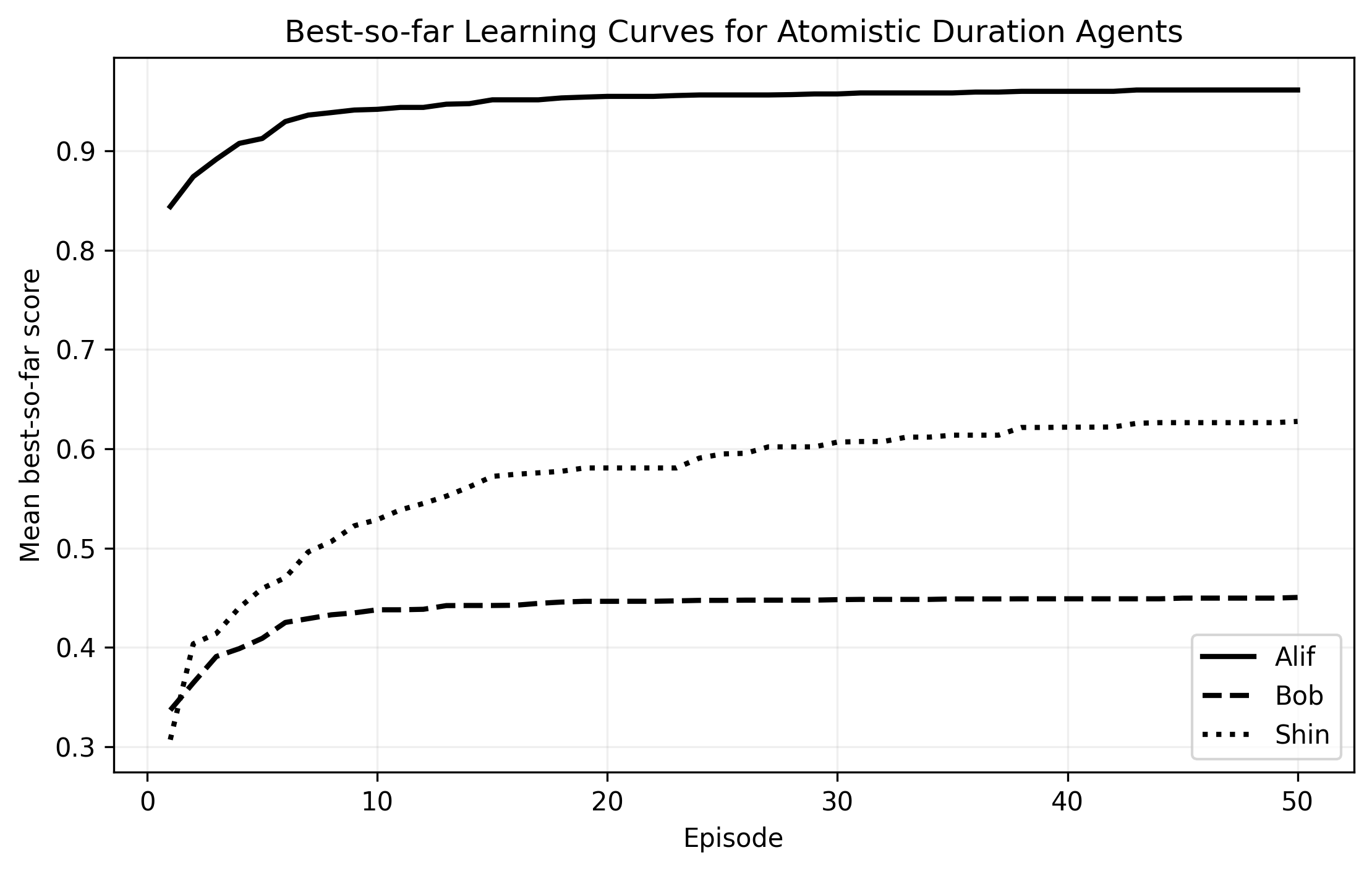}
\caption{Best-so-far learning curves for the three atomistic duration agents across 30 seeds. While Alif is a focused balance-seeking agent, Bob is a constrained multi-objective agent. Shin is a more expressive but less stable multi-objective agent.}
\label{fig:rhythm-best-so-far-comparison}
\end{center}
\end{figure}


Table~\ref{tab:duration-agent-results} summarises the quantitative results. The scores should not be interpreted as directly equivalent across all agents because Alif optimises only Perfect Balance, whereas Bob and Shin optimise the composite reward. The comparison is nevertheless informative because it reveals how different reward-action configurations produce different rhythmic mindsets.

\begin{table}[H]
\centering
\caption{Summary of best-discovered results for the atomistic duration-agent experiments across 30 seeds. While Alif optimises Perfect Balance only, Bob and Shin optimise an equally weighted composite reward $\mathcal R_{\Sigma}$.}
\label{tab:duration-agent-results}
\begin{tabular}{llll}
\hline
Agent & Reward & Action style & Best-so-far score \\
\hline
Alif &
$\mathcal R_{\text{pb}}$ &
Reordering only &
$0.9614 \pm 0.0226$ \\
Bob &
$\mathcal R_{\Sigma}$ &
Reordering only &
$0.4504 \pm 0.0130$ \\
Shin &
$\mathcal R_{\Sigma}$ &
Duration-changing &
$0.6276 \pm 0.0714$ \\
\hline
\end{tabular}
\end{table}

The comparison highlights three important observations. First, a narrow reward-action pairing seems to be highly effective when the reward and available transformations are well aligned. Alif's restricted action set is sufficient for discovering highly balanced rhythmic reorderings. Second, multi-objective rhythmic learning can be limited when the action space cannot affect all reward components. Bob improved the balance and smoothness of the rhythm, but it was not able to substantially improve diversity because it could not create new duration values. Third, expressive action spaces seem to increase the potential for musical discovery, but this can also increase variability. Shin discovered substantially stronger best-so-far rhythmic variants than Bob, but its wider action space also resulted in less predictable behaviour across runs.

\medskip

These results support the interpretation of the agentic orchestra as a collection of agents with distinct rhythmic dispositions rather than a single optimiser. Alif is focused and stable, Bob is balanced but constrained, and Shin is exploratory and transformative. From a compositional standpoint, this diversity is desirable: different agents provide different types of rhythmic material and can be selected, sequenced, or combined depending on the composer's intention.

\subsubsection{Shin: Comparison with Non-Learning Baselines}
\label{subsubsec:shin-nonlearning-baselines}

The best-so-far curves reported above capture the highest-scoring outputs found during training, which helps composition because these outputs may be retained for later use. However, a best-so-far curve is non-decreasing by construction and can also rise when actions are selected at random. To distinguish improvements in musical discovery from repeated exploration alone, we compared Shin with two non-learning baselines: uniform random action selection and an untrained parameterised quantum circuit (PQC) with fixed parameters throughout the experiment. The three conditions used the input rhythmic snippet in Figure \ref{fig:random-rhythmic-snippet}, Shin's action set (Eq. \ref{eq:shin-operations}), composite reward $\mathcal R_{\Sigma} = \frac{1}{3}(\mathcal R_{\text{div}} + \mathcal R_{\text{pb}} + \mathcal R_{\text{sm}})$, and evaluation budget of 50
episodes, with results aggregated over 30 random seeds.

\medskip

As shown in Table \ref{tab:Shin-comparison}, trained QRL achieved a mean best-so-far composite reward of $0.6276 \pm 0.0714$, compared with $0.4132 \pm 0.0989$ for random actions and $0.3790 \pm 0.0122$ for the frozen PQC. These
correspond to absolute differences of 0.2144 and 0.2486, respectively, or approximately $51.9\%$ and $65.6\%$
higher mean best-so-far scores relative to the non-learning baselines. Thus, under the tested conditions, training was associated with the discovery of higher-scoring rhythmic variants than either form of exploration without parameter updates.

\begin{table}[H]
\centering
\caption{Numerical comparison of Shin under QRL training and two non-learning baselines (mean $\pm$ one standard deviation over 30 seeds). Both metrics use the same composite reward $\mathcal R_{\Sigma} = \frac{1}{3}(\mathcal R_{\text{div}} + \mathcal R_{\text{pb}} + \mathcal R_{\text{sm}})$.}
\label{tab:Shin-comparison}
\begin{tabular}{lll}
\hline
Method & Best-so-far (mean $\pm$ SD) & Episode 50 (mean $\pm$ SD) \\
\hline
Trained QRL & $0.6276 \pm 0.0714$ & $0.3218 \pm 0.1519$ \\
Random actions & $0.4132 \pm 0.0989$ & $0.3061 \pm 0.1736$ \\
Frozen PQC & $0.3790 \pm 0.0122$ & $0.2759 \pm 0.0523$ \\
\hline
\end{tabular}
\end{table}

\medskip

The episode-50 scores were substantially closer: $0.3218 \pm 0.1519$ for trained QRL, $0.3061 \pm 0.1736$ for random actions, and $0.2759 \pm 0.0523$ for the frozen PQC. This distinction is important: the control experiment supports the value of learning for finding high-scoring variants during a fixed search budget, not a claim of strong or consistent superiority for the final-episode output.

\subsubsection{Generalisation to Unseen Rhythms}
\label{subsubsec:rhythm-unseen-placeholder}

To evaluate whether the learned rhythmic mindsets transfer beyond the training snippet, the trained duration agents were exposed to an unseen rhythmic sequence, shown in Figure~\ref{fig:unseen-rhythm}. This stage was not a continuation of training. No additional episodes were run, no replay buffer was used, and the agent parameters were not updated. Instead, following the training results reported above, we used the highest-scoring trained model from each rhythmic experiment and applied it to the unseen input in a single post-training pass.

\medskip

This evaluation tests whether the learned mindset can function as a generalisable musical disposition rather than as a response tied only to the training snippet. In a larger-scale version of the system, agents would ideally be trained on multiple snippets so that they encounter a wider range of rhythmic situations and action consequences. Here, however, each model was trained on only one short input. The purpose of this test is therefore deliberately modest: to examine whether a mindset evolved from a single snippet can still produce meaningful transformation when applied to new material.

\begin{figure}[H]
\begin{center}\vspace{0.3cm}
\includegraphics[width=0.7\linewidth]{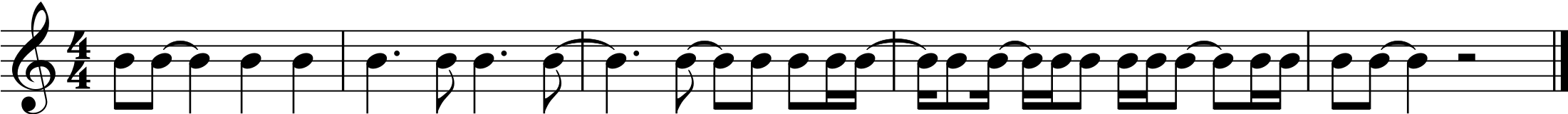}
\caption{Unseen rhythmic sequence for evaluating post-training rhythmic behaviour.}
\label{fig:unseen-rhythm}
\end{center}
\end{figure}

The unseen rhythmic input had a Perfect Balance score of $\mathcal R_{\mathrm{pb}} = 0.6903$. For the multi-objective rhythmic agents, the same unseen input had a composite score of $\mathcal R_{\Sigma} = 0.2883$. These two values are reported separately because Alif was trained and evaluated as a single-objective Perfect Balance agent, whereas Bob and Shin were trained and evaluated with the composite reward.

\medskip

To begin with, we set Alif to apply its mindset to process the unseen rhythm. The resulting output is shown in Figure~\ref{fig:alif-unseen-result}. Alif increased the Perfect Balance score from $\mathcal R_{\mathrm{pb}} = 0.6903$ to $\mathcal R_{\mathrm{pb}} = 0.9517$. 

\begin{figure}[H]
\begin{center}\vspace{0.3cm}
\includegraphics[width=0.7\linewidth]{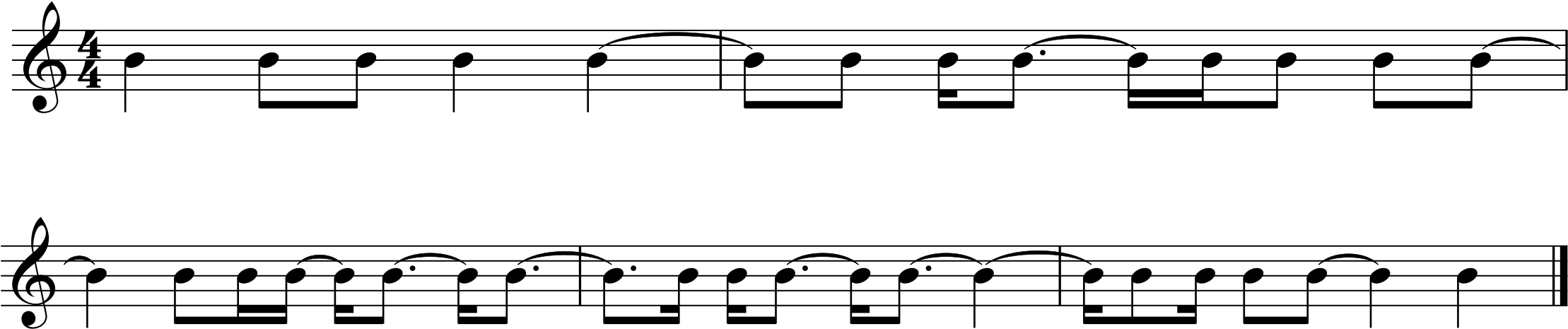}
\caption{Output from Alif acting on the unseen rhythmic sequence in Figure \ref{fig:unseen-rhythm}. The Perfect Balance score increased from $\mathcal R_{\mathrm{pb}} = 0.6903$ to $\mathcal R_{\mathrm{pb}} = 0.9517$.}
\label{fig:alif-unseen-result}
\end{center}
\end{figure}

Alif's result is consistent with its learned rhythmic mindset. As its action space is restricted to reordering operations, it does not introduce new duration values. Instead, it reorganises the existing rhythmic material to improve its temporal balance. The unseen-material result therefore suggests that Alif's balance-seeking behaviour is not specific to the training snippet, but can transfer to a new rhythmic sequence when the learned reordering strategy is appropriate.

\medskip

Next, we set Bob to apply its mindset to process the same unseen rhythm. The output is shown in Figure~\ref{fig:bob-unseen-result}. Bob increased the composite score from $\mathcal R_{\Sigma} = 0.2883$ to $\mathcal R_{\Sigma} = 0.3680$.

\medskip

Because Bob uses the same restricted reordering actions as Alif, its ability to transform the unseen rhythm remains conservative. It cannot substantially change the duration vocabulary, so its improvements must come mainly from reorganising the temporal placement of the existing values. The increase in composite score therefore indicates that Bob's constrained multi-objective mindset can transfer to unseen material, but in a relatively modest way.

\begin{figure}[H]
\begin{center}\vspace{0.3cm}
\includegraphics[width=0.7\linewidth]{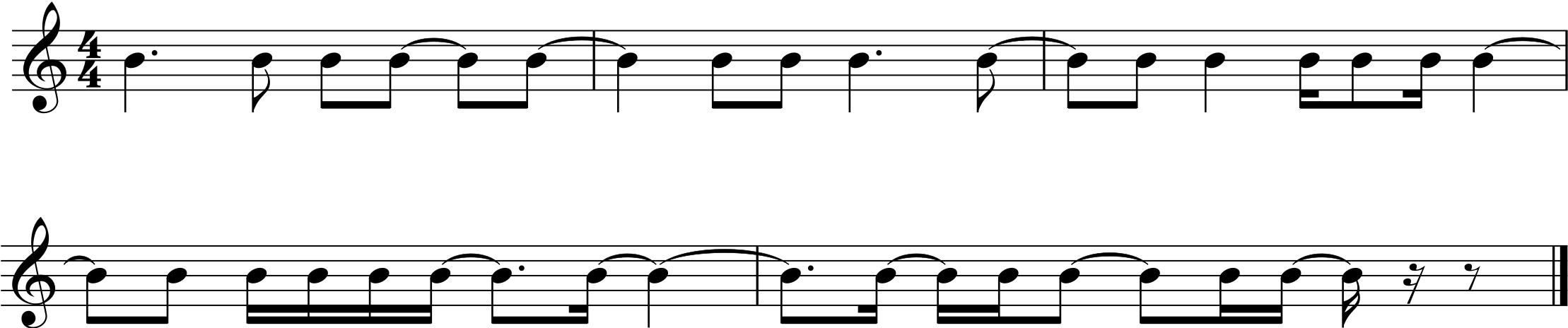}
\caption{Output from Bob acting on the unseen material shown in Figure~\ref{fig:unseen-rhythm}. The composite score increased from$\mathcal R_{\Sigma} = 0.2883$ to $\mathcal R_{\Sigma} = 0.3680$}.
\label{fig:bob-unseen-result}
\end{center}
\end{figure}

Then, it was Shin's turn to process the unseen material. The resulting output is shown in Figure~\ref{fig:shin-unseen-result}. Shin increased the composite score from $\mathcal R_{\Sigma} = 0.2883$ to $\mathcal R_{\Sigma} = 0.3906$.

\medskip

Unlike Alif and Bob, Shin can change duration values as well as reorder them. This gives it greater expressive capacity when acting on unseen material. The result suggests that Shin's learned mindset transfers as a more transformative rhythmic behaviour: it is not limited to rearranging the original rhythm, but can reshape the rhythmic vocabulary itself in pursuit of a stronger multi-objective score.

\begin{figure}[H]
\begin{center}\vspace{0.3cm}
\includegraphics[width=0.7\linewidth]{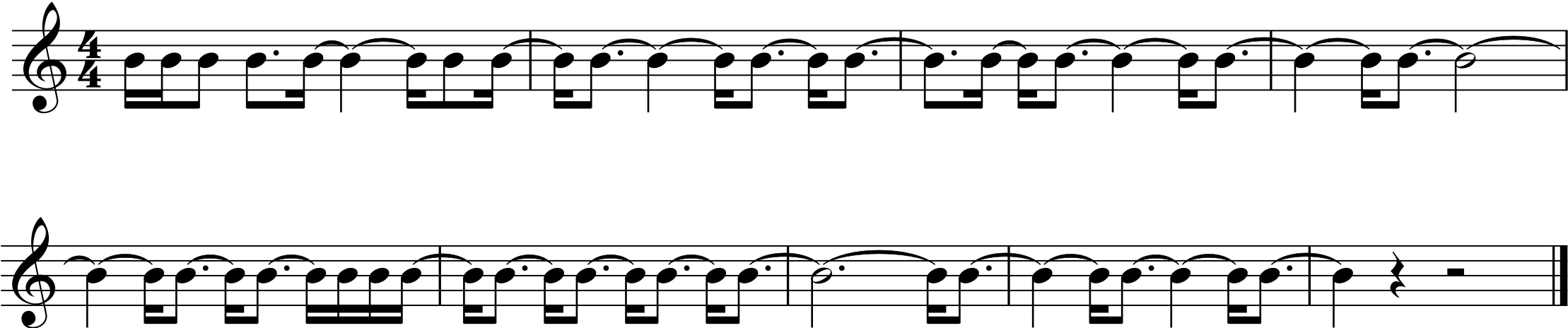}
\caption{Output from Shin acting on the unseen rhythmic sequence in Figure \ref{fig:unseen-rhythm}. The composite score increased from $\mathcal R_{\Sigma} = 0.2883$ to $\mathcal R_{\Sigma} = 0.3906$}.
\label{fig:shin-unseen-result}
\end{center}
\end{figure}

Taken together, these results show that the rhythmic agents learned reusable behaviours rather than merely producing one-off transformations of the training snippet. Alif transfers as a focused balance-seeking agent, Bob transfers as a conservative multi-objective reordering agent, and Shin transfers as a more expressive rhythmic transformer. Although stronger generalisation would likely require training on a broader collection of rhythmic inputs, these results show that even single-snippet training can produce musical behaviours that remain useful when applied to unseen material.


\subsection{Evolving Pitch Mindsets Atomistically}
\label{subsec:pitch-mindsets}

Next, let us examine how atomistic pitch agents evolve melodic mindsets under different reward and action-space configurations. First, we test whether an agent can discover more diverse melodic variants of a given pitch sequence. Second, we examine whether a controlled multi-objective pitch-action set can support a broader melodic mindset shaped by Diversity, Perfect Balance and Smoothness.

\medskip

Both pitch experiments use the pitch sequence shown in Figure~\ref{fig:pitch-seed}. The agents are trained for 50 episodes across 30 independent random seeds. As in the rhythmic experiments, multiple seeds are used because the QRL process contains stochastic elements, including parameter initialisation, exploration and action selection. The reported curves show the mean across runs, with the shaded region indicating one standard deviation.

\medskip

The main evaluation focuses on the best-so-far score, which records the highest-scoring output discovered at any point during training. This is especially relevant for the agentic orchestra because musically useful material may emerge before the final episode and can later be selected or repurposed. For completeness, we also report the episode-50 score as a fixed-budget reference.

\medskip

The pitch experiments were conducted with two agents. While Nava is a single-objective pitch-diversity agent trained only with $\mathcal R_{\text{div}}$, Min is a multi-objective pitch agent trained with a composite reward.

\begin{figure}[H]
\begin{center}\vspace{0.3cm}
\includegraphics[width=0.7\linewidth]{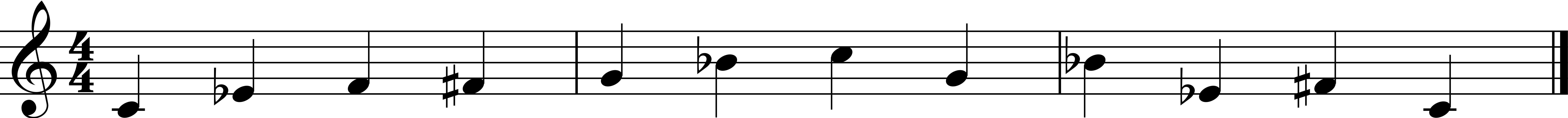}
\caption{Pitch sequence used for evolving atomistic pitch mindsets.}
\label{fig:pitch-seed}
\end{center}
\end{figure}

\subsubsection{Nava: Single-Objective Pitch Diversity}

The first pitch experiment was designed to study how Nava can develop a focused pitch-diversity mindset. Nava was trained only with a Diversity reward $\mathcal R = \mathcal R_{\text{div}}$. The pitch action set contained mutation and transposition actions, as follows (Eq. \ref{eq:nava-operations} ): 

\begin{equation}
\mathcal A_{\text{Nava}} = \{\texttt{Mutate}, \texttt{UniformTranspose}, \texttt{UnevenTranspose}\}.
\label{eq:nava-operations}
\end{equation}

\medskip

More specifically, the transposition actions included both uniform and uneven shifts across a range of intervals. This action set is well aligned with the Diversity objective because mutation and transposition can directly increase the number of distinct pitch states in the melodic material.

\medskip

Across 30 seeds, Nava reached a mean best-so-far Diversity score of $\mu_{\text{best}} = 0.9972$, with standard deviation $\sigma_{\text{best}} = 0.0152$.

\medskip

The learning curve is provided in Appendix III, Figure \ref{fig:app-nava-best-so-far}. The curve rises rapidly and approaches the maximum possible Diversity score, indicating that Nava can reliably discover highly varied pitch configurations. The average improvement from the first episode to the best discovered output was approximately 0.1389. The best-discovered melodic output is shown in Figure~\ref{fig:nava-best-output}.

\begin{figure}[H]
\begin{center}\vspace{0.3cm}
\includegraphics[width=0.7\linewidth]{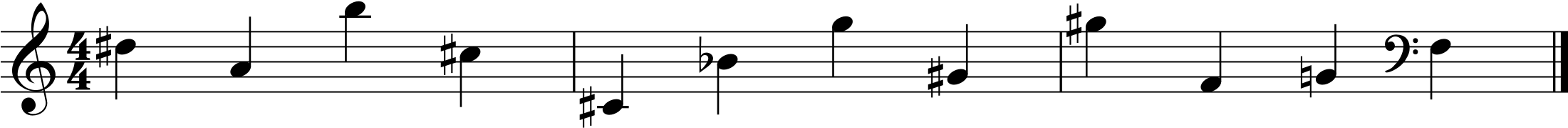}
\caption{Best-discovered output from Nava with Diversity $\mathcal R_{\mathrm{div}} = 1.0000$.}
\label{fig:nava-best-output}
\end{center}
\end{figure}

For reference, after the fixed training budget of 50 episodes, Nava achieved a mean episode-50 Diversity score of $\mu_{\text{ep50}} = 0.8111$ and $\sigma_{\text{ep50}} = 0.1157$.

\medskip

Nava is therefore best understood as a focused pitch-diversity generator. It is highly effective when the goal is to generate varied melodic material, especially when pitch variety is desired as the primary compositional feature.

\subsubsection{Min: Controlled Multi-Objective Pitch Learning}

The second pitch experiment was designed as the main multi-objective pitch-mindset experiment. Min was trained with an equally weighted composite reward $\mathcal R_{\Sigma}
= \left(\mathcal R_{\text{div}}+\mathcal R_{\text{pb}}+\mathcal R_{\text{sm}}\right) \div 3$.

\medskip

The action set was designed to be more controlled and structurally varied than the single-objective experiment, as follows (Eq. \ref{eq:min-operations}):

\begin{align}
\mathcal A_{\text{Min}} = \{ & \texttt{Invert}, \texttt{Rotate}, \texttt{Mutate}, \texttt{Retrograde}, \nonumber \\
& \texttt{UniformTranspose}, \texttt{UnevenTranspose}, \texttt{Reflect} \}.
\label{eq:min-operations}
\end{align}

\medskip

With this experiment, we tested whether a selected set of structural pitch transformations can support a broader melodic mindset, where the agent is not only encouraged to vary pitch material, but also to improve balance and smoothness.

\medskip

Across 30 seeds, Min reached a mean best-so-far composite score of $\mu_{\text{best}} = 0.5503$, with standard deviation $\sigma_{\text{best}} = 0.0340$.

\medskip

The learning curve is provided in Appendix III, Figure \ref{fig:app-min-best-so-far}. Min's best-so-far score improved reliably, although the composite objective is more difficult than Nava's single-objective Diversity task. The average improvement from the first episode to the best discovered output was approximately equal to $0.2669$. The best-discovered melodic output is shown in Figure~\ref{fig:min-best-output}.

\begin{figure}[H]
\begin{center}\vspace{0.3cm}
\includegraphics[width=0.7\linewidth]{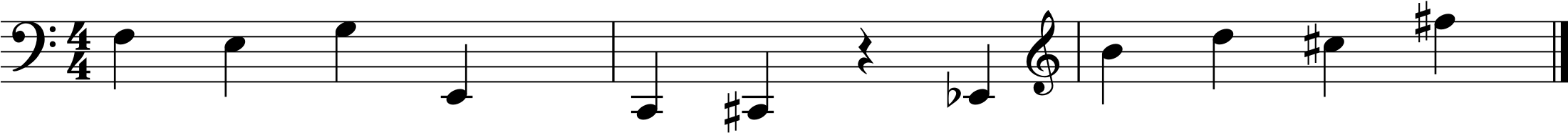}
\caption{Best-discovered output from Min with Composite reward $\mathcal R_{\Sigma} = 0.6159$.}
\label{fig:min-best-output}
\end{center}
\end{figure}

The best-so-far score is lower than in Nava's single-objective Diversity experiment, but this is expected because Min is optimising three criteria simultaneously. Component-level analysis shows that the best-so-far outputs were driven mainly by Diversity and Smoothness: $\mathcal R_{\text{div}} = 0.8833$, $\mathcal R_{\text{pb}} = 0.0341$ and $\mathcal R_{\text{sm}} = 0.7333$.

\medskip

This indicates that Min learned to produce pitch configurations with high variety and relatively strong Smoothness, but Perfect Balance remains low. The low Perfect Balance component suggests that, under the current pitch representation and action set, $\mathcal R_{\text{pb}}$ is harder to optimise than Diversity and Smoothness.

\medskip

For reference, after the fixed training budget of 50 episodes, Min achieved a mean episode-50 composite score of $\mu_{\text{ep50}} = 0.2060$, with standard deviation $\sigma_{\text{ep50}} = 0.1236$.

\medskip

Min is therefore best understood as a controlled multi-objective pitch agent. It does not maximise pitch variety as directly as Nava, but it develops a broader melodic mindset that balances diversity and smoothness while operating under a more demanding composite objective.

\subsubsection{Comparing the Pitch Agents}

Figure \ref{fig:pitch-best-so-far-comparison} compares the best-so-far learning curves across the two pitch experiments. Nava rises quickly and approaches the maximum possible Diversity score, showing that pitch diversity is easy to optimise when the reward and action set are directly aligned. Min improves more gradually and reaches a lower score, as expected from a multi-objective agent optimising three criteria simultaneously.

\begin{figure}[H]
\begin{center}\vspace{0.3cm}
\includegraphics[width=0.7\linewidth]{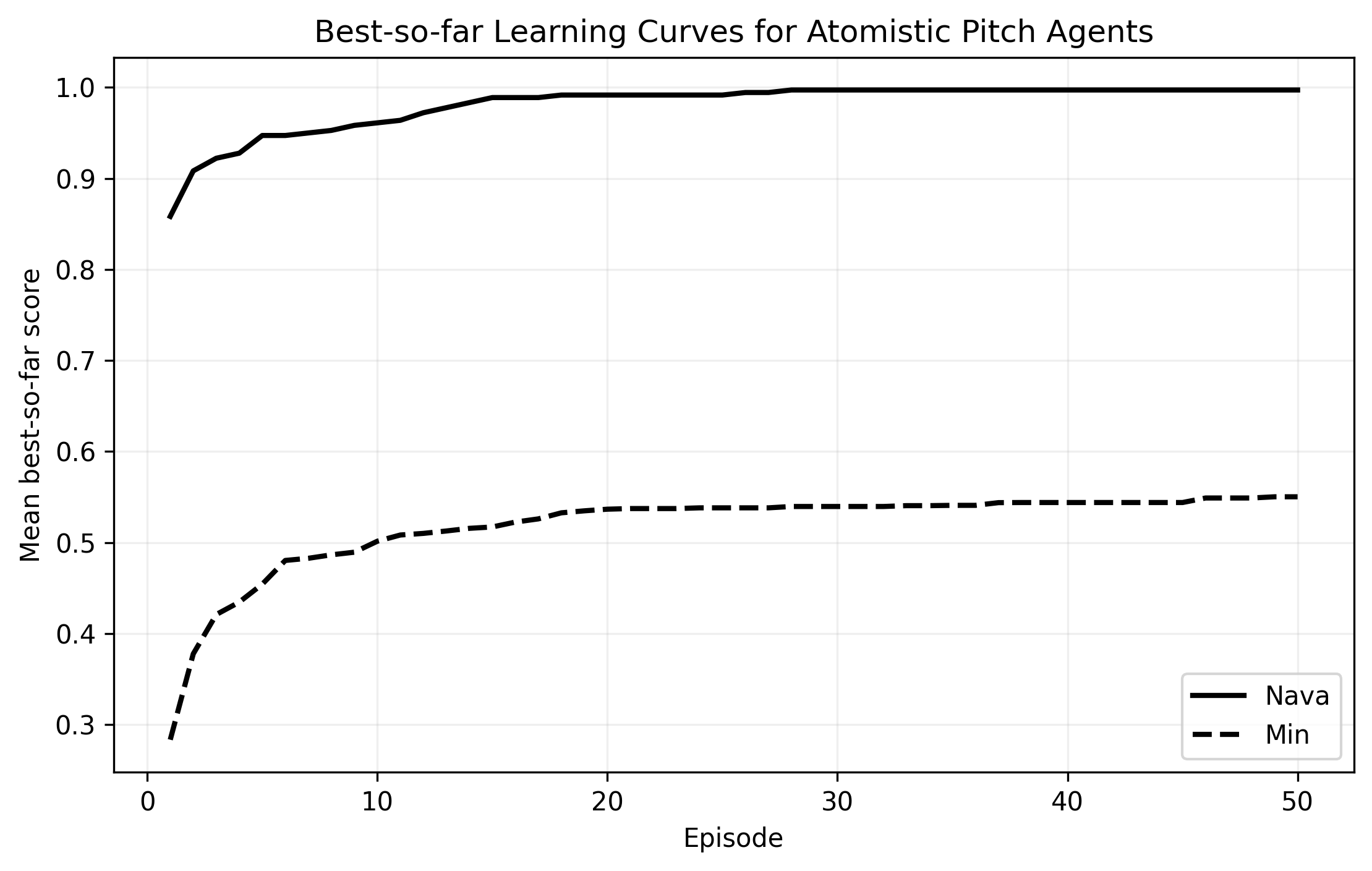}
\caption{Best-so-far learning curves for the atomistic pitch experiments across 30 seeds. Nava rapidly reaches near-maximal pitch variety, while Min provides a broader multi-objective pitch mindset.}
\label{fig:pitch-best-so-far-comparison}
\end{center}
\end{figure}

Table \ref{tab:pitch-agent-results} summarises the best-discovered results. The scores should not be interpreted as directly equivalent because Nava optimises only Diversity, whereas Min optimises the composite reward $\mathcal R_{\Sigma}$. The comparison is nevertheless useful because it shows how reward design and action-space design produce different pitch mindsets.

\begin{table}[H]
\centering
\caption{Summary of best-discovered results for the atomistic pitch-agent experiments across 30 seeds. Nava optimises Diversity only, while Min optimises the equally weighted composite reward $\mathcal R_{\Sigma}$.}
\label{tab:pitch-agent-results}
\begin{tabular}{llll}
\hline
Agent & Reward & Action style & Best-so-far score \\
\hline
Nava &
$\mathcal R_{\text{div}}$ &
Mutation + transposition &
$0.9972 \pm 0.0152$ \\
Min &
$\mathcal R_{\Sigma}$ &
Structural pitch actions &
$0.5503 \pm 0.0340$ \\
\hline
\end{tabular}
\end{table}

The comparison highlights two main observations. First, single-objective pitch diversity can be maximised effectively when the action set directly supports the reward. Nava reaches near-perfect Diversity in almost all runs. Second, multi-objective pitch learning seems to be more difficult, but Min's controlled action set still enables reliable improvement across seeds.

\medskip

Overall, these experiments show that pitch mindsets are shaped strongly by the interaction between reward design and action-space design. A focused Diversity reward produces an exploratory pitch-variety agent, while the controlled composite setting produces a broader but more difficult multi-objective pitch mindset.

\subsubsection{Generalisation to Unseen Pitch Sequences}
\label{subsubsec:pitch-unseen-placeholder}

We also evaluated whether the learned pitch mindsets could be applied to pitch sequences beyond the original training material. As in the rhythmic generalisation test, this was not a continuation of training: no additional episodes were run, no replay buffer was used, and the agent parameters were not updated. The highest-scoring trained model from each pitch experiment was applied to the unseen melodic input in a single post-training pass.

\begin{figure}[H]
\begin{center}\vspace{0.3cm}
\includegraphics[width=0.7\linewidth]{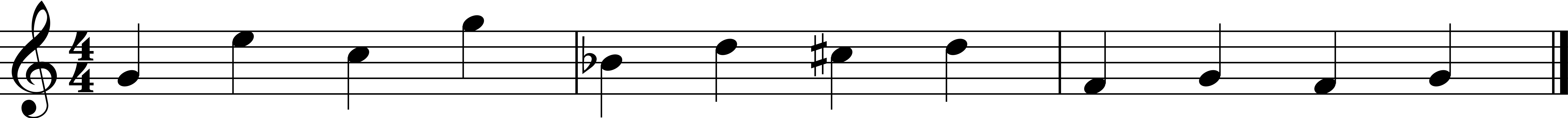}
\caption{Unseen pitch sequence for evaluating post-training pitch behaviour.}
\label{fig:unseen-pitch}
\end{center}
\end{figure}

The unseen input in Figure~\ref{fig:unseen-pitch} had a Diversity score of $\mathcal R_{\mathrm{div}} = 0.6667$. For the multi-objective pitch evaluation, the same input had a composite score of $\mathcal R_{\Sigma} = 0.2137$. These values are reported separately because Nava was trained and evaluated as a single-objective pitch-diversity agent, whereas Min was trained and evaluated with the composite reward.

\medskip

First, Nava applied its mindset to the unseen pitch sequence. The resulting output is shown in Figure~\ref{fig:nava-unseen-result}. Nava increased the Diversity score from $\mathcal R_{\mathrm{div}} = 0.6667$ to $\mathcal R_{\mathrm{div}} = 1.0000$. This result is consistent with Nava's learned mindset. Since Nava was trained specifically to increase pitch diversity, its output produces a sequence in which every pitch event is distinct according to the pitch alphabet of the environment. The result suggests that Nava's diversity-seeking behaviour is not specific to the training snippet, but can transfer directly to a new, unseen input.

\begin{figure}[H]
\begin{center}\vspace{0.3cm}
\includegraphics[width=0.7\linewidth]{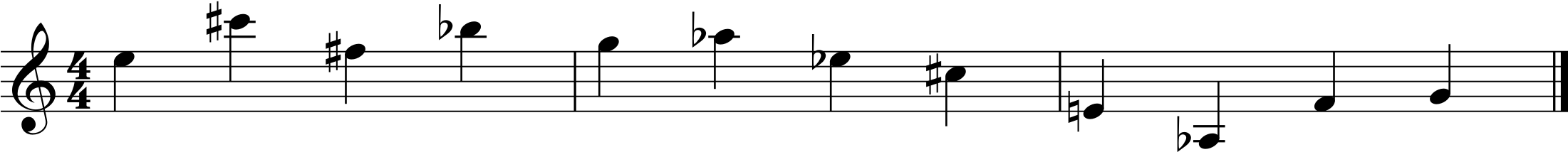}
\caption{Output from Nava acting on the unseen melodic sequence in Figure~\ref{fig:unseen-pitch}. The Diversity score increased from $\mathcal R_{\mathrm{div}} = 0.6667$ to $\mathcal R_{\mathrm{div}} = 1.0000$.}
\label{fig:nava-unseen-result}
\end{center}
\end{figure}

Then, it was Min's turn to process the unseen pitch sequence. The output is shown in Figure~\ref{fig:min-unseen-result}. Min increased the composite score from $\mathcal R_{\Sigma} = 0.2137$ to $\mathcal R_{\Sigma} = 0.299$.

\medskip

Because Min was trained with a composite reward, its behaviour is not directed only toward pitch diversity. Instead, it attempts to balance diversity with Perfect Balance and Smoothness. The improvement therefore indicates that Min's learned multi-objective mindset can transfer to unseen melodic material, producing a stronger overall configuration without optimising only one component.

\begin{figure}[H]
\begin{center}\vspace{0.3cm}
\includegraphics[width=0.7\linewidth]{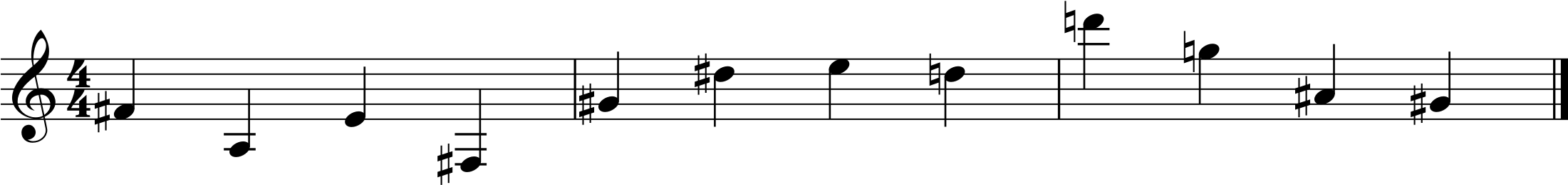}
\caption{Output from Min acting on the unseen melodic sequence in Figure~\ref{fig:unseen-pitch}. The composite score increased from $\mathcal R_{\Sigma} = 0.2137$ to $\mathcal R_{\Sigma} = 0.2991$.}
\label{fig:min-unseen-result}
\end{center}
\end{figure}

Taken together, these results show that the pitch agents also learned reusable behaviours rather than merely transforming the training snippet. Nava transfers as a focused diversity-seeking agent, producing a maximal improvement on its single target measure. Min transfers as a controlled multi-objective melodic agent, improving the overall composite score without focusing exclusively on diversity. Although more robust generalisation would likely require training on a broader collection of melodic inputs, these results show that even single-snippet training can produce pitch transformation behaviours that remain useful when applied to new material.


\subsection{Evolving Musical Mindsets Holistically}
\label{subsec:notes-mindsets}

Let us now examine how two holistic agents, Qasim and Freyja, evolve musical mindsets. In this case, the agents treat each note as a unified pitch-duration event, encoded as a single symbolic object drawn from the alphabet $\mathcal E = \mathcal P \times \mathcal D$. From here onwards we refer to elements of $\mathcal E$ simply as a musical \textit{note}. A transformation may therefore change the identity of a note, allowing pitch and rhythmic behaviour to be shaped jointly.

\medskip

We begin by testing Qasim, which uses a comparatively open action set containing event-level reordering, pitch transformation, duration transformation, silence insertion and pause replacement operations. Then, we examine how Freyja behaves with a more controlled action set designed to reduce unproductive exploration while retaining enough expressive capacity to transform the musical material. 

\medskip

Both agents used the same musical passage, shown in Figure~\ref{fig:HolisticSeed}. In both instances, Qasim and Freyja were trained for 50 episodes across 30 independent random seeds. As in the previous experiments, the main evaluation focuses on the best-so-far score, while the episode-50 score is reported as a fixed-budget reference.

\medskip

Both agents optimise the following equally weighted reward: $\mathcal R_{\Sigma} =  \left( \mathcal R_{\mathrm{div}} + \mathcal R_{\mathrm{pb}} + \mathcal R_{\mathrm{sm}} \right) \div 3$.

\begin{figure}[H]
\begin{center}\vspace{0.3cm}
\includegraphics[width=0.7\linewidth]{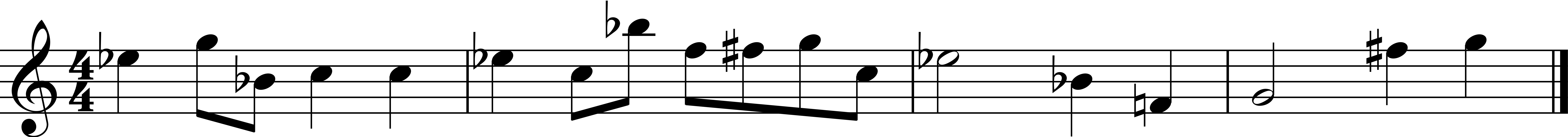}
\caption{Musical passage used for evolving holistic mindsets.}
\label{fig:HolisticSeed}
\end{center}
\end{figure}

\begin{table}[H]
\centering
\caption{Both holistic agents optimise the equally weighted composite reward $\mathcal R_{\Sigma}$.}
\label{tab:holistic-agents-summary}
\begin{tabular}{llll}
\hline
Agent & Reward & Action style & Main purpose \\
\hline
Qasim & $\mathcal R_{\Sigma}$ & Open holistic actions & Exploratory transformation \\
Freyja & $\mathcal R_{\Sigma}$ & Controlled holistic actions & Stable holistic discovery \\
\hline
\end{tabular}
\end{table}

\subsubsection{Qasim: Open Holistic Event Learning}

Qasim was trained using the equally weighted reward $\mathcal R_{\Sigma} =  \left( \mathcal R_{\mathrm{div}} + \mathcal R_{\mathrm{pb}} + \mathcal R_{\mathrm{sm}} \right) \div 3$, and a rich action set, as follows (Eq. \ref{eq:qasim-operations}):

\begin{align}
\mathcal A_{\text{Qasim}} = \{ & \texttt{Shuffle}, \texttt{Mutate}, \texttt{Invert}, \texttt{Reflect[-7, -5, +5, +7]}, \nonumber \\
&  \texttt{Silence}, \texttt{Unsilence}, \texttt{Elongate[0.5, 1.0]}, \texttt{Shorten[0.5, 1.0]}, \nonumber \\
&  \texttt{UniformTranspose[-4, -3, -2, -1, +1, +2, +3, +4]} \}
\label{eq:qasim-operations}
\end{align}

\medskip

Across 30 seeds, Qasim reached a mean best-so-far composite score equal to $\mu_{\mathrm{best}} = 0.7814$, with standard deviation $\sigma_{\mathrm{best}} = 0.0635$.

\medskip

The learning curve is provided in Appendix III, Figure \ref{fig:app-qasim-best-so-far}. Qasim's best-so-far score improves strongly during training, showing that the open action set can discover effective pitch-duration event transformations. The average improvement between the first episode and the best discovered output was approximately 0.3748. The best-discovered holistic output is shown in Figure \ref{fig:qasim-best-output}.

\begin{figure}[H]
\begin{center}\vspace{0.3cm}
\includegraphics[width=0.7\linewidth]{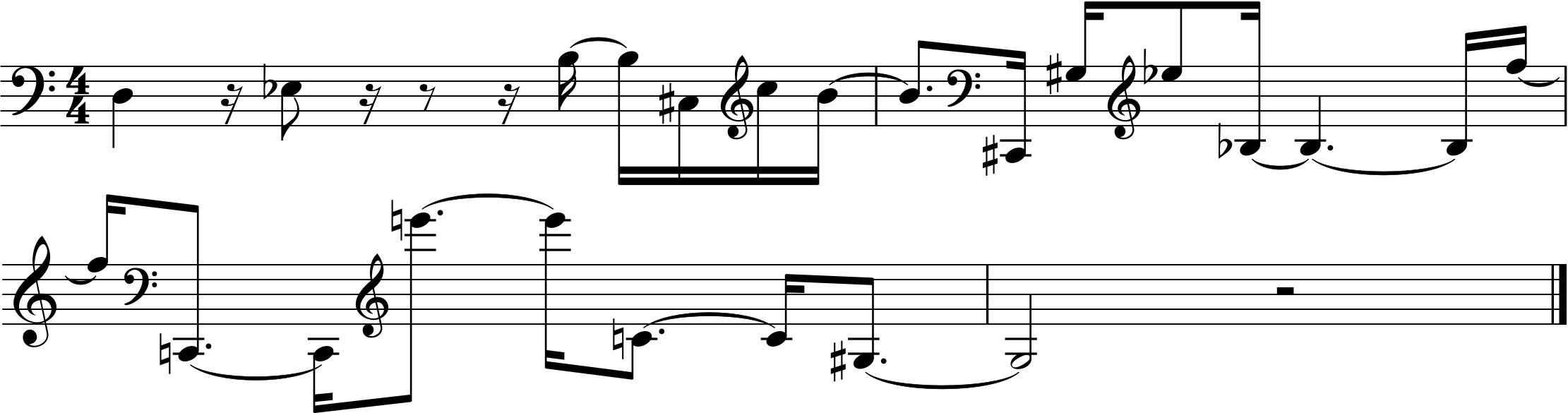}
\caption{Qasim's best-discovered output with the composite reward $\mathcal R_{\Sigma} = 0.8857$.}
\label{fig:qasim-best-output}
\end{center}
\end{figure}

Component-level analysis clarifies the nature of the discovered configurations. In the best-so-far outputs, the mean component scores were approximately $
\mathcal R_{\mathrm{div}} = 0.9241$, $\mathcal R_{\mathrm{pb}} = 0.7034$ and $\mathcal R_{\mathrm{sm}} = 0.7167$.

\medskip

The high Diversity score shows that Qasim has effectively expanded the vocabulary. Nevertheless, the Balance and Smoothness scores are also substantial, indicating that the improvement is not achieved merely by increasing event variety. Rather, the agent discovers note transformations that jointly improve diversity, balance and local continuity.

\medskip

For reference, after the fixed training budget of 50 episodes, Qasim achieved a mean episode-50 composite score of $\mu_{\mathrm{ep50}} = 0.4307$, with standard deviation $\sigma_{\mathrm{ep50}} = 0.2006$.

\medskip

Qasim is therefore best understood as an exploratory holistic agent: it does not simply optimise pitch or duration, but transforms complete note identities.

\subsubsection{Freyja: Controlled Holistic Event Learning}

Freyja was trained with the same musical passage (Figure \ref{fig:HolisticSeed}) and an equally weighted reward as Qasim, but with a more restrained action set (Eq. \ref{eq:freyja-operations}) designed to test whether constraining the available transformations could improve learning stability in the holistic event space.

\begin{align}
\mathcal A_{\text{Freyja}} = \{ & \texttt{Retrograde}, \texttt{Shuffle}, \texttt{Rotate}, \texttt{Mutate}, \texttt{Invert}, \nonumber \\
&  \texttt{Silence}, \texttt{Unsilence}, \texttt{Reflect[-6, -5, +5, +6]}, \nonumber \\
& \texttt{Elongate[0.5, 1.0]}, \texttt{Shorten[0.5, 1.0]}, \nonumber \\
&  \texttt{UniformTranspose[-3, -2, -1, +1, +2, +3]} \}
\label{eq:freyja-operations}
\end{align}

\medskip

Across 30 seeds, Freyja reached a mean best-so-far composite score equal to $\mu_{\mathrm{best}} = 0.8178$, with standard deviation $\sigma_{\mathrm{best}} = 0.0475$.

\medskip

The learning curve is provided in Appendix III, Figure \ref{fig:app-qasim-controlled-best-so-far}. Freyja achieved a higher best-so-far score than Qasim's, suggesting that action-space control can guide the search toward stronger note transformations.  Freyja's best-discovered holistic output is shown in Figure~\ref{fig:qasim-controlled-best-output}.

\begin{figure}[H]
\begin{center}\vspace{0.3cm}
\includegraphics[width=0.7\linewidth]{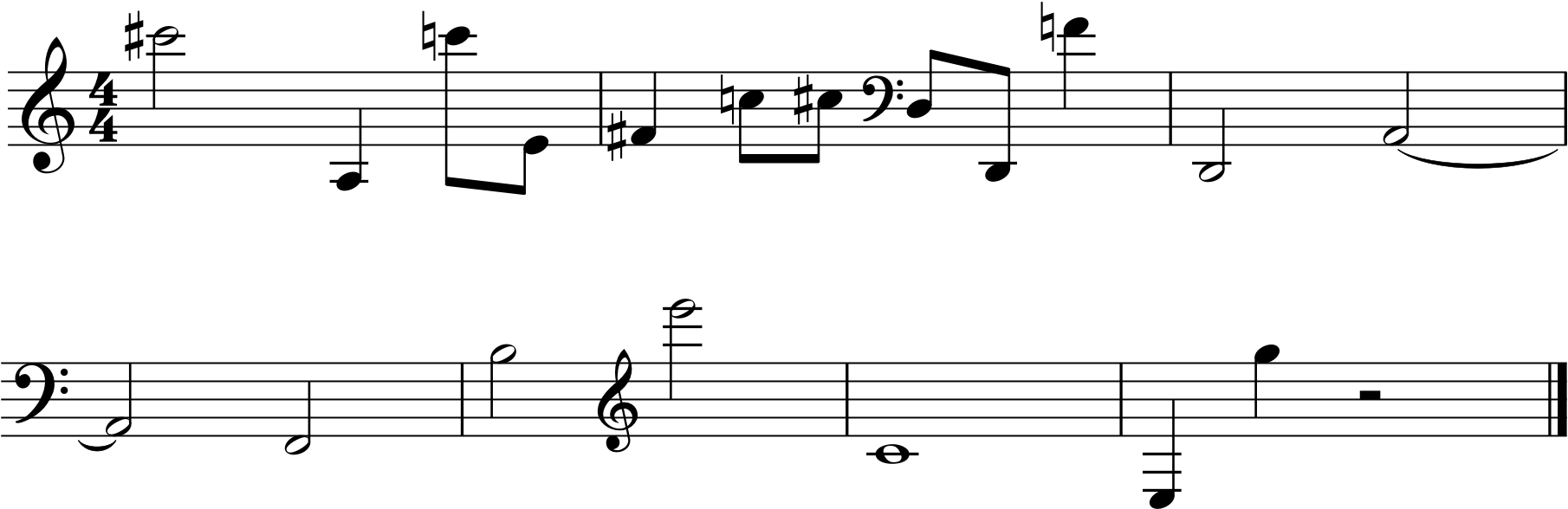}
\caption{Freyja's best-discovered output withomposite reward $\mathcal R_{\Sigma} = 0.9007$.}
\label{fig:qasim-controlled-best-output}
\end{center}
\end{figure}

The component scores for the best-so-far outputs were approximately equal to $\mathcal R_{\mathrm{div}} = 0.9444$, $\mathcal R_{\mathrm{pb}} = 0.7840$ and 
$\mathcal R_{\mathrm{sm}} = 0.7250$.

\medskip

Compared with Qasim, Freyja improved all three components, with the clearest gain in Perfect Balance. This suggests that constraining the action set does not simply reduce the agent's expressive capacity. Instead, it appears to guide the search toward transformations that are better aligned with the composite objective.

\medskip

For reference, after the fixed training budget of 50 episodes, Freyja achieved the following mean episode-50 composite score $\mu_{\mathrm{ep50}} = 0.4459$, with standard deviation 
$\sigma_{\mathrm{ep50}} = 0.1211$.

\medskip

Freyja is therefore best understood as a more stable holistic event agent. Its action-space design encourages more reliable discovery of balanced, diverse and smooth configurations.

\subsubsection{Comparing Holistic Agents}

Figure \ref{fig:holistic-comparison} compares the best-so-far learning curves for the two holistic agents. While both curves show clear improvement, Freyja's rises faster, remains above Qasim's for most of the training process, and reaches a higher final best-so-far score.

\begin{figure}[H]
\begin{center}\vspace{0.3cm}
\includegraphics[width=0.7\linewidth]{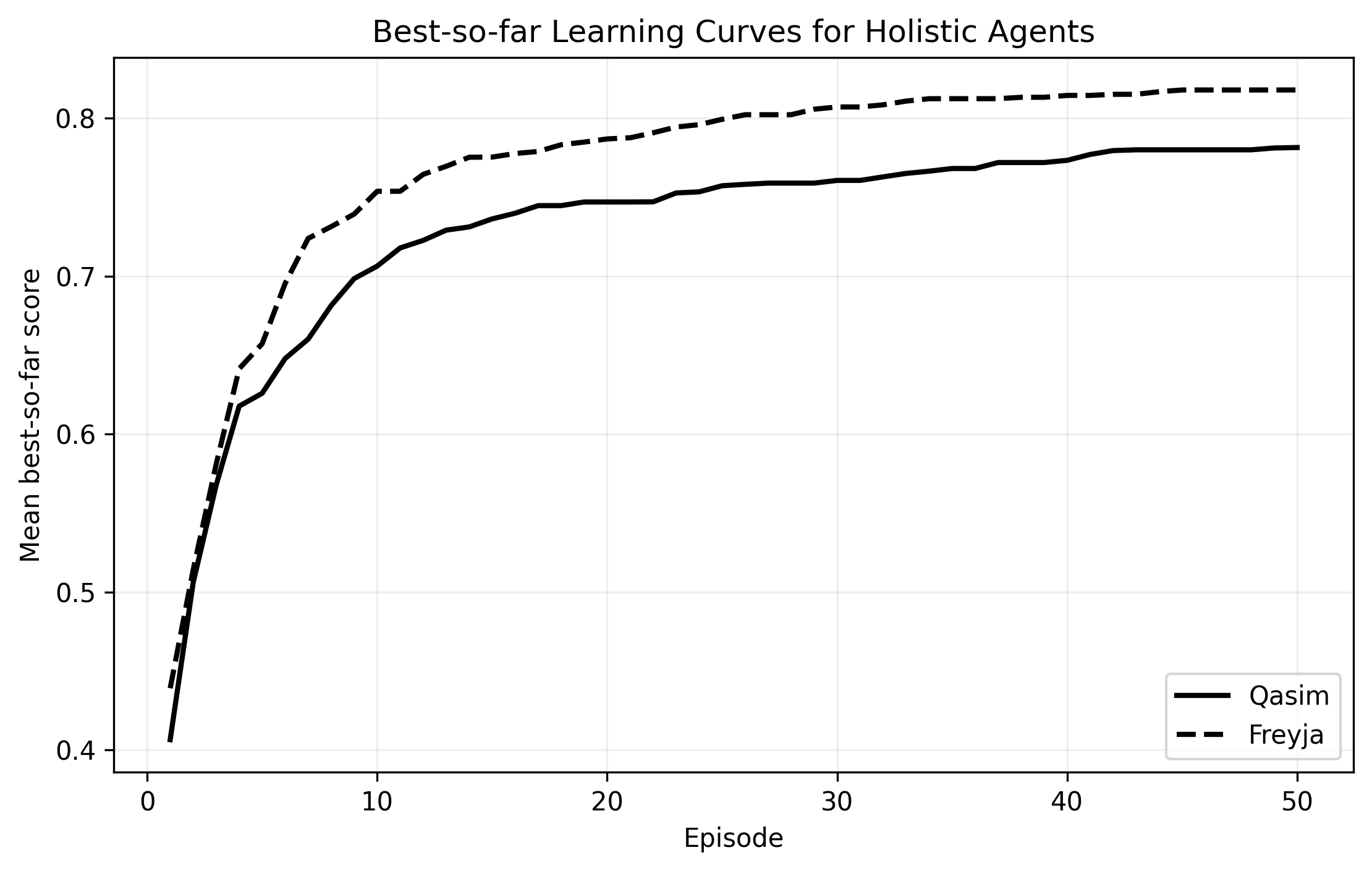}
\caption{Best-so-far learning curves for the two holistic pitch-duration agents. The controlled action set improves the efficiency and stability of holistic event learning.}
\label{fig:holistic-comparison}
\end{center}
\end{figure}

\begin{table}[H]
\centering
\caption{Summary of best-discovered results for the holistic pitch-duration experiments across 30 seeds. Both agents optimise the equally weighted composite reward $\mathcal R_{\Sigma}$.}
\label{tab:holistic-results-summary}
\begin{tabular}{llll}
\hline
Agent & Action style & Episode-50 score & Best-so-far score \\
\hline
Qasim & Open & $0.4307 \pm 0.2006$ & $0.7814 \pm 0.0635$ \\
Freyja & Controlled & $0.4459 \pm 0.1211$ & $0.8178 \pm 0.0475$ \\
\hline
\end{tabular}
\end{table}

The comparison highlights an important design principle for holistic musical agents, which may apply to atomistic ones as well. A larger or more open action set does not necessarily lead to better learning. Because the holistic agent already operates in a larger joint event space, unrestricted transformations can produce high exploratory variability. This variability may be musically useful when the goal is to generate diverse material, but it can also make learning less stable. By contrast, a more constrained action set reduces unproductive exploration while preserving enough expressive capacity for meaningful pitch-duration transformation. To conclude, Qasim behaved as an exploratory event transformer, while Freyja behaved as a more stable and efficient holistic agent. From a compositional standpoint, both are useful: the former offers broader exploratory variation, whereas the latter provides more reliable discovery of balanced, diverse and smooth pitch-duration material.

\subsubsection{Generalisation to Unseen Tunes}
\label{subsubsec:holistic-unseen-placeholder}

Finally, let us establish whether the holistic mindsets could be applied to unseen musical material. As in the previous generalisation tests, this was not a continuation of training: no additional episodes were run, no replay buffer was used, and the agent parameters were not updated. The highest-scoring trained model from each holistic experiment was applied to the unseen input in a single post-training pass.

\begin{figure}[H]
\begin{center}\vspace{0.3cm}
\includegraphics[width=0.7\linewidth]{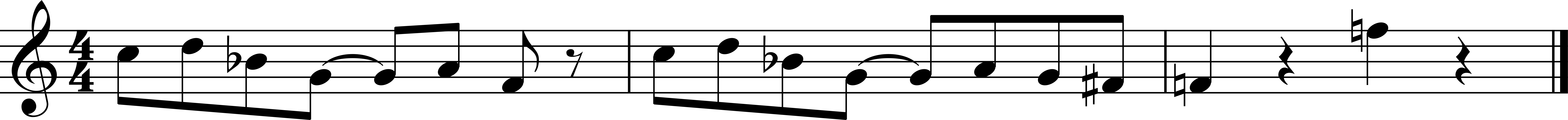}
\caption{Unseen tune for testing the holistic agents.}
\label{fig:unseen-holistic}
\end{center}
\end{figure}

The unseen tune in Figure~\ref{fig:unseen-holistic} has a composite score equal to $\mathcal R_{\Sigma} = 0.3261$. 

\medskip

To begin with, let us play the unseen tune to Qasim. The resulting output is shown in Figure \ref{fig:qasim-unseen-result}. Qasim increased the composite score from $\mathcal R_{\Sigma} = 0.3261$ to $\mathcal R_{\Sigma} = 0.6031$. This is a substantial improvement, which indicates that its evolved mindset is fit to process unseen music effectively. Note that because Qasim acts on notes, the improvement does not correspond only to pitch variation or rhythmic reorganisation, but to both, holistically.

\begin{figure}[H]
\begin{center}\vspace{0.3cm}
\includegraphics[width=0.7\linewidth]{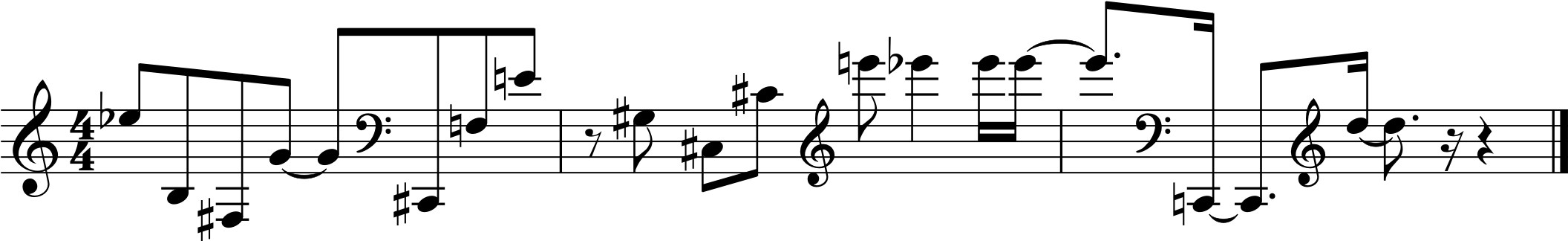}
\caption{Output from Qasim acting on the unseen tune in Figure~\ref{fig:unseen-holistic}. The composite score increased from $\mathcal R_{\Sigma} = 0.3261$ to $\mathcal R_{\Sigma} = 0.6031$.}
\label{fig:qasim-unseen-result}
\end{center}
\end{figure}

Next, it is Freyja's turn to process the unseen tune. The resulting output is shown in Figure \ref{fig:freyja-unseen-result}. Freyja increased the composite score from $\mathcal R_{\Sigma} = 0.3261$ to $\mathcal R_{\Sigma} = 0.5769$. This result confirms that the controlled holistic action set also transfers to unseen material. Although Freyja achieved a slightly lower best unseen score than Qasim in this example, it still improved the scores of the original input. This suggests that the more restrained action set remains expressive enough to reshape the new input, while retaining the more stable learning profile observed in the training exercise.

\begin{figure}[H]
\begin{center}\vspace{0.3cm}
\includegraphics[width=0.7\linewidth]{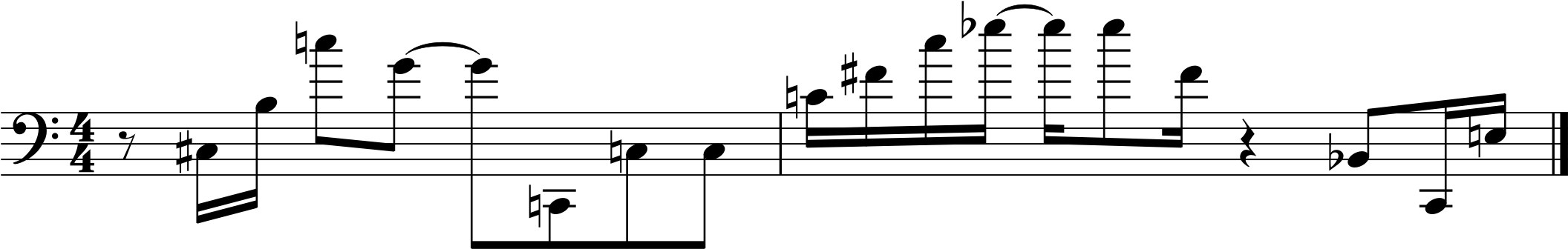}
\caption{Output from Freyja acting on the unseen tune in Figure \ref{fig:unseen-holistic}. The composite score increased from $\mathcal R_{\Sigma} = 0.3261$ to $\mathcal R_{\Sigma} = 0.5769$.}
\label{fig:freyja-unseen-result}
\end{center}
\end{figure}


\subsection{Interacting with the Agentic Orchestra}
\label{subsec:composing-agentic-orchestra}

\subsubsection{Interactive Performance}

The current version of the system works with music encoded as MIDI. In interactive mode, all agents make variations on the listened input on the fly. Then, the orchestra may select one of the variations, sequence some of them, or sequence all of them, respectively, for playback.  The system selects and sequences the variations at random or according to user-defined criteria.

\medskip

Normally, for live performance, the orchestra consists of atomistic rhythmic agents and holistic ones. Pitch-only agents are not so useful because the outputs lack rhythm. Conversely, the rhythmic agents ignore the pitches of inputs and output only durations with an irrelevant pitch, like those shown in Section \ref{subsec:rhythmic-mindsets}. This may be useful for playing non-pitched percussion instruments.

\medskip

In batch mode, the agents save their variations. This gives us the flexibility to compose offline with the generated materials, for example,  by uploading the outputs into a score editor or audio workstation for post-processing.

\medskip

The orchestra can be set to produce several variations per agent. Furthermore, the agents can be chained to transform the input one at a time. In this case, the output from one agent serves as the input for the next agent in the chain, and so on. Again, the results can be saved individually or appended into a single sequence of variations, or both. 

\medskip

As an example, Figure \ref{fig:three-sisters} shows a composition by three holistic agents: Clarice, Beatrice and Janice. In contrast to the previous examples, which were run on an IBM quantum computing simulator, these three agents were trained on a real IBM quantum computer, called \texttt{ibm\_fez} (IBM backend naming convention).

\begin{figure}[H]
\begin{center}\vspace{0.3cm}
\includegraphics[width=0.6\linewidth]{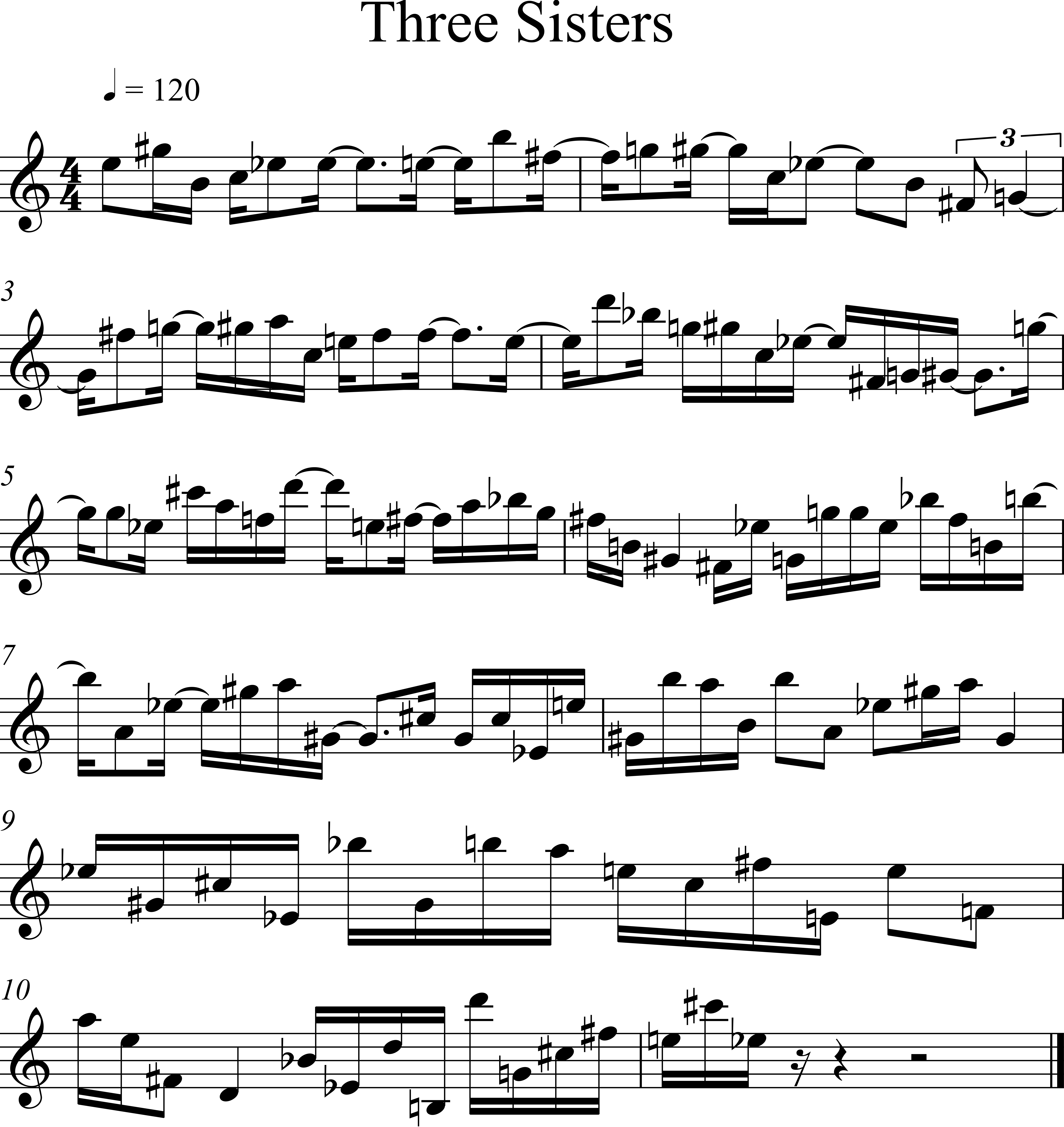}
\caption{An example of music generated by the agentic orchestra.}
\label{fig:three-sisters}
\end{center}
\end{figure}

The training took place over six epsiodes ($E = 6$) for each agent, with $B = 2$, $\xi = 1$ and $\epsilon_{\text{decay}} = 0.96$. The input was the snippet shown in Figure \ref{fig:short-jazzseed}. The reward functions and action operators were set as follows:

\begin{itemize}
\item{\textbf{Clarice}: $\mathcal R_{\text{pb}}$, $\mathcal A_{\text{Clarice}} = \{\texttt{Rotate}, \texttt{Shuffle}, \texttt{Retrograde}\}$.}
\item{\textbf{Beatrice}:  $\mathcal R_{\text{pb}}$, $\mathcal A_{\text{Beatrice}} = \{\texttt{Shuffle}, \texttt{Single-transpose}[-3, -2, -1, +1, +2, +3]\}$.}
\item{\textbf{Janice}: $ \mathcal R_{\Sigma} = (\mathcal R_{\text{div}} \mathcal R_{\text{pb}}, \mathcal R_{\text{sm}}) \div 3$,  $\mathcal A_{\text{Janice}} = \{\texttt{Shuffle}, \texttt{Silence},\texttt{Unsilence}, \\ \texttt{Single-transpose}[-3, -2, -1, +1, +2, +3], \texttt{Elongate}[0.5, 1, 2], \texttt{Shorten}[1]\}$.}
\end{itemize}

\begin{figure}[H]
\begin{center}\vspace{0.3cm}
\includegraphics[width=0.4\linewidth]{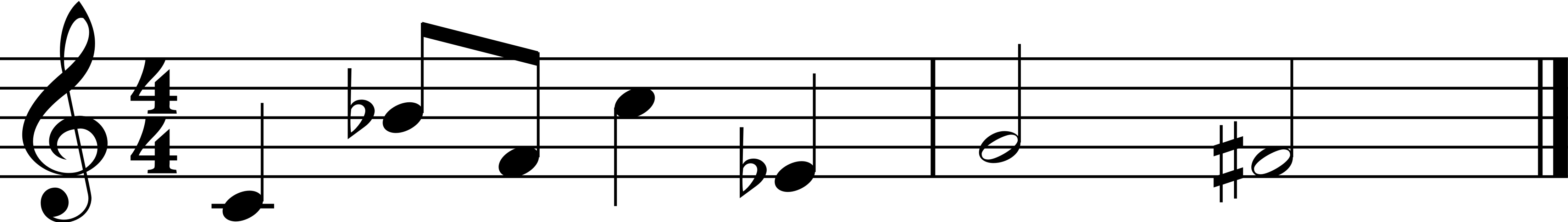}
\caption{Input tune for training Clarice, Beatrice and Janice on quantum hardware.}
\label{fig:short-jazzseed}
\end{center}
\end{figure}

\medskip

The resulting music in Figure \ref{fig:three-sisters} was composed over six iterations on the tune in Figure \ref{fig:short-jazzseed}. Each iteration was performed by the agents in chain mode: the output from one agent served as the input for the next. The outputs from each agent were appended one after another as they were executed. The order of the agents in the chain was selected by a quantum dice, as follows: Janice, Janice, Janice, Beatrice, Clarice and Beatrice.

\medskip

The hardware demonstrations were intended to explore the feasibility of the proposed musical workflow, rather than benchmark execution speed. End-to-end response times were not systematically measured. Such timings depend on the backend, workload, service scheduling, and classical processing involved. A controlled evaluation of interactive response latency across hardware configurations is therefore left for future work.

\subsubsection{Repurposing Experience}
\label{subsec:composing-single-agent}

The agents can be programmed to save the materials generated during training, that is, after each episode $E$. Thus, another option for composing offline is to repurpose the learning history of an agent, that is, the interim outputs from the learning episodes.
 
\medskip

The example in Figure \ref{fig:experience-remixing} shows an example of experience remixing with Alice's training history. In this case, the piece reflects the fact that Beatrice evolved a mindset that favours Perfect Balance ($\mathcal R_{\text{pb}}$). The Perfect Balance score for the training input (Figure \ref{fig:short-jazzseed}) is $\mathcal R_{\text{pb}} = 0.0917$. The score for the first two bars of the output is $\mathcal R_{\text{pb}} = 0.0917$, which improved to $\mathcal R_{\text{pb}} = 0.1269$ in the last two bars.

\begin{figure}[H]
\begin{center}\vspace{0.3cm}
\includegraphics[width=0.4\linewidth]{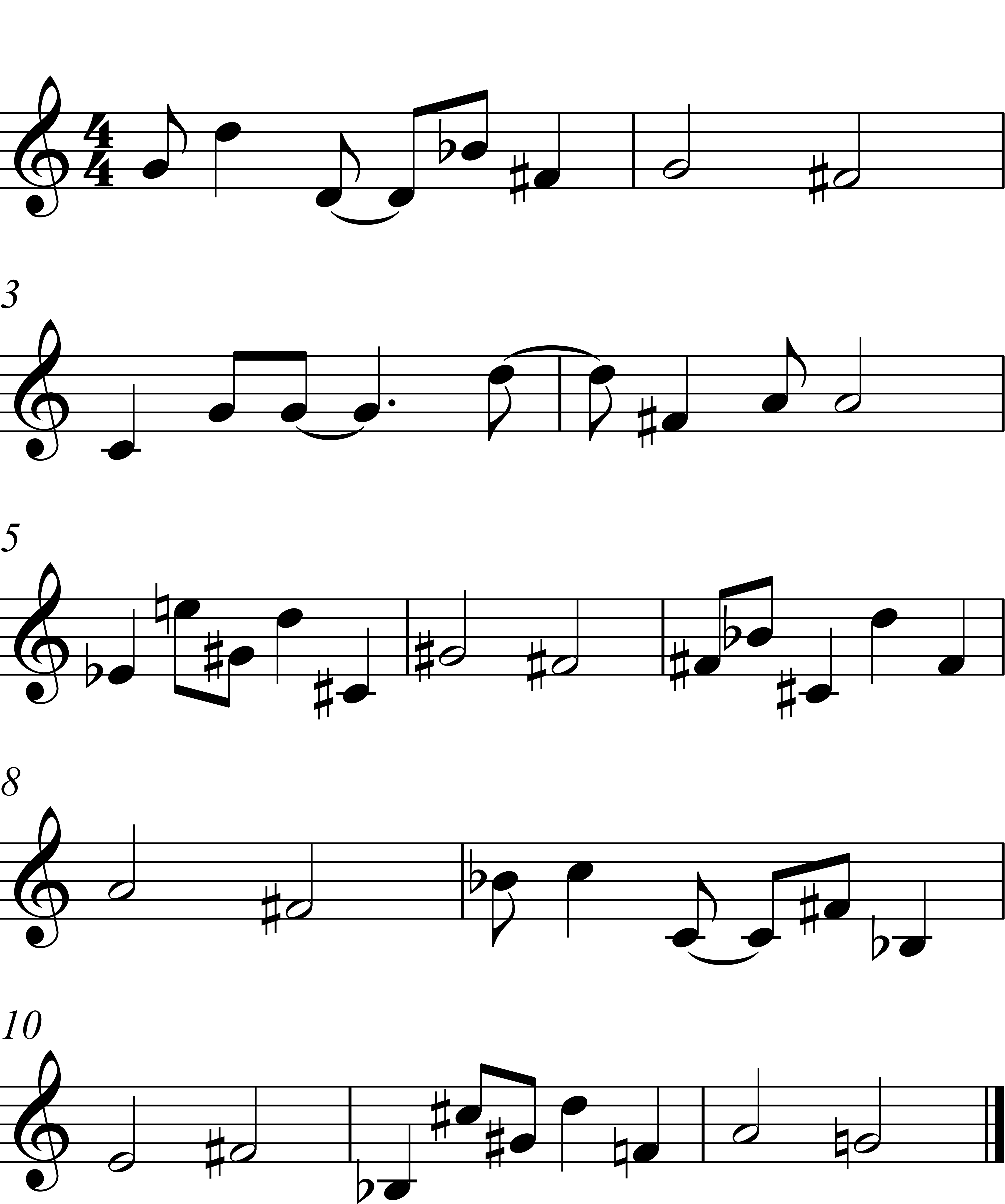}
\caption{Experience remixing with Beatrice's learning history.}
\label{fig:experience-remixing}
\end{center}
\end{figure}

\section{Conclusion}

This paper introduced a framework for computer-aided music composition and performance based on quantum reinforcement learning and multi-agent interaction. We articulated this through the concept of an agentic orchestra, in which musical agents evolve distinct mindsets via trial-and-error engagement with their environment. Implemented as parameterised quantum circuits within a hybrid quantum-classical loop, these agents use superposition and entanglement to explore musical decision spaces. We demonstrated this across two complementary architectures: atomistic and holistic. By prioritising process-oriented interaction and emergent behaviour over data-driven imitation, the framework shapes compositional styles through weighted reward functions. Consequently, musical trajectories develop gradually through active engagement rather than being reproduced from large training corpora.

\medskip

A central contribution of the work is demonstrating that parameterised quantum circuits can serve as effective decision-making mechanisms for musical agents within a reinforcement learning framework. A representative control experiment with the rhythmic agent Shin (Section \ref{subsubsec:shin-nonlearning-baselines}) distinguishes the contribution of parameter training from repeated random exploration: the trained agent achieved a mean best-so-far composite reward of 0.6276, compared with 0.4132 for random actions and 0.3790 for a frozen, untrained PQC, across 30 seeds. This result provides focused evidence that QRL training can improve the discovery of high-scoring musical variants under a fixed interaction budget, without establishing a general advantage over classical reinforcement learning. Even under current Noisy Intermediate-Scale Quantum (NISQ) limitations, the agents exhibit learning behaviour and produce distinct musical transformations, both in simulation and on real quantum hardware. While the musical representations employed here are deliberately constrained, they are sufficient to demonstrate the viability of quantum reinforcement learning for creative applications in music.

\medskip

The generalisation to unseen music examples presented above is very encouraging. The agents behave as we expected, and the agentic orchestra has been used for real musical performances. However, we ought to probe the agents' ability to transfer knowledge more rigorously to assess their behaviour systematically. 

\medskip

The system has been designed with scalability in mind. Compact encodings, modular agent definitions, and hybrid quantum-classical learning allow the framework to operate within present-day hardware limits while remaining extensible. As a natural progression from the agentic orchestra, multi-agent communication can be expanded by integrating a quantum teleportation protocol to enable direct state transfer and collaborative improvisation between agents. Implementing this protocol, as established in a recent interactive multi-agent system \cite{MirandaOshiro2026}, presents a native paradigm for distributed quantum agents. As quantum computing progresses towards larger, fault-tolerant technology, the agentic orchestra can expand. In this sense, the present work prepares the ground for scaling up quantum-assisted composition when quantum computers become larger, more reliable, and accessible to musicians.

\medskip

The framework also encourages a rethinking of compositional authorship by leveraging the distinctive computational properties of quantum mechanics. Rather than relying on deterministic rules or classical probabilistic sampling, musical outcomes arise from the interplay between reward design, agent configuration, human intervention, and inherently quantum processes. Compositional decisions are therefore distributed across multiple levels, inviting composers to work not only with musical materials but also with learning criteria, agent societies, and interaction strategies. In performance contexts, the compact expressivity of parameterised quantum circuits supports interactive music in which structure emerges organically through mutual adaptation between human performers and quantum agents.

\medskip

In conclusion, this research establishes an experimental foundation for quantum computer music grounded in reinforcement learning and agent-based interaction. By demonstrating that musically meaningful agentic systems can already be realised under NISQ conditions, the paper contributes both a working methodology and a conceptual framework for future developments. As quantum technologies mature, the agentic orchestra provides a scalable platform through which quantum computing may become an integral part of compositional practice, supporting new forms of emergent, interactive, and process-oriented creativity.

\medskip
\textbf{Acknowledgement: } The authors would like to thank James (Jim) Weaver at IBM Quantum for his valuable advice and suggestions, which helped improve the robustness of this work. Jim's contributions were made in an independent, personal capacity and his views do not necessarily reflect those of IBM.

\setcounter{secnumdepth}{0}
\section{Appendix I: Action Operators}
\label{app:appendix-operators}

\subsection{I.1 Duration Actions for Atomistic Agents}
\label{app:dur-operators}

When focusing on durations, the atomic agents do not take into account pitch information.

\begin{itemize}

\item{\texttt{Retrograde} - Reverse the order of the durations in the window: [\musHalf, \musEighth, \musEighth, \musQuarter] $\Rightarrow$ [{\color{red}{\musQuarter}}, {\color{red}{\musEighth}}, {\color{red}{\musEighth}}, {\color{red}{\musHalf}}].}

\item{\texttt{Rotate} - Reorder the durations in the window by moving the first duration figure to the end: [\musHalf, \musEighth, \musEighth, \musQuarter] $\Rightarrow$ [\musEighth, \musEighth, \musQuarter, {\color{red}{\musHalf}}].}

\item{\texttt{Shuffle}  - Shuffle the order of the durations in the window: [\musHalf, \musEighth, \musEighth, \musQuarter] $\Rightarrow$ [{\color{red}{\musEighth}}, {\color{red}{\musQuarter}}, {\color{red}{\musHalf}}, {\color{red}{\musEighth}}].} 

\item{\texttt{Mutate} - Change a random duration in the window by replacing it with one of the following \{\musWhole, \musHalf, \musQuarter, \musEighth, \musSixteenth\}:  [\musHalf, \musEighth, \musEighth, \musQuarter] $\Rightarrow$ [\musHalf, \musEighth, \musEighth,{\color{red}{ \musWhole}}]}.

\item{\texttt{Elongate} - Elongate the durations in the window by a factor of $0.25$ beat, $0.5$ beat, $1$ beat, or $2$: [\musHalf, \musEighth, \musEighth, \musQuarter]  $(+\musQuarter)\Rightarrow$  [{\color{red}{\musHalfDotted}}, {\color{red}{\musQuarterDotted}}, {\color{red}{\musQuarterDotted}}, {\color{red}{\musHalf}}].}

\item{\texttt{Shorten} - Shorten the durations of the notes in the window by a factor of  $0.25$ beat, $0.5$ beat, $1$ beat, or $2$ beats; if the result is zero or negative, then round to a sixteenth note: [\musHalf, \musEighth, \musEighth, \musQuarter]  $(\minus\musEighth)\Rightarrow$ [{\color{red}{\musQuarterDotted}}, {\color{red}{\musSixteenth}}, {\color{red}{\musSixteenth}}, {\color{red}{\musEighth}}].}

\end{itemize}

\subsection{I.2 Pitch Actions for Atomistic Agents}
\label{app:pitch-operators}

When focusing on pitch, the atomic agents do not take into account rhythmic information.

\begin{itemize}

\item{\texttt{Retrograde} - Reverse the order of the pitches in the window: [$C_3$, $D_3$, $E_3$, $F_3$] $\Rightarrow$ [{\color{red}{$F_3$}}, {\color{red}{$E_3$}}, {\color{red}{$D_3$}}, {\color{red}{$C_3$}}].}

\item{\texttt{Rotate} - Reorder the pitches in the window by moving the first pitch to the end: [$C_3$, $D_3$, $E_3$, $F_3$] $\Rightarrow$ [$D_3$, $E_3$, $F_3$, {\color{red}{$C_3$}}].}

\item{\texttt{Shuffle} - Shuffle the order of the pitches in the window: [$C_3$, $D_3$, $E_3$, $F_3$] $\Rightarrow$ [{\color{red}{$F_3$}}, {\color{red}{$C_3$}}, {\color{red}{$E_3$}}, {\color{red}{$D_3$}}].}

\item{\texttt{Mutate} - Change any pitch in the window randomly, within the interval of $\minus$7 and +7 semitones:  [$C_3$, $D_3$, $E_3$, $F_3$] $\Rightarrow$ [$C_3$, {\color{red}{$G_3$}}, $E_3$, $F_3$].}

\item{\texttt{Invert} - Flip upside down the pitch intervals in the window, changing ascending intervals to descending ones and vice-versa:  [$C_3$, $D_3$, $E_3$, $F_3$] ($\circ$, 2$\uparrow$, 2$\uparrow$, 2$\uparrow$) $\Rightarrow$ ($\circ$, 2$\downarrow$, 2$\downarrow$, 2$\downarrow$) [$C_3$, {\color{red}{${B\musFlat}_3$}}, {\color{red}{${G\musSharp}_2$}}, {\color{red}{$G_2$}}].}

\item{\texttt{Single-transpose} - Transpose one randomly selected pitch in the window by a random number of semitones, within a user-specified range; normally between $\minus$7 and +7 semitones: [$C_3$, $D_3$, $E_3$, $F_3$] (+5)$\Rightarrow$ [$C_3$, {\color{red}{$G_3$}}, $E_3$, $F_3$].}

\item{\texttt{Window-transpose} - Transpose the whole window by a randomly selected number of semitones within a user-specified range; normally between $\minus$7 and +7 semitones: [$C_3$, $D_3$, $E_3$, $F_3$] (+4)$\Rightarrow$ [{\color{red}{$E_3$}}, {\color{red}{$F\musSharp_3$}}, {\color{red}{$G_3$}}, {\color{red}{$B_4$}}].}

\item{\texttt{Reflect} - A type of pitch inversion, but around a given centre point. In practice, it is a pitch inversion followed by a window transposition: [$C_3$, $D_3$, $E_3$, $F_3$] ($\circ$, 2$\uparrow$, 2$\uparrow$, 2$\uparrow$) $(\minus 2)\Rightarrow$ ($\circ$, 2$\downarrow$, 2$\downarrow$, 2$\downarrow$) [{\color{red}{$B\musFlat_3$}}, {\color{red}{${G\musSharp}_2$}}, {\color{red}{${F\musSharp}_2$}}, {\color{red}{$F_2$}}] .}

\item{\texttt{Silence} - Replace a random pitch in the window by a pause: [$C_3$, $D_3$, $E_3$, $F_3$] $\Rightarrow$ [$C_3$, $D_3$, {\color{red}{$\varnothing$}}, $F_3$].}

\item{\texttt{Unsilence} - Replace a random pause (if any) in the window by a pitch that is already present in the window. Assign a random pitch if the entire window has only pauses.  [$C_3$, $\varnothing$, $E_3$, $F_3$] $\Rightarrow$  [$C_3$, {\color{red}{$F_3$}}, $E_3$, $F_3$].}

\end{itemize}

\subsection{I.3 Actions for Holistic Agents}
\label{app:note-operators}

\begin{itemize}

\item{\texttt{Retrograde} - Reverse the order of the notes in the window: [$C^{\musQuarter}_3$, $D^{\musQuarter}_3$, $E^{\musQuarter}_3$, $F^{\musQuarter}_3$] $\Rightarrow$ [{\color{red}{$F^{\musQuarter}_3$}}, {\color{red}{$E^{\musQuarter}_3$}}, {\color{red}{$D^{\musQuarter}_3$}}, {\color{red}{$C^{\musQuarter}_3$}}].}

\item{\texttt{Rotate} - Reorder the notes in the window by moving the first note to the end: [$C^{\musQuarter}_3$, $D^{\musQuarter}_3$, $E^{\musQuarter}_3$, $F^{\musQuarter}_3$] $\Rightarrow$ [$D^{\musQuarter}_3$, $E^{\musQuarter}_3$, $F^{\musQuarter}_3$, {\color{red}{$C^{\musQuarter}_3$}}].}

\item{\texttt{Shuffle} - Shuffle the order of the notes in the window: [$C^{\musQuarter}_3$, $D^{\musQuarter}_3$, $E^{\musQuarter}_3$, $F^{\musQuarter}_3$] $\Rightarrow$ [{\color{red}{$F^{\musQuarter}_3$}}, {\color{red}{$C^{\musQuarter}_3$}}, {\color{red}{$E^{\musQuarter}_3$}}, {\color{red}{$D^{\musQuarter}_3$}}].}

\item{\texttt{Mutate pitch} - Change any note in the window randomly, within the interval of $\minus$7 and +7 semitones:  [$C^{\musQuarter}_3$, $D^{\musQuarter}_3$, $E^{\musQuarter}_3$, $F^{\musQuarter}_3$] $\Rightarrow$ [$C^{\musQuarter}_3$, {\color{red}{$G^{\musQuarter}_3$}}, $E^{\musQuarter}_3$, $F^{\musQuarter}_3$].}

\item{\texttt{Invert pitches} - Flip upside down the pitch intervals in the window, changing ascending intervals to descending ones and vice-versa:  [$C^{\musQuarter}_3$, $D^{\musQuarter}_3$, $E^{\musQuarter}_3$, $F^{\musQuarter}_3$] ($\circ$, 2$\uparrow$, 2$\uparrow$, 2$\uparrow$) $\Rightarrow$ ($\circ$, 2$\downarrow$, 2$\downarrow$, 2$\downarrow$) [$C^{\musQuarter}_3$, {\color{red}{${B\musFlat}^{\musQuarter}_3$}}, {\color{red}{${G\musSharp}^{\musQuarter}_2$}}, {\color{red}{$
G^{\musQuarter}_2$}}].}

\item{\texttt{Single-transpose} - Tranpose one randomly selected note in the window by a random number of semitones between $\minus$7 and +7 semitones: [$C^{\musQuarter}_3$, $D^{\musQuarter}_3$, $E^{\musQuarter}_3$, $F^{\musQuarter}_3$] (+5)$\Rightarrow$ [$C^{\musQuarter}_3$, {\color{red}{$G^{\musQuarter}_3$}}, $E^{\musQuarter}_3$, $F^{\musQuarter}_3$].}

\item{\texttt{Window-transpose} - Tranpose the whole window by a randomly selected number of semitones between $\minus$7 and +7 semitones: [$C^{\musQuarter}_3$, $D^{\musQuarter}_3$, $E^{\musQuarter}_3$, $F^{\musQuarter}_3$] (+4)$\Rightarrow$ [{\color{red}{$E^{\musQuarter}_3$}}, {\color{red}{$F\musSharp^{\musQuarter}_3$}}, {\color{red}{$G^{\musQuarter}_3$}}, {\color{red}{$B^{\musQuarter}_4$}}].}

\item{\texttt{Reflect pitches} - A type of pitch inversion, but around a given centre point. In practice, tt is a pitch inversion followed by a window transposition:   [$C^{\musQuarter}_3$, $D^{\musQuarter}_3$, $E^{\musQuarter}_3$, $F^{\musQuarter}_3$] ($\circ$, 2$\uparrow$, 2$\uparrow$, 2$\uparrow$) $ (\minus 2) \Rightarrow$ ($\circ$, 2$\downarrow$, 2$\downarrow$, 2$\downarrow$) [{\color{red}{$B\musFlat^{\musQuarter}_3$}}, {\color{red}{${G\musSharp}^{\musQuarter}_2$}}, {\color{red}{${F\musSharp}^{\musQuarter}_2$}}, {\color{red}{$F^{\musQuarter}_2$}}].}

\item{\texttt{Silence} - Replace a random note in the window by a pause, of equivalent duration: [$C^{\musQuarter}_3$, $D^{\musQuarter}_3$, $E^{\musQuarter}_3$, $F^{\musQuarter}_3$] $\Rightarrow$ [$C^{\musQuarter}_3$, $D^{\musQuarter}_3$, {\color{red}{$\varnothing^{\musQuarter}$}}, $F^{\musQuarter}_3$] }

\item{\texttt{Unsilence} - Replace a random pause (if any) in the window by a pitch that is already present in the window. Assign a random pitch if the entire window has only pauses: [$C^{\musQuarter}_3$, $\varnothing^{\musQuarter}$, $E^{\musQuarter}_3$, $F^{\musQuarter}_3$] $\Rightarrow$ [$C^{\musQuarter}_3$, {\color{red}{$F^{\musQuarter}_3$}}, $E^{\musQuarter}_3$, $F^{\musQuarter}_3$].}

\item{\texttt{Elongate} - Elongate the durations of the notes in the window by a factor of $0.25$ beat, $0.5$ beat, $1$ beat, or $2$: [$C^{\musQuarter}_3$, $D^{\musQuarter}_3$, $E^{\musQuarter}_3$, $F^{\musQuarter}_3$] (+\musQuarter)$\Rightarrow$ [$C^{\color{red}{{\musHalf}}}_3$, $D^{\color{red}{{\musHalf}}}_3$, $E^{\color{red}{{\musHalf}}}_3$, $F^{\color{red}{{\musHalf}}}_3$].}

\item{\texttt{Shorten} - Shorten the durations of the notes in the window by a factor of  $0.25$ beat, $0.5$ beat, $1$ beat, or $2$ beats; if the result is zero or negative, then round to a sixteenth note: [$C^{\musQuarter}_3$, $D^{\musQuarter}_3$, $E^{\musQuarter}_3$, $F^{\musQuarter}_3$] ($\minus$\musEighth)$\Rightarrow$ [$C^{\color{red}{{\musEighth}}}_3$, $D^{\color{red}{{\musEighth}}}_3$, $E^{\color{red}{{\musEighth}}}_3$, $F^{\color{red}{{\musEighth}}}_3$].}

\end{itemize}

\setcounter{secnumdepth}{0}
\section{Appendix II: Qiskit Code}
\label{app:qiskit-code}

\begin{python}
import numpy as np
from qiskit import QuantumCircuit

# ----------------------------
# Configuration
# ----------------------------
nq = 24          # number of qubits (atomistic pitch agent, window size 4)
L = 2            # number of layers
theta = np.random.uniform(0, 2*np.pi, size=2*L*nq)

qc = QuantumCircuit(nq)

# ----------------------------
# 1. State encoding (example placeholder)
# ----------------------------
# Assume binary_state[q] = 1 if the q-th bit of s_t is 1
# Here we just illustrate the mechanism
binary_state = np.random.randint(0, 2, nq)

for q in range(nq):
    if binary_state[q] == 1:
        qc.ry(np.pi / 2, q)

# ----------------------------
# 2. PQC Ansatz (Eq. 2)
# ----------------------------
param_idx = 0

for layer in range(L):
    # Rotation layer
    for q in range(nq):
        qc.rz(theta[param_idx], q)
        param_idx += 1
        qc.ry(theta[param_idx], q)
        param_idx += 1

    # Entangling layer (linear chain)
    for q in range(nq - 1):
        qc.cx(q, q + 1)

# ----------------------------
# 3. Measurement
# ----------------------------
qc.measure_all()

qc
\end{python}

\section{Appendix III: Learning Curves}
\label{app:individual-learning-curves}

This appendix provides the individual best-so-far learning curves for the agents discussed in Section~\ref{sec:demos}. These plots are separated from the main demonstrations section to keep the main text focused on musical interpretation and comparative results.

\subsection{III.1 Atomistic Duration Agents}
\begin{figure}[H]
\begin{center}\vspace{0.3cm}
\includegraphics[width=0.6\linewidth]{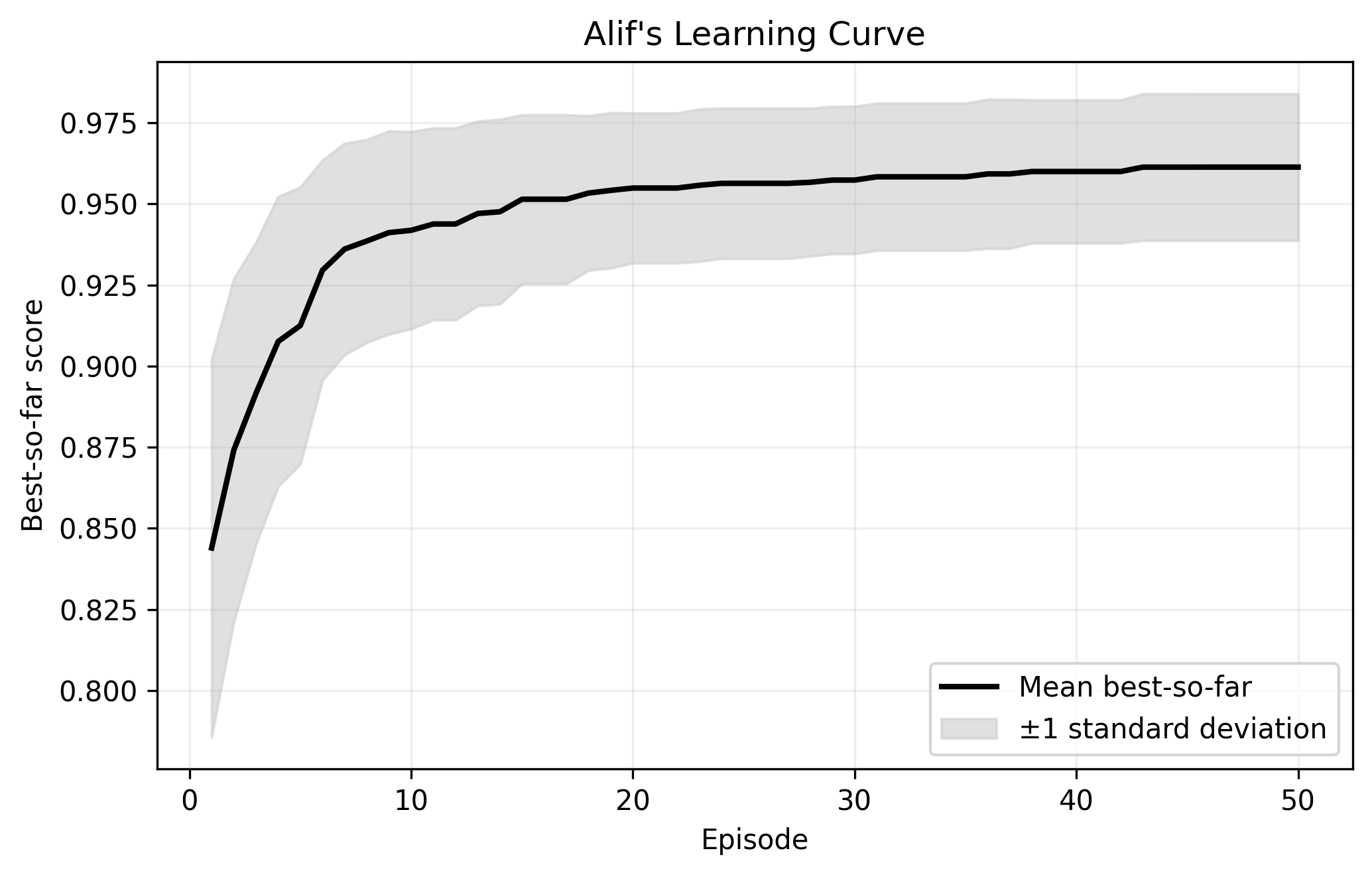}
\caption{Alif's best-so-far Perfect Balance score across 30 seeds.}
\label{fig:app-alif-best-so-far}
\end{center}
\end{figure}

\begin{figure}[H]
\begin{center}\vspace{0.3cm}
\includegraphics[width=0.6\linewidth]{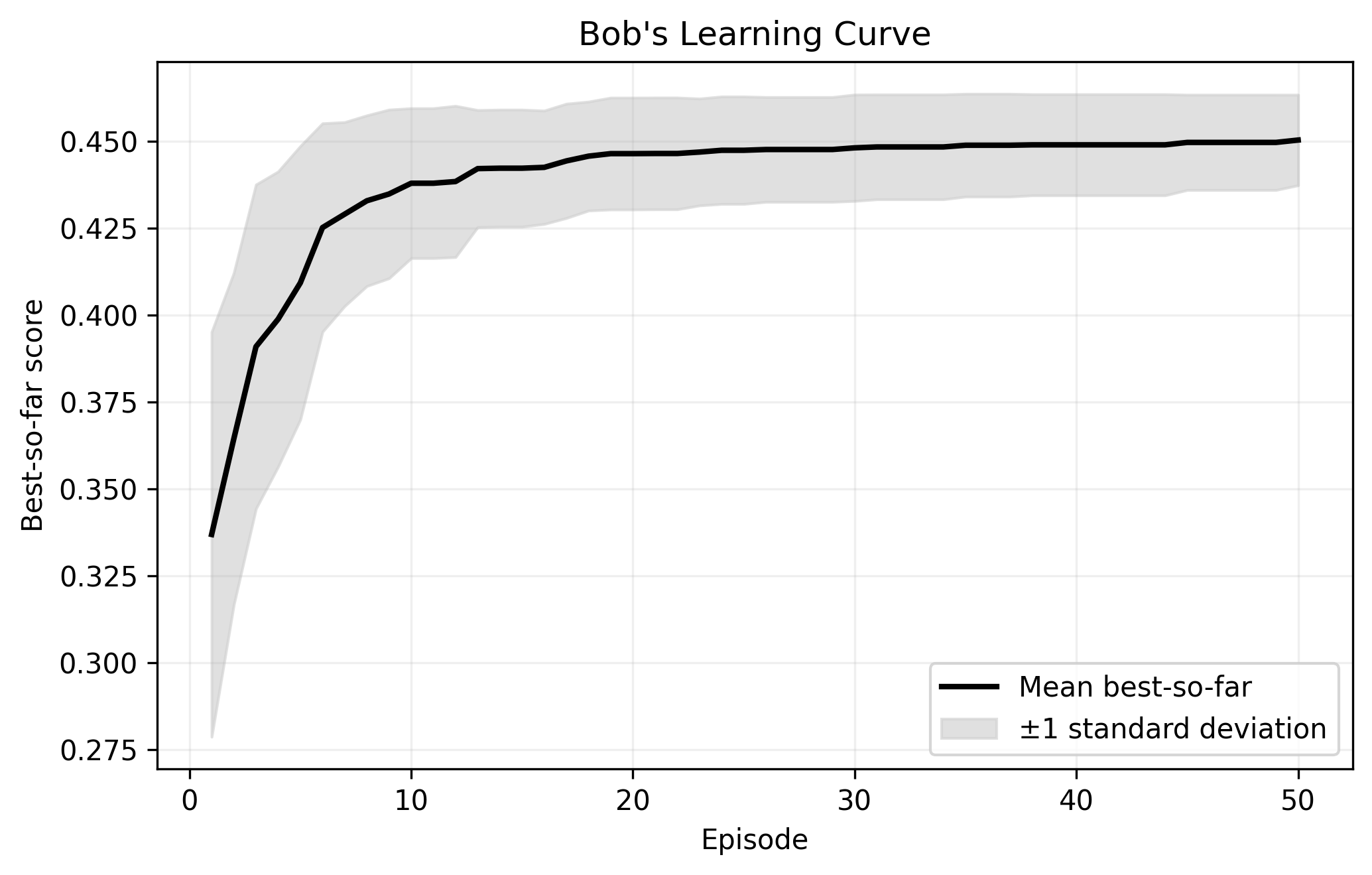}
\caption{Bob's best-so-far composite rhythmic score across 30 seeds.}
\label{fig:app-bob-best-so-far}
\end{center}
\end{figure}

\begin{figure}[H]
\begin{center}\vspace{0.3cm}
\includegraphics[width=0.6\linewidth]{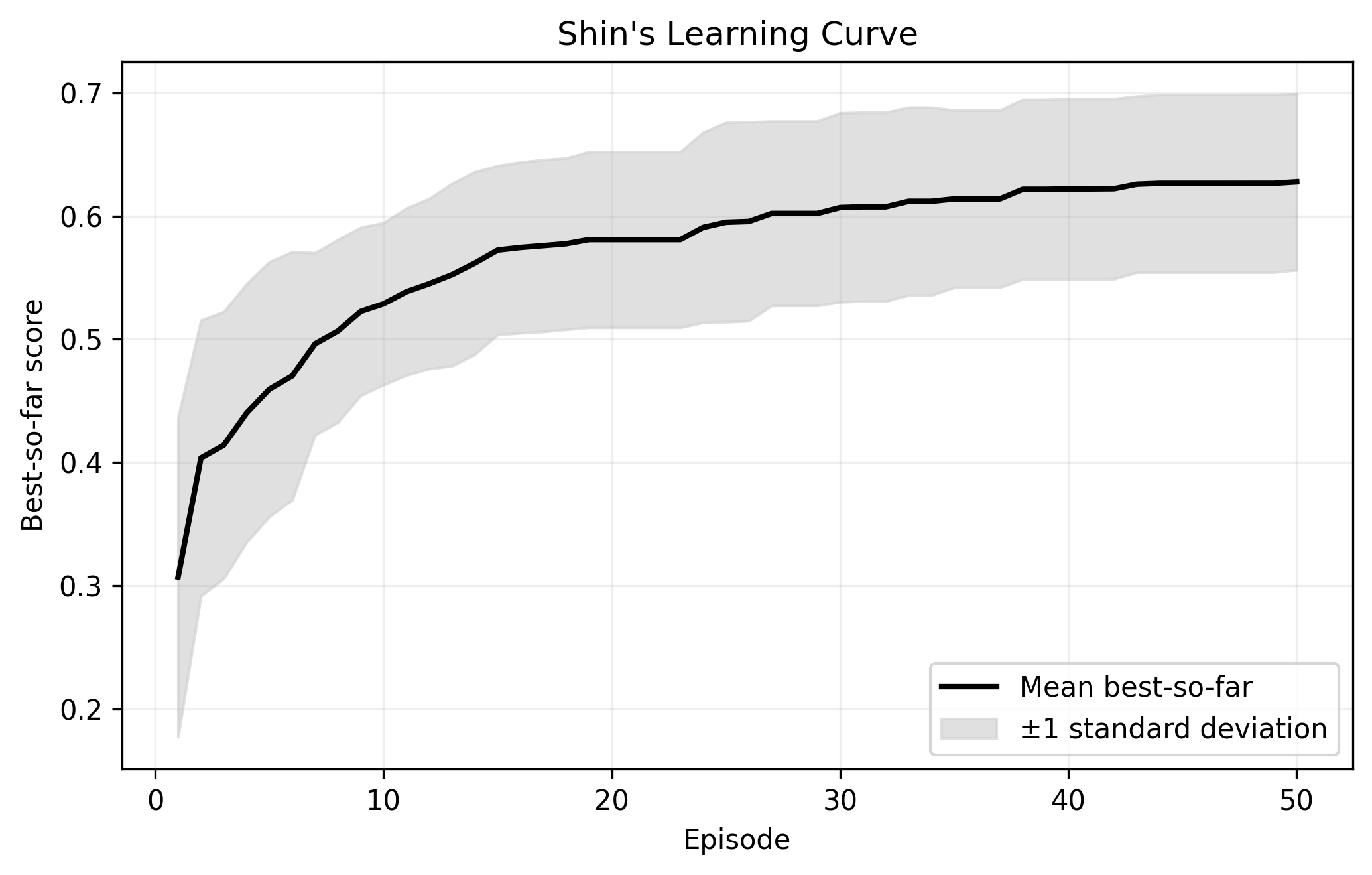}
\caption{Shin's best-so-far composite rhythmic score across 30 seeds.}
\label{fig:app-shin-best-so-far}
\end{center}
\end{figure}

\subsection{III.2 Atomistic Pitch Agents}

\begin{figure}[H]
\begin{center}\vspace{0.3cm}
\includegraphics[width=0.6\linewidth]{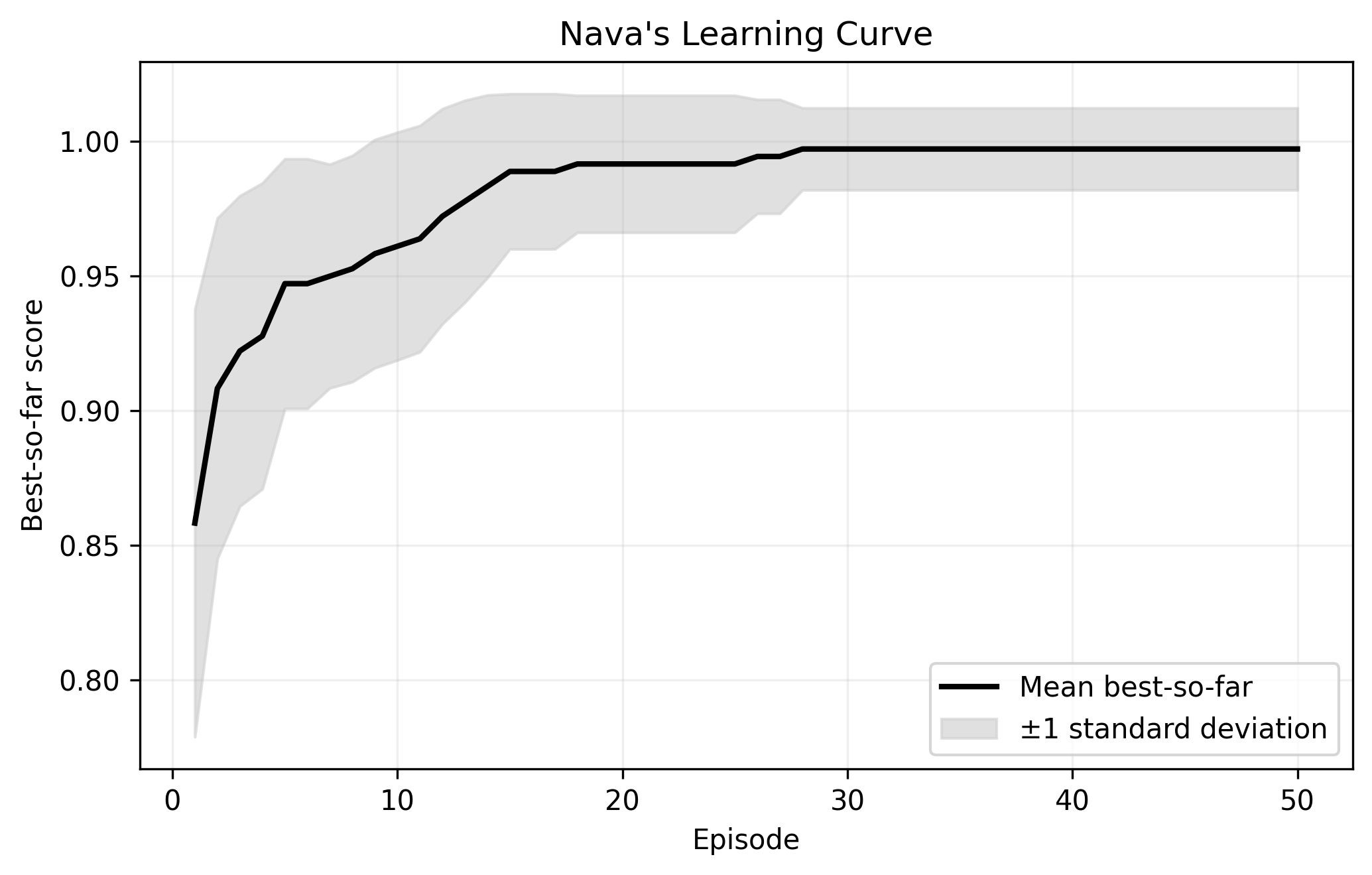}
\caption{Nava's best-so-far Diversity score across 30 seeds.}
\label{fig:app-nava-best-so-far}
\end{center}
\end{figure}

\begin{figure}[H]
\begin{center}\vspace{0.3cm}
\includegraphics[width=0.6\linewidth]{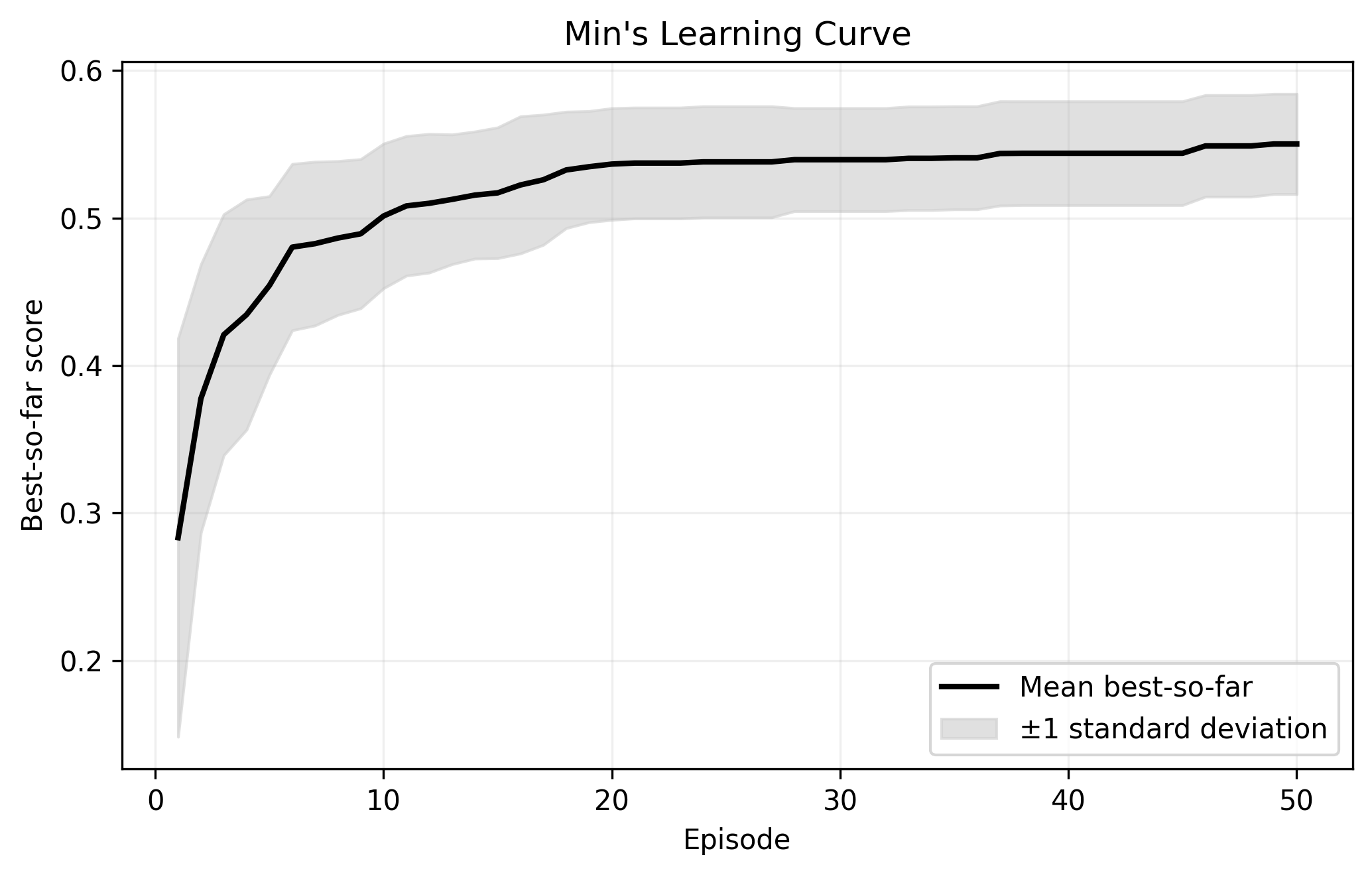}
\caption{Min's best-so-far composite pitch score across 30 seeds.}
\label{fig:app-min-best-so-far}
\end{center}
\end{figure}

\subsection{III.3 Holistic Pitch-Duration Agents}

\begin{figure}[H]
\begin{center}\vspace{0.3cm}
\includegraphics[width=0.6\linewidth]{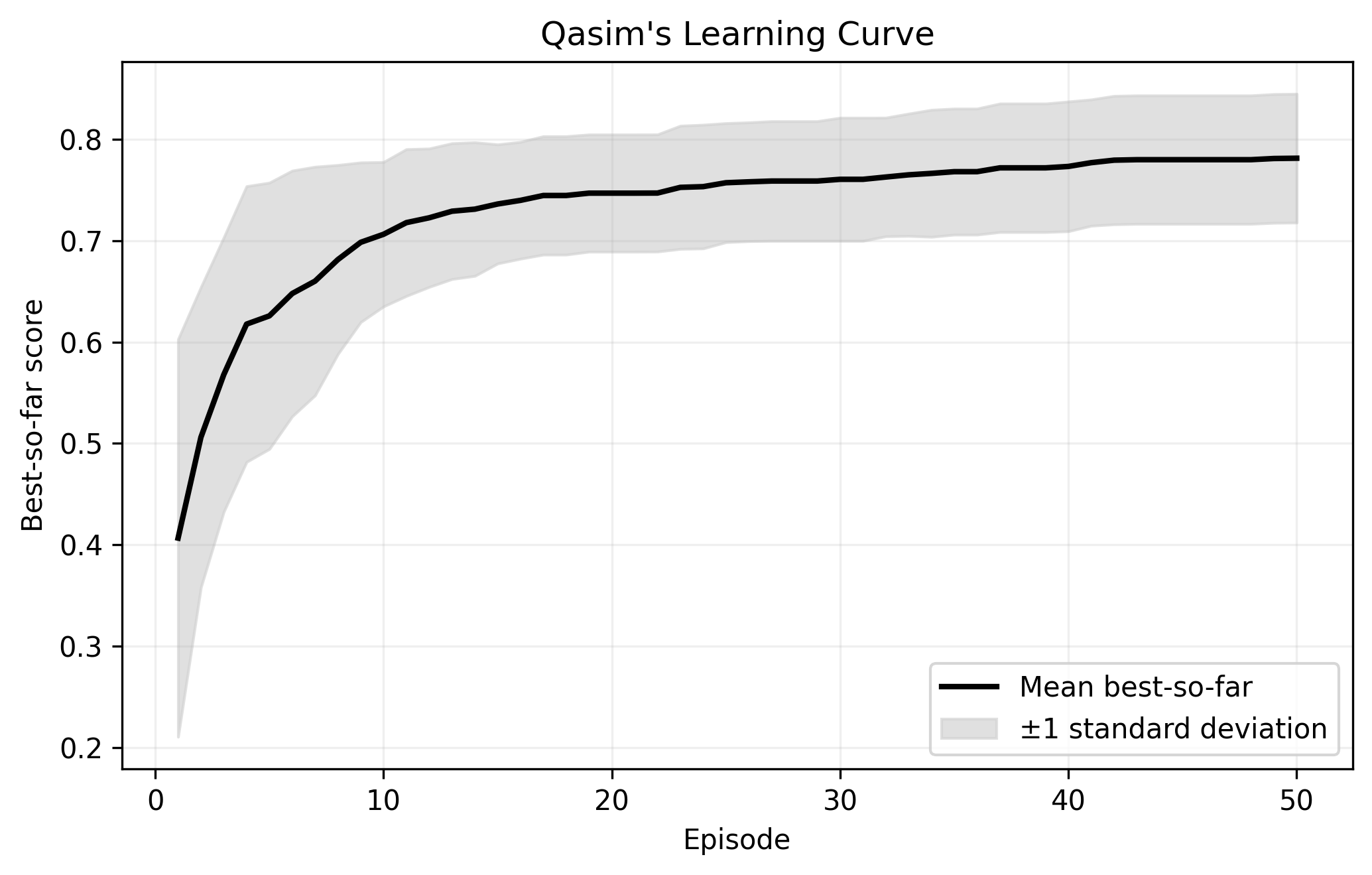}
\caption{Qasim's best-so-far composite score across 30 seeds.}
\label{fig:app-qasim-best-so-far}
\end{center}
\end{figure}

\begin{figure}[H]
\begin{center}\vspace{0.3cm}
\includegraphics[width=0.6\linewidth]{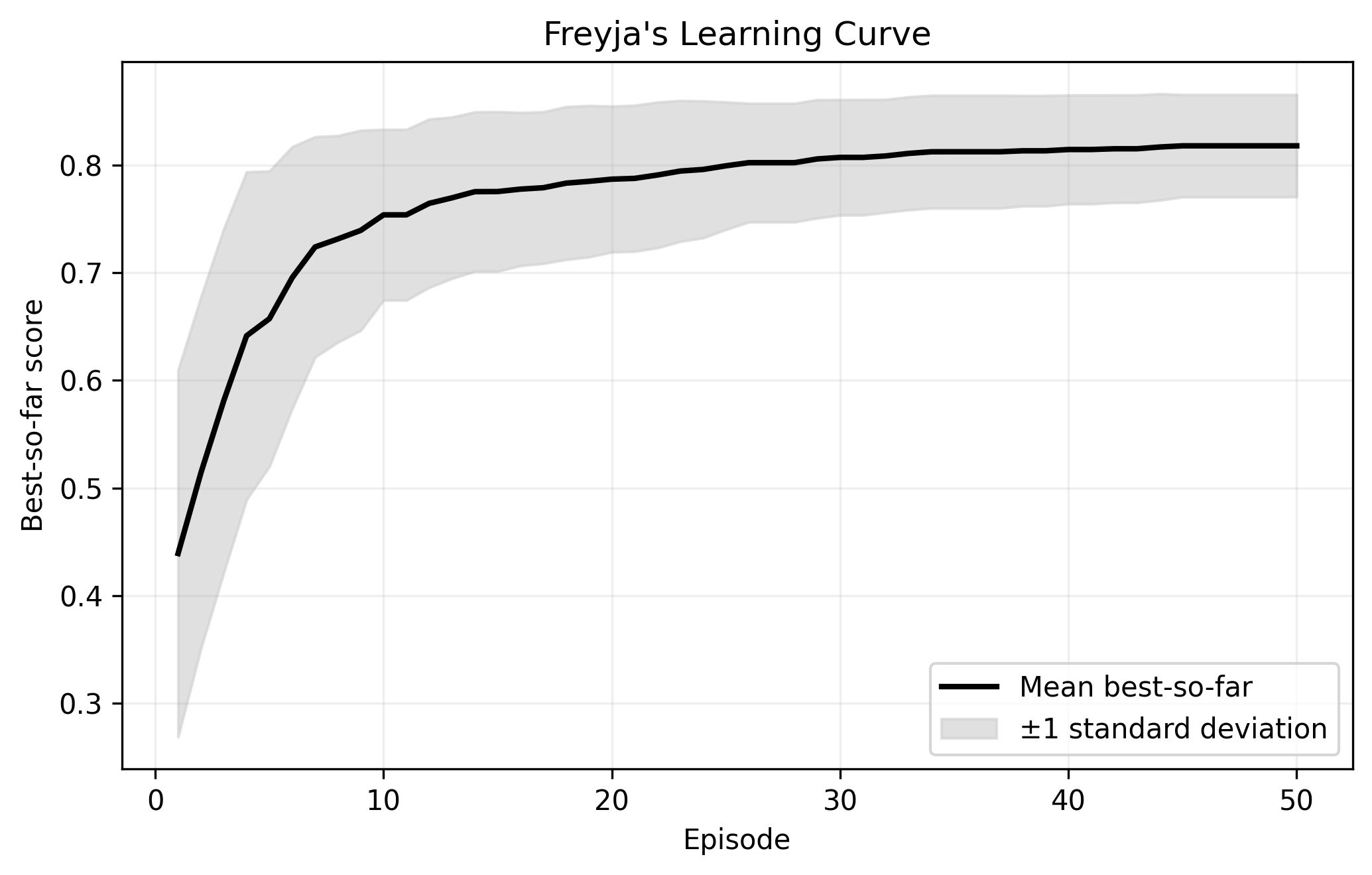}
\caption{Qasim-controlled's best-so-far composite score across 30 seeds.}
\label{fig:app-qasim-controlled-best-so-far}
\end{center}
\end{figure}



\end{document}

%% file: qcs.tikzstyles
\tikzstyle{node}=[minimum size=0.3cm]
\tikzstyle{Z}=[fill={rgb,255:red,230; green,254; blue,230}, draw={rgb,255: red,61; green,77; blue,61}, shape=circle]
\tikzstyle{X}=[fill={rgb,255:red,255; green,135; blue,136}, draw={rgb,255: red,102; green,54; blue,54}, shape=circle]
\tikzstyle{Y}=[fill=zxblue, draw=zxdblue, shape=circle]
\tikzstyle{Z_big}=[fill={rgb,255:red,230; green,254; blue,230}, draw={rgb,255: red,61; green,77; blue,61}, shape=circle, minimum width=1.6em, font={\small}]
\tikzstyle{X_big}=[fill={rgb,255:red,255; green,135; blue,136}, draw={rgb,255: red,102; green,54; blue,54}, shape=circle, minimum width=1.6em, font={\small}]
\tikzstyle{Y_big}=[fill=zxblue, draw=zxdblue, shape=circle, minimum width=1.6em, font={\small}]
\tikzstyle{Z_tri}=[fill={rgb,255:red,230; green,254; blue,230}, draw={rgb,255: red,61; green,77; blue,61}, regular polygon, regular polygon sides=3, draw, shape border rotate=0, inner sep=0pt, minimum width=15pt, line width=0.75]
\tikzstyle{X_tri}=[fill={rgb,255:red,255; green,135; blue,136}, draw={rgb,255: red,102; green,54; blue,54}, regular polygon, regular polygon sides=3, draw, shape border rotate=0, inner sep=0pt, minimum width=15pt, line width=0.75]
\tikzstyle{Z_tri_inv}=[fill={rgb,255:red,230; green,254; blue,230}, draw={rgb,255: red,61; green,77; blue,61}, regular polygon, regular polygon sides=3, draw, shape border rotate=180, inner sep=0pt, minimum width=15pt, line width=0.75, font={\footnotesize}]
\tikzstyle{X_tri_inv}=[fill={rgb,255:red,255; green,135; blue,136}, draw={rgb,255: red,102; green,54; blue,54}, regular polygon, regular polygon sides=3, draw, shape border rotate=180, inner sep=0pt, minimum width=15pt, line width=0.75]
\tikzstyle{Z_tri_l}=[fill={rgb,255:red,230; green,254; blue,230}, draw={rgb,255: red,61; green,77; blue,61}, regular polygon, regular polygon sides=3, draw, shape border rotate=90, inner sep=0pt, minimum width=15pt, line width=0.75]
\tikzstyle{X_tri_l}=[fill={rgb,255:red,255; green,135; blue,136}, draw={rgb,255: red,102; green,54; blue,54}, regular polygon, regular polygon sides=3, draw, shape border rotate=90, inner sep=0pt, minimum width=15pt, line width=0.75]
\tikzstyle{H}=[fill=yellow, draw=black, shape=rectangle, minimum width=2mm, minimum height=2mm]
\tikzstyle{Y_box}=[draw=black, shape=rectangle, minimum width=2mm, minimum height=2mm, tikzit fill={rgb,255: red,255; green,128; blue,0}]
\tikzstyle{Z_med}=[fill={rgb,255:red,230; green,254; blue,230}, draw={rgb,255: red,61; green,77; blue,61}, shape=circle, minimum width=1.1em, font={\footnotesize}]
\tikzstyle{X_med}=[fill={rgb,255:red,255; green,135; blue,136}, draw={rgb,255: red,102; green,54; blue,54}, shape=circle, minimum width=1.1em, font={\footnotesize}]
\tikzstyle{Y_med}=[fill=zxblue, draw=zxdblue, font={\footnotesize}, minimum width=1.1em, font={\footnotesize}, shape=circle]
\tikzstyle{ZP}=[fill={rgb,255:red,230; green,254; blue,230}, draw={rgb,255: red,61; green,77; blue,61}, regular polygon, regular polygon sides=3, shape border rotate=30]
\tikzstyle{ZM}=[fill={rgb,255:red,230; green,254; blue,230}, draw={rgb,255: red,61; green,77; blue,61}, regular polygon, regular polygon sides=3, shape border rotate=-30]
\tikzstyle{XP}=[fill={rgb,255:red,255; green,135; blue,136}, draw={rgb,255: red,102; green,54; blue,54}, regular polygon, regular polygon sides=3, shape border rotate=30]
\tikzstyle{XM}=[fill={rgb,255:red,255; green,135; blue,136}, draw={rgb,255: red,102; green,54; blue,54}, regular polygon, regular polygon sides=3, shape border rotate=-30]
\tikzstyle{small_box}=[fill=white, draw=black, shape=rectangle, minimum width=1.5cm, minimum height=1.5cm, font={\footnotesize}]
\tikzstyle{med_rectangle}=[fill=white, draw=black, shape=rectangle, minimum width=2.8cm, minimum height=1.5cm, font={\footnotesize}]
\tikzstyle{big_rectangle}=[fill=white, draw=black, shape=rectangle, minimum width=4.5cm, minimum height=1.5cm, font={\footnotesize}]
\tikzstyle{smol_box}=[fill=white, draw=black, shape=rectangle, minimum width=1cm, minimum height=1cm]
\tikzstyle{poo}=[minimum height=0.7cm, minimum width=0.7cm, path picture={\node at (path picture bounding box.center) {\includegraphics[width=0.7cm] {figures/poo}};}]
\tikzstyle{Z_long}=[fill={rgb,255:red,230; green,254; blue,230}, draw={rgb,255: red,61; green,77; blue,61}, shape=rectangle, rounded corners=0.25cm, minimum height=0.5cm, inner sep=0.25em, font={\scriptsize}]
\tikzstyle{X_long}=[fill={rgb,255:red,255; green,135; blue,136}, draw={rgb,255: red,102; green,54; blue,54}, shape=rectangle, rounded corners=0.25cm, minimum height=0.5cm, inner sep=0.25em, font={\scriptsize}]
\tikzstyle{med_rectv}=[fill=white, draw=black, shape=rectangle, minimum width=1.1cm, minimum height=2.4cm, font={\scriptsize}, inner sep=0.2cm]
\tikzstyle{Z dot}=[fill={rgb,255: red,0; green,127; blue,0}, draw=black, shape=circle, minimum width=1.5mm]
\tikzstyle{X dot}=[fill={rgb,255:red,255; green,21; blue,0}, draw=black, shape=circle, minimum width=1.5mm]
\tikzstyle{Z phase dot}=[fill={rgb,255: red,0; green,127; blue,0}, draw=black, shape=circle, font={\footnotesize}]
\tikzstyle{X phase dot}=[fill={rgb,255:red,255; green,21; blue,0}, draw=black, shape=circle, font={\footnotesize}]
\tikzstyle{ZYa}=[draw=black, shape=rectangle, rectangle split, rectangle split parts=2, rectangle split horizontal, rectangle split part fill={zxgreen, zxblue}, rectangle split draw splits=false, minimum height=2mm, font={\tiny}]
\tikzstyle{YZ}=[draw=black, shape=rectangle, rectangle split, rectangle split parts=2, rectangle split horizontal, rectangle split part fill={zxblue, zxgreen}, rectangle split draw splits=false, minimum height=2mm, font={\tiny}]
\tikzstyle{XYa}=[draw=black, shape=rectangle, rectangle split, rectangle split parts=2, rectangle split horizontal, rectangle split part fill={zxred, zxblue}, rectangle split draw splits=false, minimum height=2mm, font={\tiny}]
\tikzstyle{YX}=[draw=black, shape=rectangle, rectangle split, rectangle split parts=2, rectangle split horizontal, rectangle split part fill={zxblue, zxred}, rectangle split draw splits=false, minimum height=2mm, font={\tiny}]
\tikzstyle{tiny_box}=[fill=white, draw=black, shape=rectangle, minimum width=1cm, minimum height=1cm, font={\footnotesize}]
\tikzstyle{scalar}=[fill=white, draw=black, shape=diamond, font={\scriptsize}]
\tikzstyle{black_dot}=[fill=black, draw=black, shape=circle, inner sep=0pt, minimum size=0.2cm, text height=2pt, text depth=0pt]
\tikzstyle{white_dot}=[fill=none, draw=black, shape=circle, inner sep=0pt, minimum size=0.4cm]

\tikzstyle{dashs}=[-, dashed, line width=0.15mm]
\tikzstyle{thick}=[-, line width=0.5mm]
\tikzstyle{arrow}=[->]
\tikzstyle{invisible}=[-, draw=none]
\tikzstyle{functor}=[-, fill={rgb,255: red,240; green,240; blue,240}]
\tikzstyle{boxedge}=[-, fill=white]
\tikzstyle{dashed_line}=[dashed, gray, line width=0.25mm]

%% file: dots.tikzstyles
\tikzstyle{black_dot}=[fill=black, draw=black, shape=circle, inner sep=0pt, minimum size=0.1cm, text height=2pt, text depth=0pt]
\tikzstyle{white_dot}=[fill=none, draw=black, shape=circle, inner sep=0pt, minimum size=0.2cm]

\tikzstyle{dashed_line}=[dashed, gray, line width=0.25mm]

%% file: colors.tikzstyles
\tikzstyle{color1}=[fill={rgb,255: red,155; green,155; blue,155}, line width=0.25mm]
\tikzstyle{color2}=[fill={rgb,255: red,180; green,180; blue,180}, line width=0.25mm]
\tikzstyle{color3}=[fill={rgb,255: red,205; green,205; blue,205}, line width=0.25mm]
\tikzstyle{color4}=[fill={rgb,255: red,230; green,230; blue,230}, line width=0.25mm]
\tikzstyle{color5}=[fill={rgb,255: red,255; green,255; blue,255}, line width=0.25mm]